\documentclass[nohyperref]{article}
\usepackage{hyperref}

\usepackage[accepted]{icml2023}

\usepackage{amsmath}
\usepackage{amssymb}
\usepackage{mathtools}
\usepackage{amsthm}

\usepackage[capitalize,noabbrev]{cleveref}

\usepackage{tikz}
\usepackage{thmtools,thm-restate}

\usetikzlibrary{arrows,automata,positioning}
\newcommand\interval{0.3cm}
\tikzset{ 
    roundnode/.style={
        circle, draw = black,
        minimum size=0.9cm
    },
    shadownode/.style={
        circle, draw = black,
        fill=black!10, 
        minimum size=0.9cm
    },
}

\usepackage{url}
\usepackage{color, float}
\usepackage{xspace}
\usepackage{multirow}
\usepackage{fancyhdr}
\usepackage{microtype}
\usepackage{booktabs} 
\usepackage[ruled, vlined, linesnumbered]{algorithm2e}
\usepackage[utf8]{inputenc}
\usepackage{microtype}
\usepackage{graphicx}
\usepackage{booktabs} 
\usepackage{url}            
\usepackage{nicefrac}       
\usepackage{xspace}
\usepackage{enumerate}
\usepackage{amsthm, amsmath, mathtools, amsfonts, amssymb, dsfont, enumitem}
\usepackage{bm}
\renewcommand{\cite}{\citep}

\usepackage[mathscr]{euscript}
\usepackage[capitalize,noabbrev]{cleveref}
\usepackage{adjustbox}
\usepackage{footmisc}
\usepackage[skip=5pt]{caption}
\usepackage[skip=3pt]{subcaption}
\usepackage{wrapfig}
\usepackage{titlesec}
\usepackage{xcolor, soul}
\sethlcolor{lightgray}

\titlespacing{\section}{0pt}{5pt}{0pt}
\titlespacing{\subsection}{0pt}{5pt}{0pt}
\titlespacing{\subsubsection}{0pt}{5pt}{0pt}
\titlespacing{\paragraph}{0pt}{0pt}{5pt}
\graphicspath{{./figure/}}

\numberwithin{figure}{section}
\numberwithin{table}{section}

\newcommand{\ahat}{\widehat{a}}
\newcommand{\eps}{\varepsilon}
\renewcommand{\hat}{\widehat}

\newcommand{\offlinedata}{\mathscr{T}_\text{offline}}
\newcommand{\ldata}{\mathscr{T}_\text{labelled}}
\newcommand{\unlabeldata}{\mathscr{T}_\text{unlabelled}}
\newcommand{\proxyactiondata}{\mathscr{T}_\text{proxy}}

\newcommand{\Plabel}{\mathcal{P}_\text{labelled}}
\newcommand{\Punlabel}{\mathcal{P}_\text{unlabelled}}

\newcommand{\dt}{\texttt{DT}\xspace}
\newcommand{\cql}{\texttt{CQL}\xspace}
\newcommand{\td}{\texttt{TD3BC}\xspace}

\newcommand{\ssdt}{\texttt{SS-DT}\xspace}
\newcommand{\sscql}{\texttt{SS-CQL}\xspace}
\newcommand{\sstd}{\texttt{SS-TD3BC}\xspace}

\newcommand{\hopper}{\texttt{hopper}\xspace}
\newcommand{\walker}{\texttt{walker}\xspace}
\newcommand{\cheetah}{\texttt{halfcheetah}\xspace}
\newcommand{\maze}{\texttt{maze2d}\xspace}
\newcommand{\gym}{\texttt{Gym}\xspace}

\newcommand{\medexpert}{\texttt{medium-expert}\xspace}
\newcommand{\medreplay}{\texttt{medium-replay}\xspace}
\newcommand{\medium}{\texttt{medium}\xspace}
\newcommand{\med}{\texttt{medium}\xspace}

\newcommand{\abs}[1]{\left|#1\right|}
\newcommand{\set}[1]{\left\{#1\right\}}

\newcommand{\vs}{\mathbf{s}}

\newcommand{\D}{\mathcal{D}}
\newcommand{\Dlabel}{\mathcal{D}_\text{labelled}}
\newcommand{\Dunlabel}{\mathcal{D}_\text{unlabelled}}

\newcommand{\N}{\mathcal{N}}
\newcommand{\A}{\mathcal{A}}

\renewcommand{\P}{\operatorname{\mathbb{P}}}
\newcommand{\E}{\operatorname{\mathbb{E}}}

\renewcommand{\S}{\mathcal{S}}
\renewcommand{\emptyset}{\varnothing}
\newcommand{\gato}{\texttt{DT-Joint}\xspace}
\newcommand{\ssa}{\texttt{SS-ORL}\xspace}
\newcommand{\norm}[1]{\|#1\|}

\DeclareMathOperator*{\argmin}{argmin}

\theoremstyle{plain}

\theoremstyle{definition}

\theoremstyle{remark}

\usepackage[textsize=tiny]{todonotes}

\icmltitlerunning{Semi-Supervised Offline Reinforcement Learning with Action-Free Trajectories}

\begin{document}

\twocolumn[
\icmltitle{Semi-Supervised Offline Reinforcement Learning with Action-Free Trajectories}



\icmlsetsymbol{equal}{*}

\begin{icmlauthorlist}
\icmlauthor{Qinqing Zheng}{fair}
\icmlauthor{Mikael Henaff}{fair}
\icmlauthor{Brandon Amos}{fair}
\icmlauthor{Aditya Grover}{ucla}
\end{icmlauthorlist}

\icmlaffiliation{fair}{Meta AI Research}
\icmlaffiliation{ucla}{UCLA}

\icmlcorrespondingauthor{Qinqing Zheng}{zhengqinqing@gmail.com}

\icmlkeywords{Machine Learning, ICML}

\vskip 0.3in
]



\printAffiliationsAndNotice{}  

\begin{abstract}
Natural agents can effectively learn from multiple data sources that differ in size, quality, and types of measurements. We study this heterogeneity in the context of offline reinforcement learning (RL) by introducing a new, practically motivated semi-supervised setting. Here, an agent has access to two sets of trajectories: labelled trajectories containing state, action and reward triplets at every timestep, along with unlabelled trajectories that contain only state and reward information.  For this setting, we develop and study a simple meta-algorithmic pipeline that learns an inverse dynamics model on the labelled data to obtain proxy-labels for the unlabelled data, followed by the use of any offline RL algorithm on the true and proxy-labelled trajectories. Empirically, we find this simple pipeline to be highly successful --- on several D4RL benchmarks~\cite{fu2020d4rl}, certain offline RL algorithms can match the performance of variants trained on a fully labelled dataset even when we label only 10\% of trajectories which are highly suboptimal. To strengthen our understanding, we perform a large-scale controlled empirical study investigating the interplay of data-centric properties of the labelled and unlabelled datasets, with algorithmic design choices (e.g., choice of inverse dynamics, offline RL algorithm) to identify general trends and best practices for training RL agents on semi-supervised offline datasets. \looseness=-1
\end{abstract}
\section{Introduction}
\label{sec:intro}
One of the key challenges with deploying reinforcement learning (RL) agents is their prohibitive sample complexity for real-world applications.
Offline reinforcement learning (RL) can significantly reduce the sample complexity by exploiting logged demonstrations from auxiliary data sources~\cite{levine2020offline}.
Standard offline RL assumes fully logged datasets: the trajectories are complete sequences of observations, actions, and rewards.
However, contrary to curated benchmarks in use today, the nature of offline demonstrations in the real world can be highly varied.
For example, the demonstrations could be misaligned due to frequency mismatch~\cite{burns2022offline}, use different sensors, actuators, or dynamics~\cite{reed2022generalist, lee2022multi}, or lack partial state~
\cite{ghosh2022offline, rafailov2021offline, mazoure2021improving} or reward information~\cite{yu2022leverage}.
Successful offline RL in the real world requires embracing these heterogeneous aspects for maximal data efficiency, similar to learning in humans.

In this work, we propose a new and practically motivated \emph{semi-supervised} setup for offline RL: the offline dataset consists of some
action-free trajectories (which we call \emph{unlabelled}) in addition to the standard action-complete trajectories (which we call \emph{labelled}).
In particular, we are mainly interested in the case where a significant majority of the trajectories in the offline dataset are unlabelled, and the unlabelled data might have different qualities than the labelled ones. One motivating example for this setup is learning from videos 
~\cite{schmeckpeper2020reinforcement, schmeckpeper2020learning}
or third-person demonstrations~\cite{thirdpersonIL, sharma19thirdperson}.
There are tremendous amounts of internet videos that can be potentially used to train RL agents, yet they are 
without action labels and are of varying quality.
Notably, our setup has two key properties that differentiate it from traditional semi-supervised learning:
\vspace{-\medskipamount}
\begin{itemize}[leftmargin=*]\itemsep0em
    \item  First, we do not assume that the distribution of the labelled and unlabelled trajectories are necessarily identical. 
        In realistic scenarios, we expect these to be different with unlabelled data having higher returns than labelled data e.g., videos of a human professional are easy to obtain whereas precisely measuring their actions is challenging.
        We replicate such varied data quality setups in some of our experiments; Figure~\ref{fig:density_semi_example} shows an illustration of the difference in returns between the labelled and unlabelled dataset splits using the \hopper-\medexpert D4RL dataset.
        \begin{figure}[h]
        \centering
        \includegraphics[width=0.8\columnwidth]{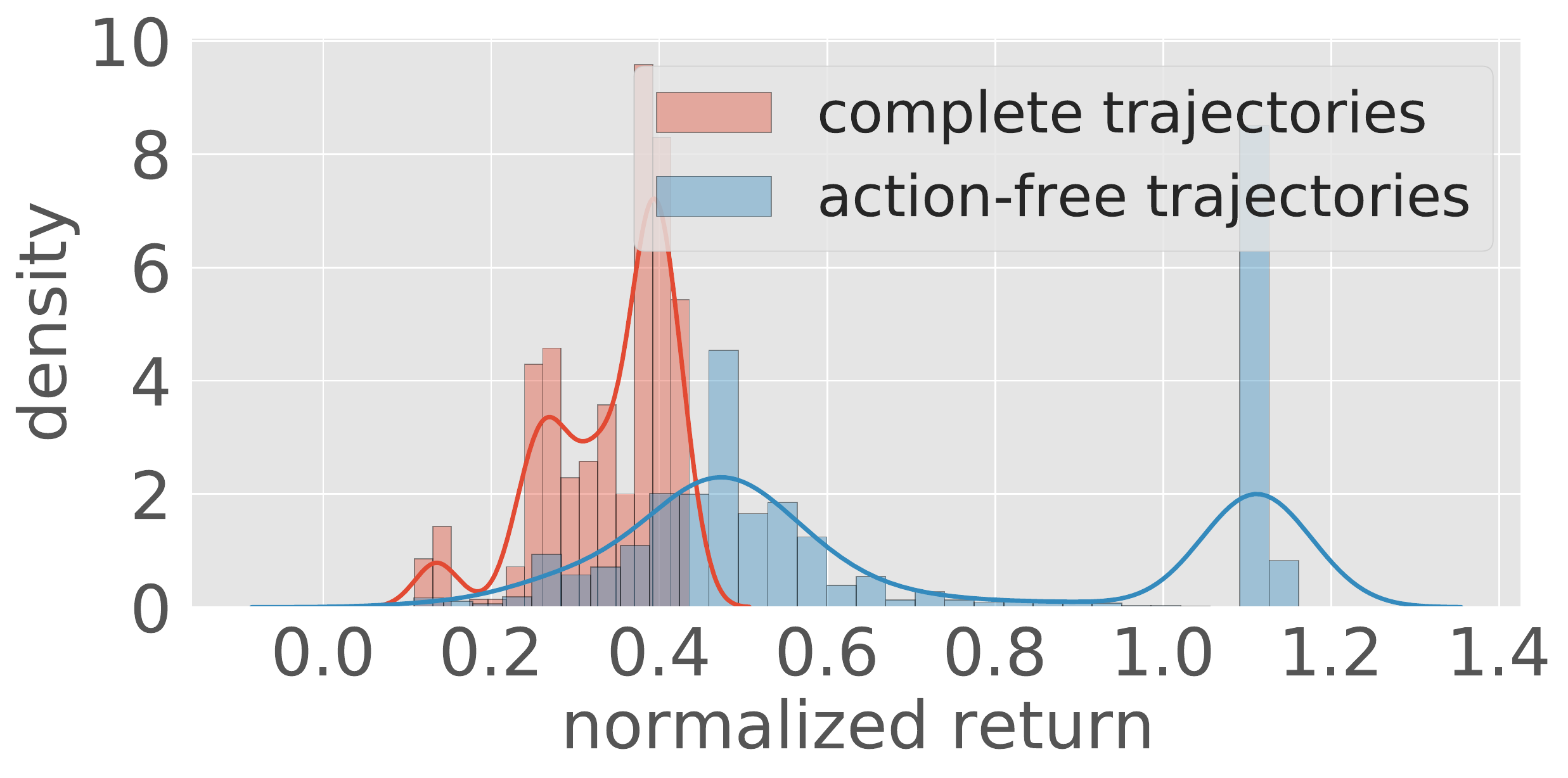}
        \caption{An example of the return distribution of the labelled and unlabelled datasets.}
        \label{fig:density_semi_example}
        \end{figure}
    \item Second, our end goal goes beyond labelling the actions in the unlabelled trajectories, but rather we intend to use the unlabelled data for learning a downstream policy that is better than the behavioral policies used for generating the offline datasets. \looseness=-1
\end{itemize}
\vspace{-1em}
Correspondingly, there are two kinds of generalization challenges in the proposed setup: (i) generalizing from the labelled to the unlabelled data distribution and then (ii) going beyond the offline data distributions to get closer to the expert distribution. Regular offline RL is only concerned with the latter, and standard algorithms such as Conservative Q Learning~(\cql; \citet{kumar2020conservative}), TD3BC~(\td; \citet{fujimoto2021minimalist}) or Decision Transformer~(\dt; \citet{chen2021decision}), cannot directly operate on such unlabelled trajectories.
At the same time, na\"ively throwing out the unlabelled trajectories can be wasteful, especially when they have high returns.
Thus, our paper seeks to answer the following question:

\emph{\hspace*{2em}How can we best leverage the unlabelled data to improve the performance of offline RL algorithms?}

To answer this question, we study different approaches to train policies in the semi-supervised setup described above, and propose a meta-algorithmic pipeline \emph{\textbf{S}emi-\textbf{S}upervised \textbf{O}ffline
\textbf{R}einforcement \textbf{L}earning (\ssa)}.
\ssa contains three simple steps: (1) train an inverse dynamics model (IDM) on the labelled data, which predicts actions based on transition sequences, (2) fill in proxy-actions for the unlabelled data, and finally (3) train an offline RL agent on the combined dataset. \looseness=-1

The \textbf{\emph{main takeaway}} of our paper is:

    \emph{\hspace*{2em}Given low-quality labelled data, \ssa agents can exploit unlabelled data containing high-quality trajectories to improve performance. The absolute performance of \ssa is close to or even matches that of the oracle agents, which have access to complete action information of both labelled and unlabelled trajectories.}

From a technical standpoint, we address the limitations of the classic IDM~\cite{pathak2017curiosity} by proposing a novel stochastic multi-transition IDM that incorporates previous states to account for non-Markovian behavior policies. 
To enable compute and data efficient learning,
we conduct thorough ablation studies to understand how the performance of \ssa agents are affected by the algorithmic design choices,
and how it varies as a function of data-centric properties such as the size and return distributions of labelled and unlabelled datasets.
We highlight a few predominant trends from our experimental findings below:
\vspace{-1em}
\begin{enumerate}[leftmargin=*]\itemsep0em

    \item Proxy-labelling is an effective way to utilize unlabelled data.
    For example, \ssa instantiated with \dt as the offline RL method significantly outperforms an alternative DT-based approach without proxy-labelling. \looseness=-1
    
    \item Simply training the IDM on the labelled dataset outperforms more sophisticated semi-supervised protocols such as self-training~\cite{fralick1967learning}.

    \item Incorporating past information into the IDM improves generalization.
    \looseness=-1
    
    \item  The performance of \ssa agents critically depend on factors such as size and quality of the labelled and unlabelled datasets, but the effect magnitudes depend on the offline RL method.  For example, we found that \td is less sensitive to missing actions then \dt and \cql.
\end{enumerate}
\section{Related Work}
\label{sec:related}
\paragraph{Offline RL} 
The goal of offline RL is to learn effective policies from fixed datasets which are generated by unknown behavior policies. There are two main categories of model-free offline RL methods: value-based methods
and behavior cloning (BC) based methods. 

Value-based methods attempt to learn value functions based on temporal difference (TD) updates.
There is a line of work that aims to port existing off-policy value-based online RL methods to the offline setting, with various types of additional regularization components that encourage the learned policy to stay close to the behavior policy. Several representative techniques include specifically tailored policy parameterizations~\cite{fujimoto2019off,ghasemipour2021emaq}, divergence-based regularization on the learned policy~\cite{wu2019behavior, jaques2019way, kumar2019stabilizing}, and regularized value function estimation~\cite{nachum2019algaedice, kumar2020conservative, kostrikov2021offline_divergence, fujimoto2021minimalist, kostrikov2021offline}.

A growing body of recent work formulates offline RL as a supervised learning problem~\cite{chen2021decision, janner2021offline, emmons2021rvs}. Compared with value-based methods,
these supervised methods enjoy several appealing properties including algorithmic simplicity and training stability.
Generally speaking, these approaches can be viewed as conditional behavior cloning methods~\cite{bain1995framework}, where the conditioning is based on goals or returns. 
Similar to value-based methods, these can be extended to the online setup as well~\cite{zheng2022online} and demonstrate excellent performance in hybrid setups involving both offline data and online interactions.

\paragraph{Semi-Supervised Learning}
Semi-supervised learning (SSL) is a sub-area of machine learning that studies approaches to train predictors from a small amount of labelled data combined with a large amount of unlabelled data. 
In supervised learning, predictors only learn from labelled data. However, labelled training examples often require human annotation efforts and are thus hard to obtain, whereas unlabelled data can be comparatively easy to collect. 
The research on semi-supervised learning spans several decades.
One of the oldest SSL techniques, \emph{self-training}, was originally proposed in the 1960s~\cite{fralick1967learning}.
There, the predictor is first trained on the labelled data. Then, at each training round, according to certain selection criteria such as model uncertainty, a portion of the unlabelled data is annotated by the predictor and added into the training set for the next round. Such process is repeated multiple times.
We refer the readers to~\citet{zhu05survey, chapelle2006semi, ouali2020overview, van2020survey} for comprehensive literature surveys.

\paragraph{Imitation Learning from Observations}
There have been several works in imitation learning (IL) which do not assume access to the full set of actions, such as BCO~\citep{BCO}, MoBILE~\cite{kidambi2021mobile}, GAIfO~\citep{gailfo} or third-person IL approaches \cite{thirdpersonIL, sharma19thirdperson}. The recent work of~\citet{VPT} also considered a setup where a small number of labelled actions are available in addition to a large unlabelled dataset. A key difference with our work is that the IL setup typically assumes that all trajectories are generated by an expert, unlike our offline setup.
Further, some of these methods even permit reward-free interactions with the environment which is not possible in the offline setup.
\looseness=-1

\paragraph{Learning from Videos} Several works consider training agents
with human video demonstrations~\cite{schmeckpeper2020reinforcement, schmeckpeper2020learning}, which are without action annotations. Distinct from our setup, some of these works allow for online interactions, assume expert videos, and more broadly, video data typically specifies agents with different embodiments. 
\section{Semi-Supervised Offline RL}
\label{sec:algo}

\paragraph{Preliminaries}
\label{sec:prelim}
We model our environment as a Markov decision process (MDP)~\citep{bellman1957mdp} denoted by $\langle \S, \A, p, P, R, \gamma \rangle$, where $\S$ is the state space, $\A$ is the action space, $p(s_1)$ is the distribution of the initial state, $P(s_{t+1}|s_t, a_t)$ is the transition probability distribution, $R(s_t, a_t)$ is the deterministic reward function, and $\gamma$ is the discount factor. At each timestep $t$, the agent observes a state $s_t \in \S$ and executes an action $a_t \in \A$.
The environment then moves the agent to the next state $s_{t+1} \sim P(\cdot| s_t, a_t)$, and also returns the agent a reward $r_t = R(s_{t}, a_t)$.
\looseness=-1

\subsection{Proposed Setup}
\label{sec:algo_setup}
We assume the agent has access to a static offline dataset $\offlinedata$. The dataset consists of trajectories
collected by unknown policies, which are generally suboptimal. Let $\tau$ denote a trajectory and $|\tau|$ denote its length.  We assume that all the trajectories in $\offlinedata$ contain complete rewards and states. However, only a small subset of them contain actions. 
\looseness=-1

We are interested in learning a policy by leveraging the offline dataset without interacting with the environment. 
This setup is analogous to semi-supervised learning, where actions serve the role of \emph{labels}. Hence, we also refer to the complete trajectories as \emph{labelled} data (denoted by $\ldata$) and the action-free trajectories as \emph{unlabelled} data (denoted by $\unlabeldata$).
Further, we assume the labelled and unlabelled data are sampled from two distributions $\Plabel$ and $\Punlabel$, respectively. In general, the two distributions can be different. One case we are particularly interested in is when $\Plabel$ generates low-to-moderate quality trajectories, whereas $\Punlabel$ generates trajectories of diverse qualities including ones with high returns, as shown in Fig~\ref{fig:density_semi_example}.

Our setup shares some similarities with state-only imitation learning~\cite{ijspeert2002movement, bentivegna2002humanoid, torabi2019recent} in the use of action-unlabelled trajectories. However, there are two fundamental differences. 
   First,  in state-only IL, the unlabelled demonstrations are from the same distribution as the labelled demonstrations, and both are generated by a near-optimal expert policy. In our setting, $\Plabel$ and $\Punlabel$ can be different and are not assumed to be optimal.
    Second, many state-only imitation learning algorithms (e.g., ~\citet{gupta2017learning, BCO, gailfo, liu2018imitation, sermanet2018time}) permit (reward-free) interactions with the environments similar to their original counterparts (e.g., \citet{ho2016generative, kim2020domain}). This is not allowed in our offline setup, where the agents are only provided with $\ldata$ and $\unlabeldata$.


\subsection{Training Pipeline} \label{sec:algo_pipeline}
RL policies trained on low to moderate quality offline trajectories are often sub-optimal,
as many of the trajectories might not have high returns and only cover a limited part of the state space. 
Our goal is to find a way to combine the action labelled trajectories and the unlabelled action-free trajectories,
so that the offline agent can exploit structures in the unlabelled data to improve performance.

One natural strategy is to fill in \textit{proxy actions} for those unlabelled trajectories, and use the proxy-labelled data together with the labelled data as a whole to train an offline RL agent. Since we assume both the labelled and unlabelled trajectories contain the states,
we can train an inverse dynamics model (IDM) $\phi$ that predicts actions using the states. 
Once we obtain the IDM, we use it to generate the proxy actions for the unlabelled trajectories.
Finally, we combine those proxy-labelled trajectories with the labelled trajectories, and train
an agent using the offline RL algorithm of choice. Our meta-algorithmic pipeline is summarized in Algorithm~\ref{algo:semi}. 

\begin{algorithm}[th]
\DontPrintSemicolon
\small
\caption{Semi-supervised offline RL (\ssa)}\label{algo:semi}
\textbf{Input:} trajectories $\ldata$ and $\unlabeldata$, IDM transition size $k$, offline RL algorithm \texttt{ORL} \;
\tcp{train a stochastic multi-transition IDM using the labelled data}
$\hat{\theta} \leftarrow \argmin_{\theta} \sum_{(a_t, \vs_{t, -k})\; \text{in}\; \ldata}\left[ -\log \phi_\theta(a_t | \vs_{t, -k})  \right]$\;
\vskip5pt
\tcp{fill in the proxy actions for the unlabelled data}
$\proxyactiondata \leftarrow \emptyset$\; 
\For{each trajectory $\tau \in \unlabeldata$ }{
    $\ahat_{t} \leftarrow \mu_{\hat{\theta}}(\vs_{t, -k})$, i.e. mean of $\N \left(\mu_{\hat{\theta}}(\vs_{t, -k}), \, \Sigma_{\hat{\theta}}(\vs_{t, -k}) \right)$, $t = 1, \ldots, |\tau|$\\
    $\tau_\text{proxy} \leftarrow \tau$ with proxy actions $\set{\ahat_t}_{t=1}^{|\tau|}$ filled in\\
    $\proxyactiondata \leftarrow \proxyactiondata \bigcup \set{\tau_\text{proxy}}$
}
\vskip5pt
\tcp{train an offline RL agent using the combined data}
$\pi \leftarrow$ policy trained by $\texttt{ORL}$ using dataset $\ldata \bigcup \proxyactiondata$\;
\textbf{Output: $\pi$}\;
\end{algorithm}

Particularly, we propose a novel stochastic multi-transition IDM that incorporates past information
to enhance the treatment for stochastic MDPs and non-Markovian behavior policies. Section~\ref{sec:algo_idm} discusses the details.

Of note, \ssa is a \emph{multi-stage} pipeline, where the IDM is trained only on the labelled data in a \emph{single} round. 
There are other possible ways to combine the labelled and unlabelled data. 
In Section~\ref{sec:algo_alternative_design}, we discuss several alternative design choices and the key reasons why we do not employ them. Additionally, we present the ablation experiments in Section~\ref{sec:expr_design}. 

\subsubsection{Stochastic Multi-transition IDM} 
\label{sec:algo_idm}
In past work~\cite{pathak2017curiosity, pathak18largescale, henaff2022exploration}, the IDM typically learns to map two subsequent states of the $t$-th transition, $(s_t, s_{t+1})$, to $a_t$. 
In theory, this is sufficient when the offline dataset is generated by a single Markovian policy 
in a deterministic environment, see Appendix~\ref{app:idm_proof} for the analysis. 
However, in practice, the offline dataset might contain trajectories logged from multiple sources.

To provide better treatment for multiple behavior policies,
we introduce a multi-transition IDM that predicts 
the distribution of $a_t$ using the most recent $k+1$ transitions.
More precisely, let $\vs_{t, -k}$ denote the sequence $s_{\min(0, t-k)}, \ldots, s_t, s_{t+1}$. We model $\P(a_t|\vs_{t, -k})$ as a multivariate Gaussian with a diagonal covariance matrix:
\begin{equation}
\label{eq:idm}
    a_t \sim \N\big(\mu_\theta(\vs_{t, -k}), \, \Sigma_\theta(\vs_{t, -k})\big).
\end{equation}
Let $\phi_\theta(a_t | \vs_{t, -k})$ be the probability density function
of $\N\big(\mu_\theta(\vs_{t, -k}), \, \Sigma_\theta(\vs_{t, -k})\big)$.
Given the labelled trajectories $\ldata$, we minimize the negative log-likelihood loss $\sum_{(a_t, \vs_{t, -k}) \; \text{in} \; \ldata}\left[ -\log \phi_\theta(a_t | \vs_{t, -k})  \right]$.
We call $k$ the transition size parameter.
Note that the standard IDM which predicts $a_t$ from $(s_t, s_{t+1})$ under the $\ell_2$ loss, is a special case subsumed by our model: it is equivalent to the case $k=0$ and the diagonal entries of $\Sigma_\theta$ (i.e., the variances of each action dimension) are all the same.

In essence, we approximate $p(a_t|s_{t+1}, \ldots, s_1)$ by $p(a_t|\vs_{t, -k})$, and choosing $k>0$ allows us to take
past state information into account. Meanwhile, the theory also indicates that incorporating future states like $s_{t+2}$ would not help 
to predict $a_t$ (see the analysis in Appendix~\ref{app:idm_proof} for details).
For all the experiments in this paper, we use $k=1$. We ablate this design choice in Section~\ref{sec:expr_design}.
Moreover, our IDM naturally extends to non-Markovian policies and stochastic MDPs. This is beyond the scope of
this paper, but we consider them as potential directions for future work.

\subsubsection{Alternative Design Choices}
\label{sec:algo_alternative_design}
\paragraph{Training without Proxy Labelling} 
\ssa fills in proxy actions for the unlabelled trajectories before training the agent. There, 
the policy learning task is defined on the combined dataset of the labelled and unlabelled data.
An alternative approach is to only use the labelled data to define the policy learning task, but create
certain auxiliary tasks using the unlabelled data. These auxiliary tasks do not depend on actions, so
that proxy-labelling is not needed. Multitask learning approaches can be employed to train an agent that solves those tasks together. For example, \citet{reed2022generalist} train a generalist agent that processes diverse sequences with a single transformer model.
In a similar vein, we consider \gato, a variant of \dt, that trains on 
both labelled and unlabelled data simultaneously. In a nutshell, \gato predicts actions for the labelled trajectories, and states and rewards for both labelled and unlabelled trajectories. See Appendix~\ref{app:gato} for the implementation details. Nonetheless, our ablation experiment in Section~\ref{sec:expr_design} shows that \ssa significantly outperforms \gato. 
\looseness=-1

\paragraph{Self-Training for the IDM} The annotation process in \ssa, which involves training an IDM on the labelled data and generating proxy actions for the unlabelled trajectories, is similar to one step of \emph{self-training}~\cite[Cf. Section~\ref{sec:related}]{fralick1967learning}, one commonly used approach in standard semi-supervised learning. 
However, a key difference is that we do not retrain the IDM but directly move to the next stage of training the agent using the combined data. There are a few reasons that we do not employ self-training for the IDM. First, it is computationally expensive to execute multiple rounds of training. More importantly, our end goal is to obtain a downstream policy with improved performance via utilizing the proxy-labelled data.
As a baseline, we consider self-training for the IDM, where after each training round
we add the proxy-labelled data with low predictive uncertainties into the training set for the next round.
Empirically, we found that this variant underperforms our approach. See Section~\ref{sec:expr_design}
and Appendix~\ref{app:self_training} for more details. 

\section{Experiments}
\label{sec:expr}
Our main objectives are to answer four sets of questions:\looseness=-1
\vspace*{-1em}
\begin{enumerate}[leftmargin=*]\itemsep0em
    \item[Q1.] How close can \ssa agents match the performance of fully supervised offline RL agents, especially when only a small subset of trajectories is labelled?
    \item[Q2.] How do the \ssa agents perform under different design choices for training the IDM, or even avoiding proxy-labelling completely? 
    \item[Q3.] How does the performance of \ssa agents vary as a function of the size and quality of the labelled and unlabelled datasets? 
    \item[Q4.] Do different offline RL methods respond differently to various setups of the dataset size and quality?
\end{enumerate}
\vspace*{-0.5em}

We focus on two \gym locomotion tasks, \hopper and \walker, with the v2 \medexpert, \medium and \medreplay datasets
from the D4RL benchmark \citep{fu2020d4rl}. 
Due to space constraints, the results on \medium and \medreplay datasets are deferred to Appendix~\ref{app:main_extra_coupled}. We respond to the above questions in Section~\ref{sec:expr_main}, ~\ref{sec:expr_design}, \ref{sec:expr_data} and~\ref{sec:expr_offline}, respectively. 
We also include additional experiments on the \texttt{maze2d} environments in Appendix~\ref{app:additional_expr_maze2d}.
For all experiments, we train $5$ instances of each method with different seeds, and for each instance we roll out $30$ evaluation trajectories.  
Our code is available at \url{https://github.com/facebookresearch/ssorl/}.
\looseness=-1

\subsection{Main Evaluation (Q1)}
\label{sec:expr_main}

\paragraph{Data Setup} We subsample $10\%$ of the total offline trajectories
whose returns are from the bottom $q\%$ as the labelled trajectories, $10 \leq q \leq 100$.
The actions of the remaining trajectories are discarded to create the unlabelled ones.
We refer to this setup as the \emph{coupled} setup, since the labelled data distribution $\Plabel$ and the unlabelled data distribution $\Punlabel$ will change simultaneously as we vary the value of $q$. 
As $q$ increases, the labelled data quality increases and the distributions $\Plabel$ and $\Punlabel$ become closer. When $q=100$, our setup is equivalent to sampling the labelled trajectories uniformly and $\Plabel = \Punlabel$. Note that under our setup, we always have $10\%$ trajectories labelled and $90\%$ unlabelled, and the total amount of data used to train the offline RL agent is the same as the original offline dataset. This allows for easy comparison with results under the standard, fully labelled setup. In Section~\ref{sec:expr_data}, we will decouple $\Plabel$ and $\Punlabel$ for an in-depth understanding of their individual influences on the \ssa agents.
\looseness=-1

\paragraph{Inverse Dynamics Model} We train an IDM as described in Section~\ref{sec:algo} with $k=1$. That is, the IDM predicts $a_t$ using 3 consecutive states: $s_{t-1}, s_t$ and $s_{t+1}$, where the mean and the covariance matrix are predicted by two independent multilayer perceptrons (MLPs), each containing two hidden layers and $1024$ hidden units per layer. To prevent overfitting, we randomly sample $10\%$ of the labelled trajectories as the validation set, and use the IDM that yields the best validation error within $100$k iterations. \looseness=-1

\paragraph{Offline RL Methods} We instantiate Algorithm~\ref{algo:semi} with \dt, \cql and \td as the underlying offline RL methods. 
\dt is a recently proposed conditional behaviour cloning (BC) method that uses sequence modelling tools to model the trajectories. \cql is a representative value-based offline RL method. \td is a hybrid method which adds a BC term to regularize the Q-learning updates. We refer to these instantiations as \ssdt, \sscql and \sstd, respectively. See Appendix~\ref{app:hp} for the implementation details.  

\begin{figure}[t!]
    \centering
    \includegraphics[width=\columnwidth]{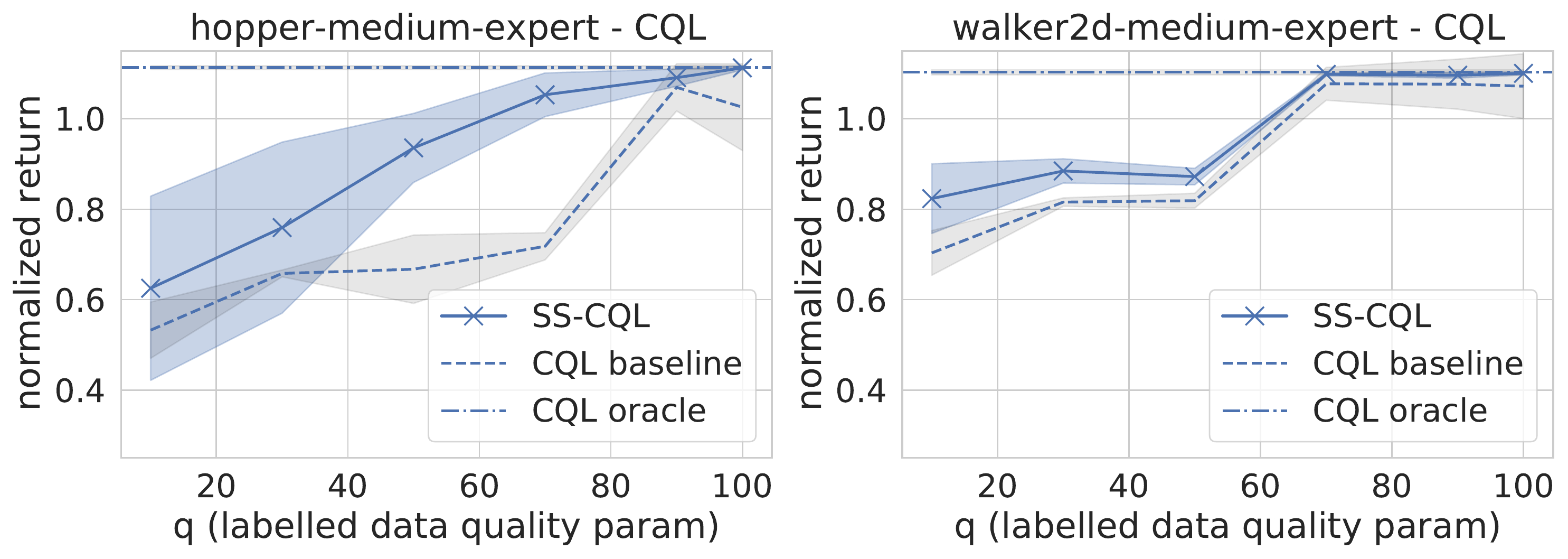}\\
    \includegraphics[width=\columnwidth]{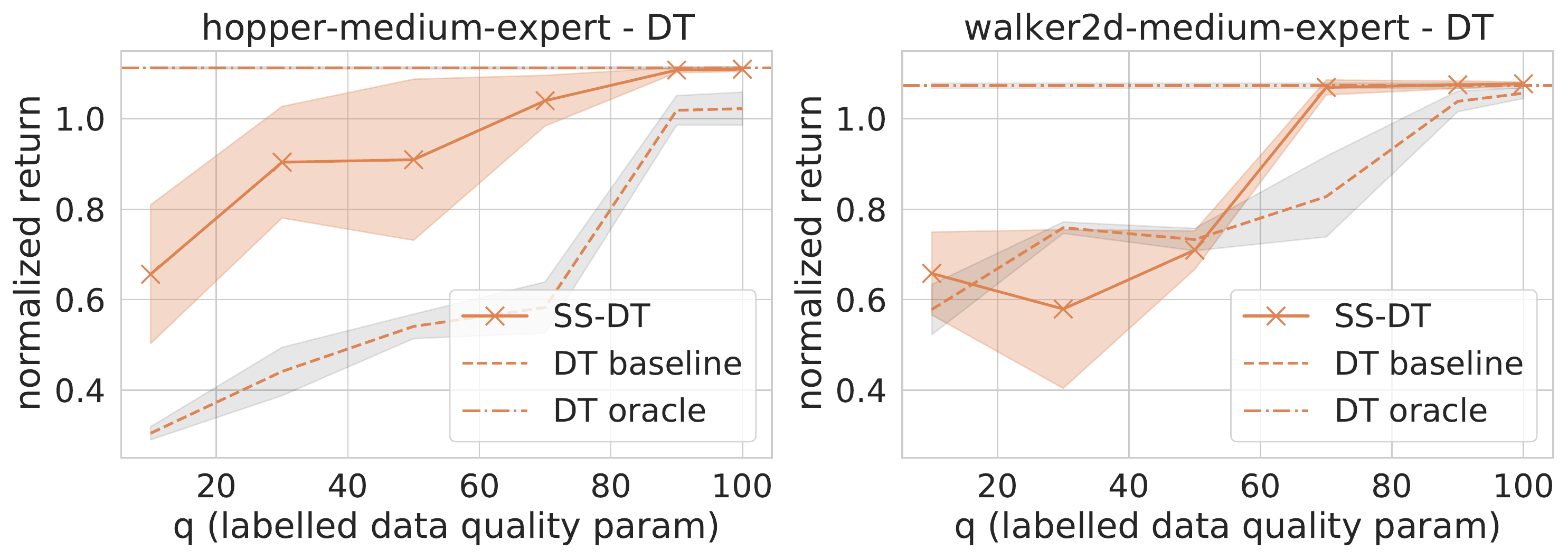}\\
    \includegraphics[width=\columnwidth]{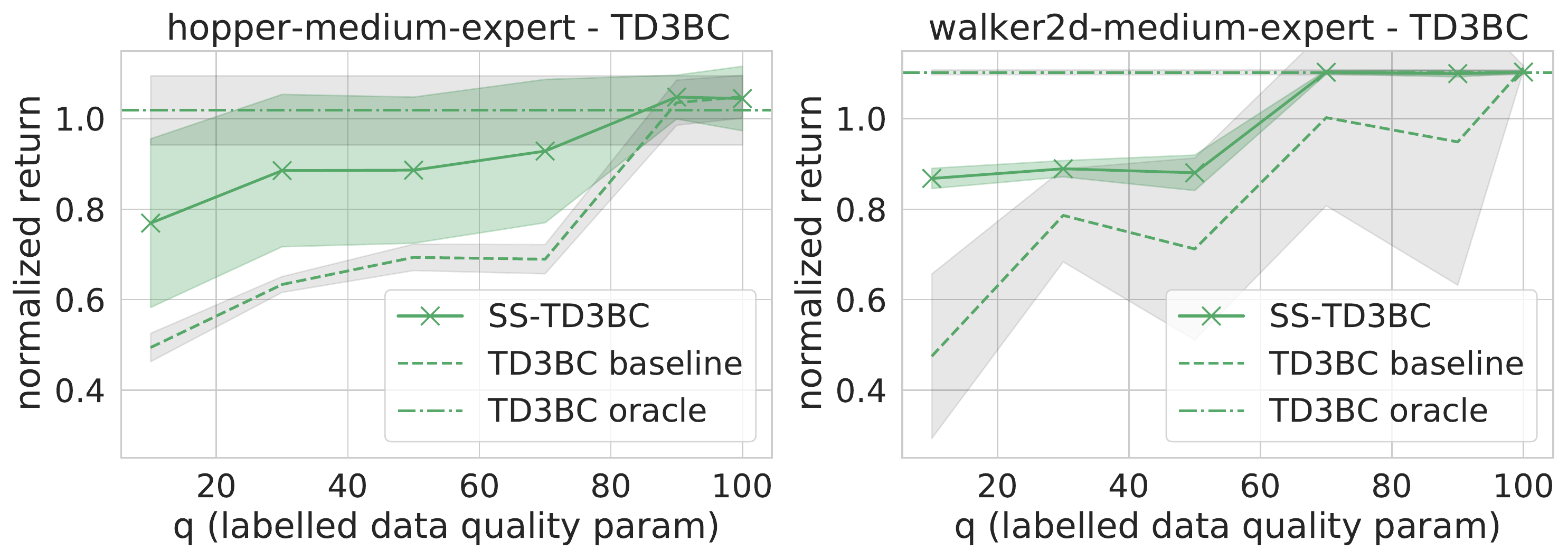}\\
    \caption{Return (average and standard deviation) of \ssa agents trained on the D4RL \medexpert datasets. The \ssa agents
    are able to utilize the unlabelled data to improve their performance upon the baselines and even match 
    the performance of the oracle agents. 
    }
    \label{fig:main_medium-expert}
    \vspace{0.5em}
    \includegraphics[width=\columnwidth]{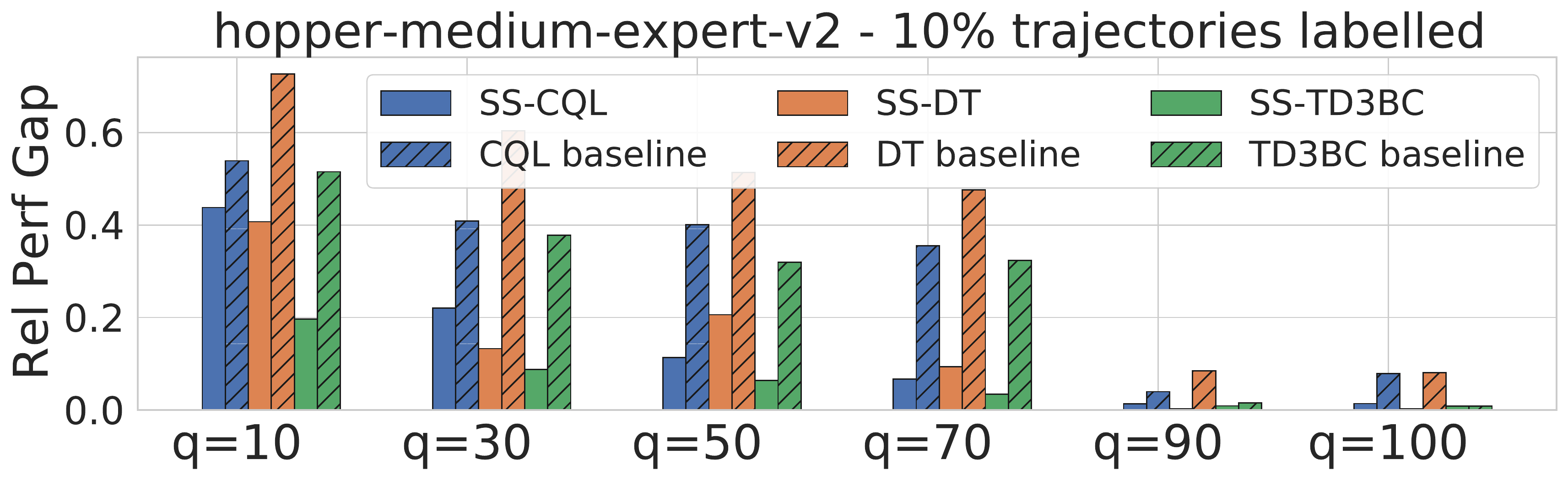}\\
    \includegraphics[width=\columnwidth]{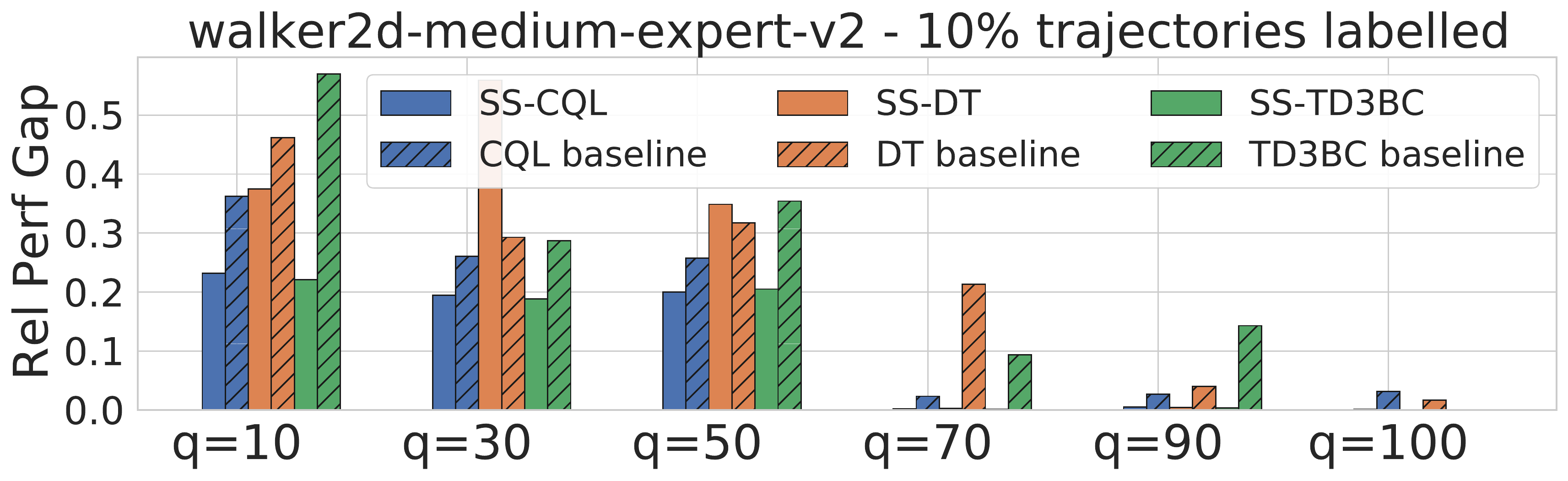}
    \caption{Relative performance gap of \ssa agents and corresponding baselines on \hopper and \walker-\medexpert datasets.}
    \label{fig:perf_gap_medium_expert}
    \vspace{0.5em}
    \includegraphics[width=\columnwidth]{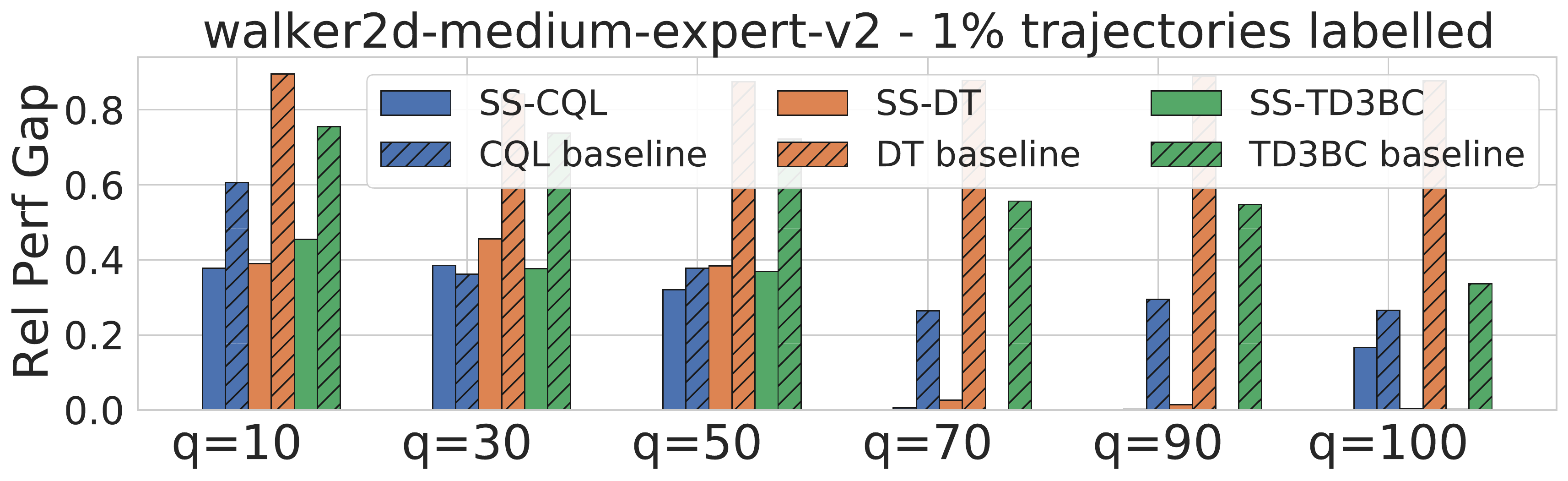}
    \caption{Relative performance gap of \ssa agents and corresponding baselines with $1\%$ labelled trajectories. }
    \label{fig:perf_gap_medium_expert_label_1}
\end{figure}
\paragraph{Results} We compare the performance of the \ssa agents with corresponding \textit{baseline} and \textit{oracle} agents. The baseline agents are trained on the labelled trajectories only, and the oracle agents are trained on the full offline dataset with complete action labels.
Intuitively, the performance of the baseline and the oracle agents can be considered as the (estimated) lower and upper bounds for the performance of the \ssa agents.
We consider $6$ different values of $q$: $10, 30, 50, 70, 90$ and $100$, and 
we report the average return and standard deviation after $200$k iterations.
Figure~\ref{fig:main_medium-expert} plots the results on the \medexpert datasets. On both datasets, the \ssa agents consistently improve upon the baselines. Remarkably, even when the labelled data quality is low, the \ssa agents are able to obtain decent returns. 
As $q$ increases, the performance of the \ssa agents also keeps increasing and finally matches the performance of the oracle agents. 

To quantitatively measure how a \ssa agent tracks the performance of the corresponding oracle agent, we define the \emph{relative performance gap} of \ssa agents as
\begin{equation}
    \label{eq:perf_gap}
    \small
    \frac{\texttt{Perf(Oracle-ORL)} - \texttt{Perf(\ssa)}}{\texttt{Perf(Oracle-ORL)}},
\end{equation}
and similarly for the baseline agents. Figure~\ref{fig:perf_gap_medium_expert} plots the average relative performance gap of these agents. Compared with the baselines, the \ssa agents notably reduce the relative performance gap.

Our results generalize to even fewer percentage of labelled data. Figure~\ref{fig:perf_gap_medium_expert_label_1} plots the relative performance gap of the agents  trained on \walker-\medexpert datasets, when only $1\%$ of the total trajectories are labelled. See Appendix~\ref{app:fewer_labelled_data} for more experiments.
Similar observations can be found in the results of \medium and \medreplay datasets, see Figure~\ref{fig:main_medium} and~\ref{fig:main_medium-replay}.
\looseness=-1
\subsection{Comparison with Alternative Design Choices (Q2)}
\label{sec:expr_design}

\begin{figure}[t]
    \centering
    \includegraphics[width=\columnwidth]{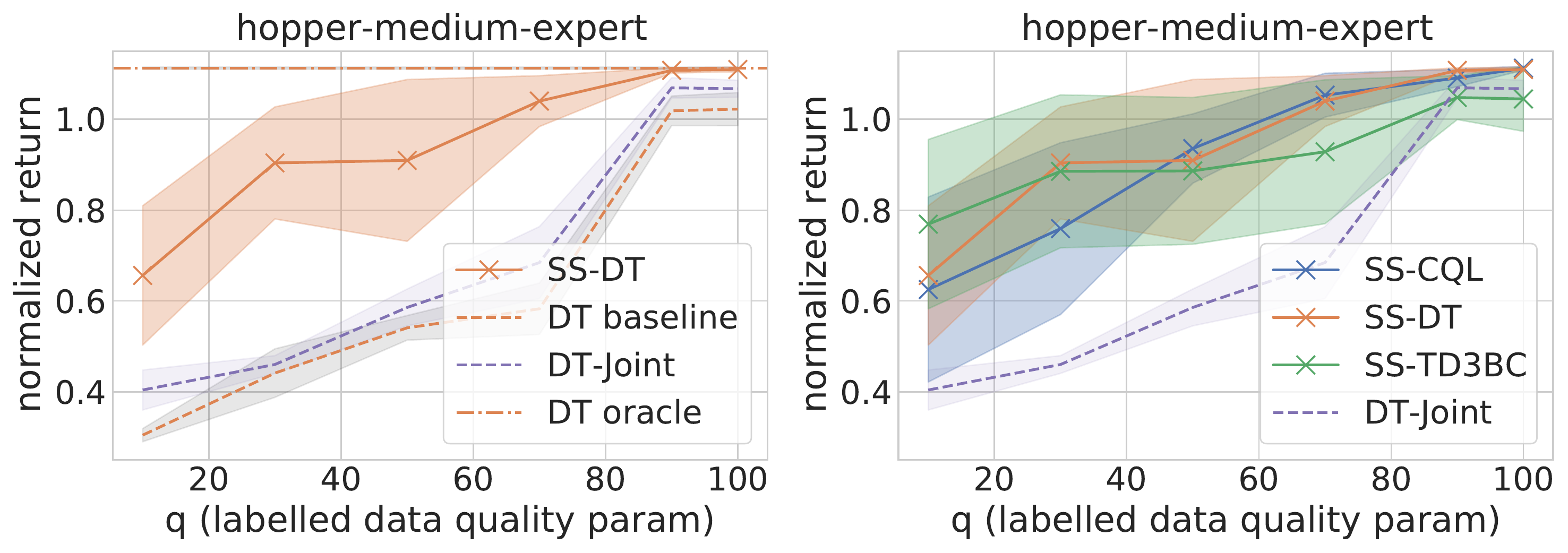}
    \caption{(L) \ssdt significantly outperforms \gato on the \hopper-\medexpert dataset. The latter only slightly improves upon the baseline. (R) \sscql and \sstd also outperform \gato. \looseness=-1
   }
    \label{fig:gato}
\end{figure}

\paragraph{Training without Proxy-Labelling}
Figure~\ref{fig:gato} plots the performance of \gato and the \ssa agents on the \hopper-\medexpert dataset, using the coupled setup as in Section~\ref{sec:expr_main}. 
Since \gato is a variant of \dt, the left panel compares \gato with \ssdt as well as the \dt baseline and the \dt oracle.
\gato only marginally outperforms the \dt baseline and performs significantly worse than \ssdt.
In addition, the right panel shows that \sscql, \ssdt and \sstd all perform much better than \gato.
The implementation details of \gato can be found in Appendix~\ref{app:gato}.

\paragraph{Self-Training for the IDM}
\label{sec:expr_idm_uncertainty}
\begin{figure}[t]
\centering
\includegraphics[width=0.9\linewidth]{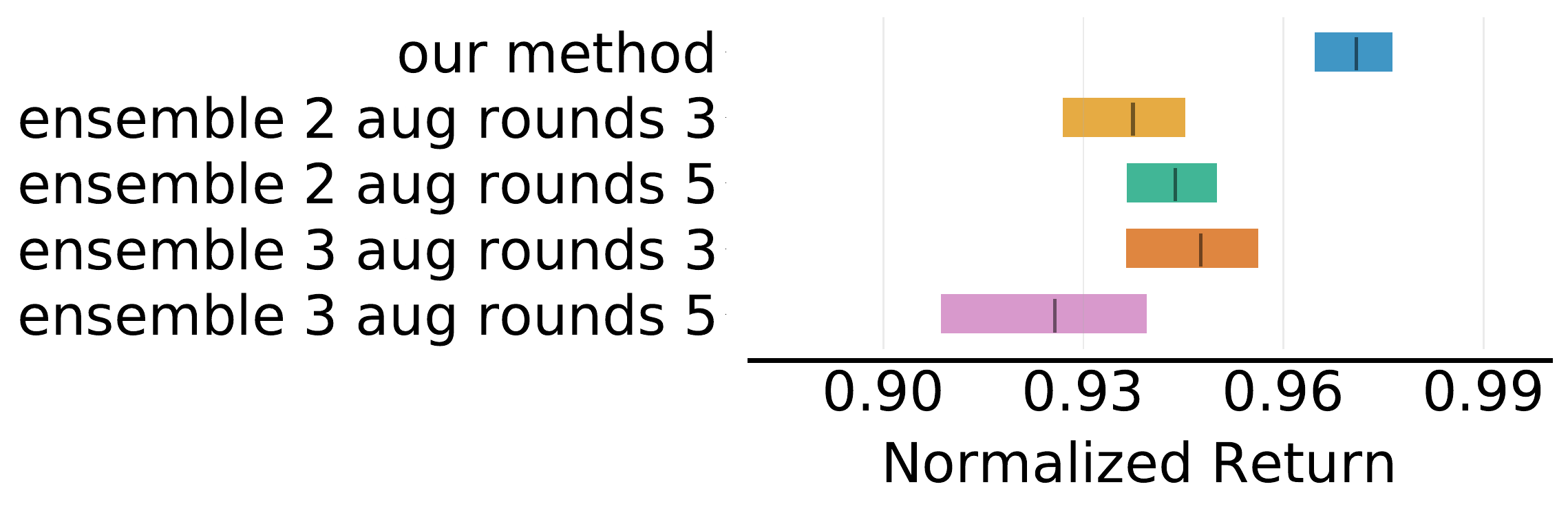}
\caption{
The $95\%$ bootstrap CIs of the IQM return obtained by 
the \ssa agents and the variants with self-training IDMs.
}
\label{fig:idm_ablation_self_training}
\vskip5pt
\includegraphics[width=\linewidth]{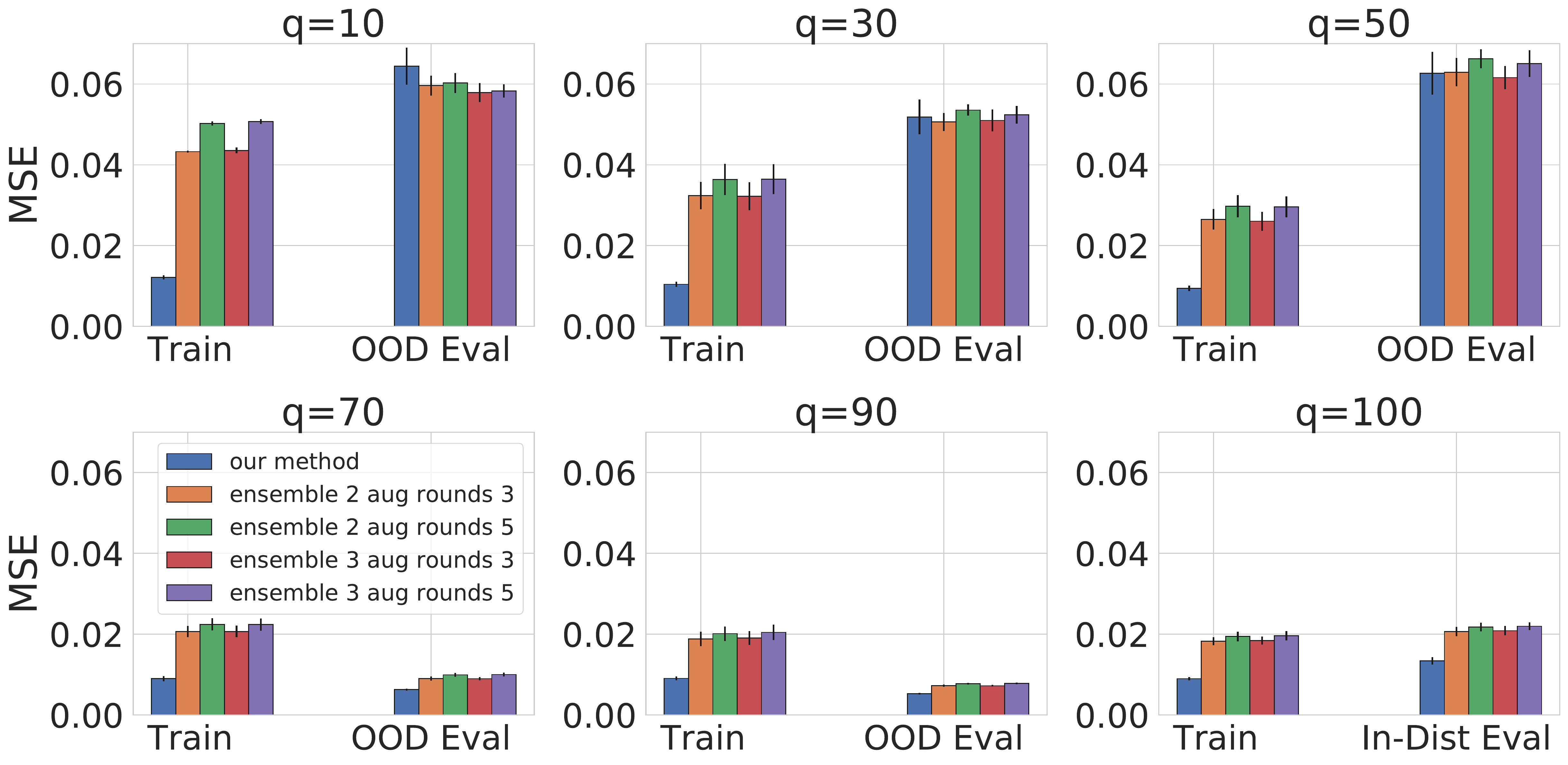}
\caption{The action prediction MSE of different IDMs.}
\label{fig:idm_mse_self_training}
\vskip5pt
\includegraphics[width=0.9\linewidth]{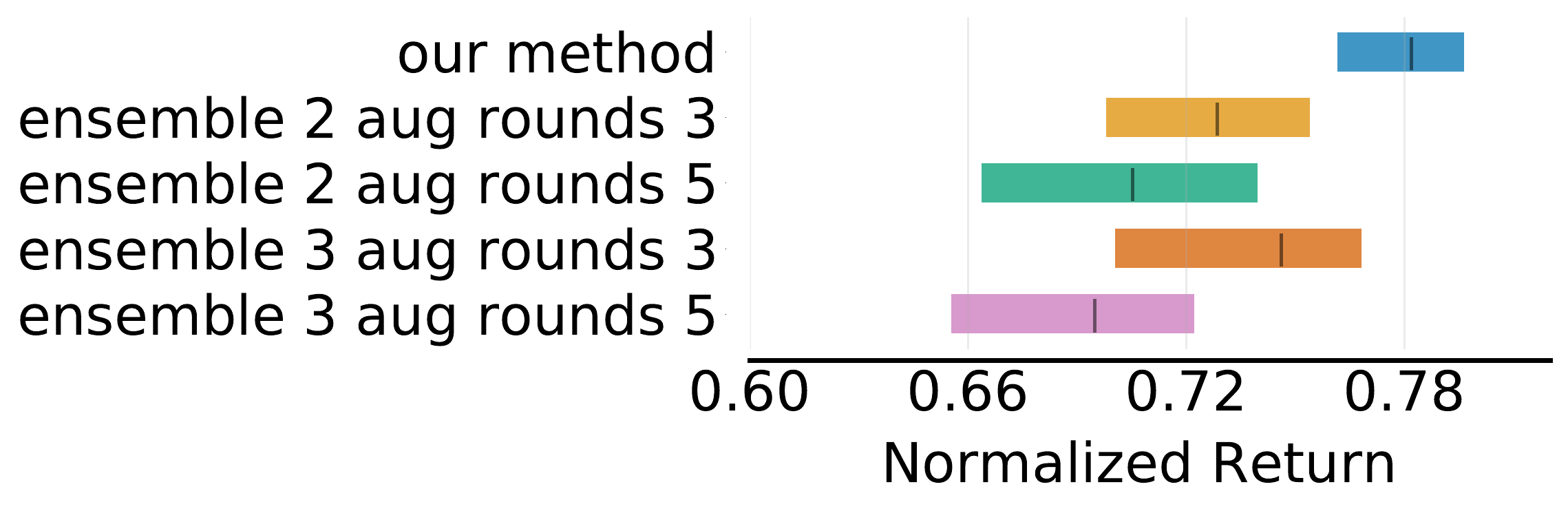}
\caption{
The $95\%$ bootstrap CIs of the IQM return,
when the labelled data is of low or moderate quality.}
\label{fig:idm_ablation_self_training_low_quality}
\end{figure}
We consider a variant of \ssa where self-training is used to train the IDM.
Recall that self-training involves an initial training round using only the labelled data, followed by multiple additional rounds using the augmented training sets.
After each training round, we need to measure the uncertainties of our action predictions and add the most ones into the training set. To do this, we use the ensemble based method~\cite{lakshminarayanan2017simple} where
we train $m$ independent stochastic IDMs. 
We model the action distribution as the mixture of those $m$ estimated distributions. The whole self-training algorithm is presented in Algorithm~\ref{algo:self_training} in Appendix~\ref{app:self_training}.

We compare \sscql, \ssdt with their self-training variant on the \walker-\medexpert datasets, using
IDM with $k=1$. All the hyperparameters and the architecture are the same.
We have tested the variant with ensemble size $2$ and $3$, and with $3$ and $5$ augmentation rounds. As before, we use the coupled setup with $6$ different $q$ varying between $10$ and $100$.
To take account of different models and different data setups, we report the $95\%$ stratified bootstrap confidence intervals (CIs)
of the interquartile mean (IQM)\footnote{
The interquartile mean of a list of sorted numbers is the mean of the middle $50\%$ numbers. } of the return for all these cases
and training instances~\cite{agarwal2021deep}. We use $50000$ bootstrap replications to generate the CIs.
Compared with the other statistics like the mean or the median, the IQM is both robust to outliers and also a good representative of the overall performance. The stratified bootstrapping is a handy tool to obtain CIs with decent coverage rate, even if one only have a small number of training instances per setup. We refer the readers to \citet{agarwal2021deep} for the complete introduction. \looseness=-1
Figure~\ref{fig:idm_ablation_self_training} plots the $95\%$ bootstrap CIs of the IQM return across all the setups.
Our approach notably outperforms the other variants.

It is intriguing to investigate the MSE of action predictions for different IDMs. Figure~\ref{fig:idm_mse_self_training} shows that our IDM is favourable when the labelled data quality is relatively high ($q=70, 90$ and $100$),
yet it is comparable with the self-training IDMs when the labelled data quality is low or moderate ($q=10, 30$ or $50$). Interestingly, we have found that
the final performance of \ssa still clearly outperforms in those cases, see Figure~\ref{fig:idm_ablation_self_training_low_quality}.

\begin{figure}[tb]
\centering
\includegraphics[width=0.85\columnwidth]{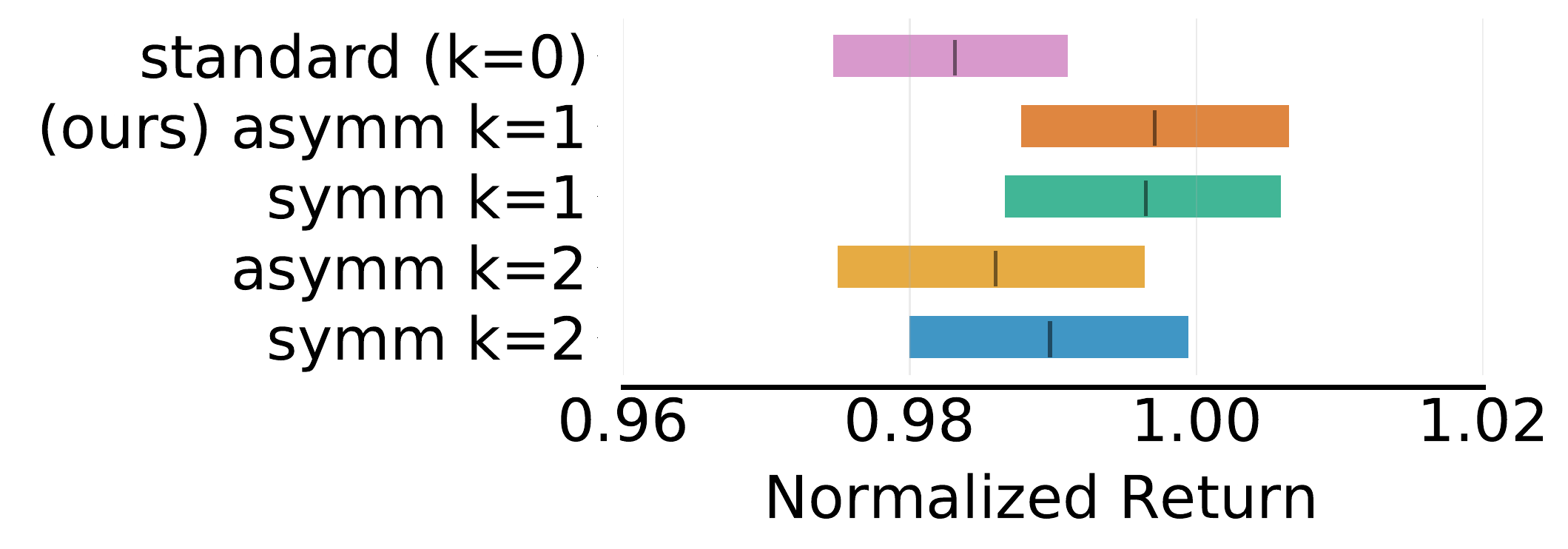}
\caption{The $95\%$ bootstrap CIs of the IQM return of the \ssa agents with different IDM architectures.}
\label{fig:idm_arch_perf_iqm}
\end{figure}
\paragraph{IDM Architecture} 
We consider the multi-transition IDM with transition window size $k=0, 1, 2$, respectively. 
To verify the influence of future states on predicting the actions, we also consider the variant that incorporates future $k$ transitions. We refer to those models \emph{symmetric} IDMs
and our IDMs \emph{asymmetric} IDMs. When $k=2$, the symmetric IDM will predict $a_t$ using the states $s_{t-2}, \ldots, s_t, s_{t+1}, \ldots, s_{t+3}$, while our asymmetric IDM will only use states up to $s_{t+1}$.
We train \sscql and \ssdt agents on the \walker-\medexpert datasets using those IDMs.
Again, we use the coupled set with $6$ different values of $q$.
Figure~\ref{fig:idm_arch_perf_iqm} plots the $95\%$ bootstrap CIs
of the IQM return across all the setups and training instances.
The symmetric IDMs perform comparably to the asymmetric IDMs, providing
empirical justifications that the future states beyond timestep $t+1$
are independent of $a_t$ given state $s_{t+1}$, see Appendix~\ref{app:idm_proof}.
The choice $k=1$ furthermore outperforms the other two options.
Since the behavior policy of the \medexpert dataset can be viewed as mixture of two policies~\cite{fu2020d4rl},
this provides empirical evidences that our IDM better copes with multiple behaviour policies than classic IDM.
Our intuition is that it might be easier to infer the actual behavior policy by a sequence of past states 
rather than a single one. 

\subsection{Albation Study for Data-Centric Properties (Q3)}
\label{sec:expr_data}
 We conduct experiments to investigate the performance of \ssa in variety of
 data settings. To enable a systematic study, we depart from the coupled setup in Section~\ref{sec:expr_main} and consider a decoupling of $\Plabel$ and $\Punlabel$. 
We will vary four configurable values: the quality and size of both the labelled and unlabelled trajectories,
individually while keeping the other values fixed. We
examine how the performance of the \ssa agents change with these variations.\looseness=-1
 \begin{figure}[t]
     \centering
     \includegraphics[width=0.75\columnwidth]{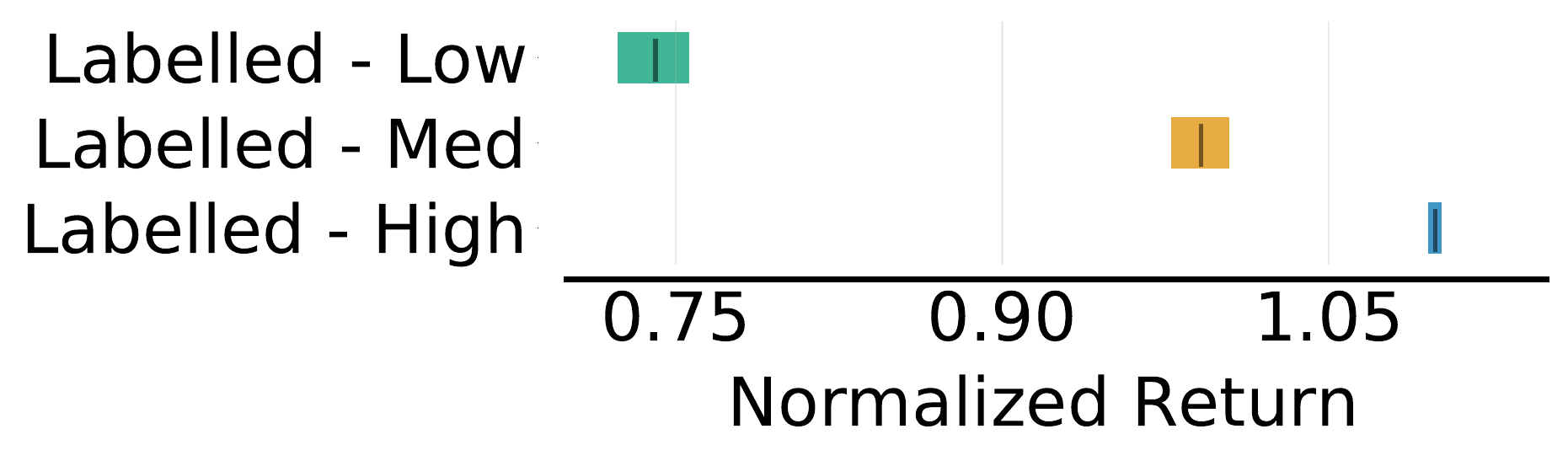}    
     \caption{The 95\% bootstrap CIs of the IQM return of the \ssa agents with varying labelled data quality.}
     \label{fig:decouple_quality_labelled}
     \vspace{3pt}
     \includegraphics[width=0.75\columnwidth]{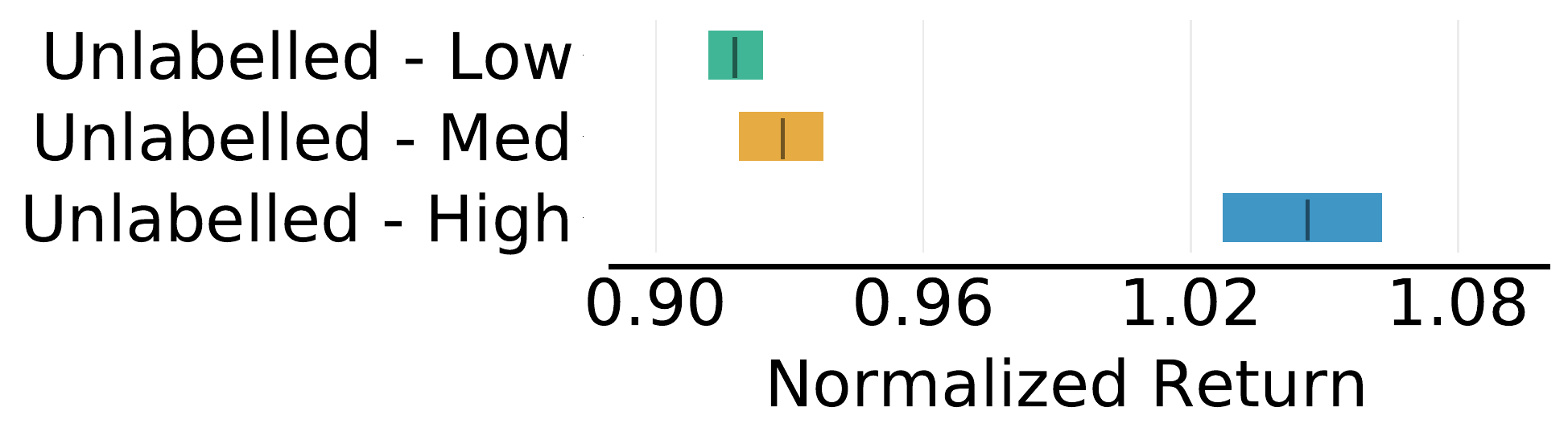}
     \caption{The 95\% bootstrap CIs of the IQM return of the \ssa agents with varying unlabelled data quality.}
     \label{fig:decouple_quality_unlabelled}
 \end{figure}
\paragraph{Quality of Labelled Data} 
We divide the offline trajectories into 3 groups, whose returns are the bottom 0\% to 33\%, 33\% to 67\%, and 67\% to 100\%, respectively. We refer to them as \texttt{Low}, \texttt{Medium}, and \texttt{High} groups. We evaluate the performance of \ssa when the labelled trajectories are sampled from three different groups: \texttt{Low}, \texttt{Med}, and \texttt{High}.
To account for different environment, offline RL methods, and the unlabelled data qualities, 
we consider a total of $12$ cases that cover:\looseness=-1
\vspace{-1em}
\begin{itemize}[leftmargin=*]\itemsep0pt
    \item $2$ datasets \hopper-\medexpert and \walker-\medexpert,
    \item $2$ agents \sscql and \ssdt, and
    \item $3$ quality setups where the unlabelled trajectories are sampled from \texttt{Low}, \texttt{Med}, and \texttt{High} groups.
\end{itemize}
\vspace{-1em}
Both the number of labelled and unlabelled trajectories are set to be $10\%$ of the total number of offline trajectories.
Figure~\ref{fig:decouple_quality_labelled} report the $95\%$ bootstrap CIs of the IQM return for all the $12$ cases and $5$ training instances per case. 
Clearly, as the labelled data quality goes up, the performance of \ssa significantly increases by large margins.

\paragraph{Quality of Unlabelled Data} 
Similar to the above experiment, we sample the unlabelled trajectories from one of the three groups, and train the \ssa agents under $12$ different cases where the labelled data quality varies. Figure~\ref{fig:decouple_quality_unlabelled} reports the $95\%$ bootstrap CIs of the IQM return. 
The performance of \ssa agents increases as the unlabelled data quality increases, and using high quality unlabelled data significantly outperforms the other two cases.

\begin{figure}[t]
    \centering
        \includegraphics[width=\columnwidth]{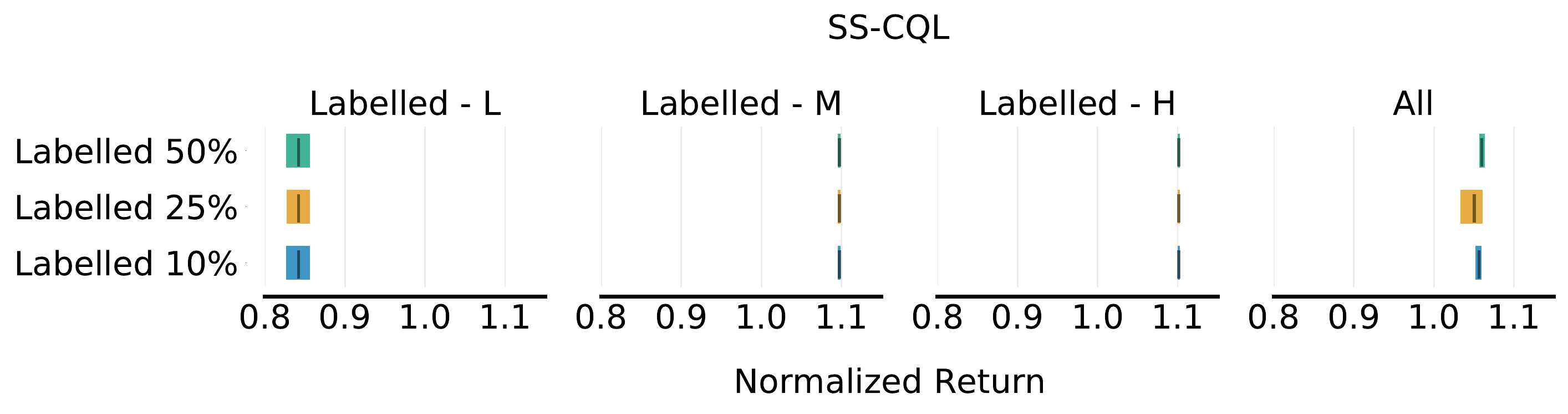}\\
        \includegraphics[width=\columnwidth]{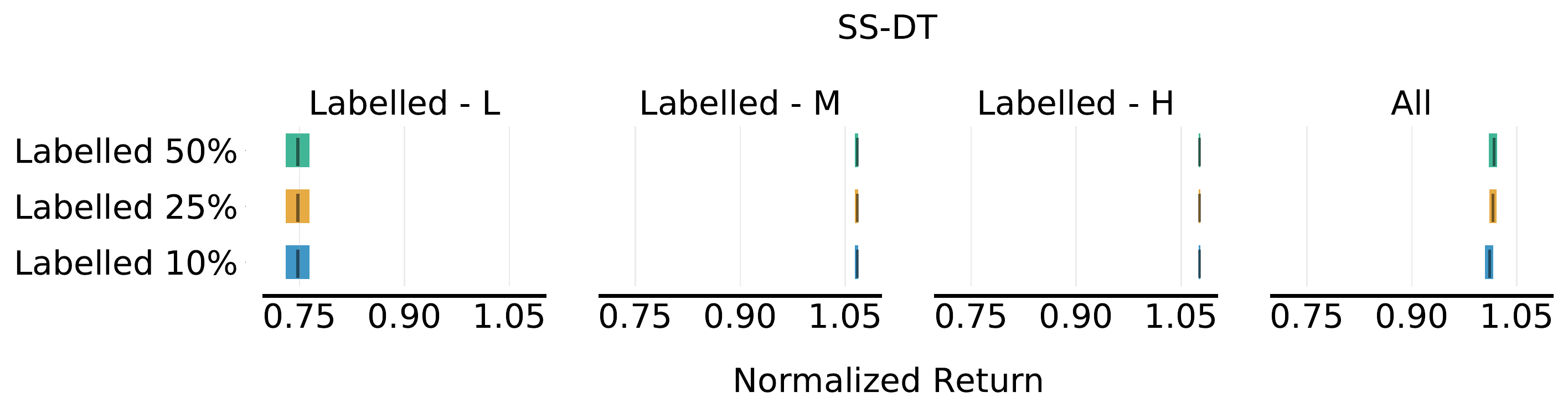}\\
    \caption{The $95\%$ bootstrap CIs of the IQM return of \ssdt and \sscql when the size of the labelled data changes.
        We fix the unlabelled data size to be $10\%$ of the offline dataset size.}
        \label{fig:ci_size_labelled_iqm}
        \includegraphics[width=\columnwidth]{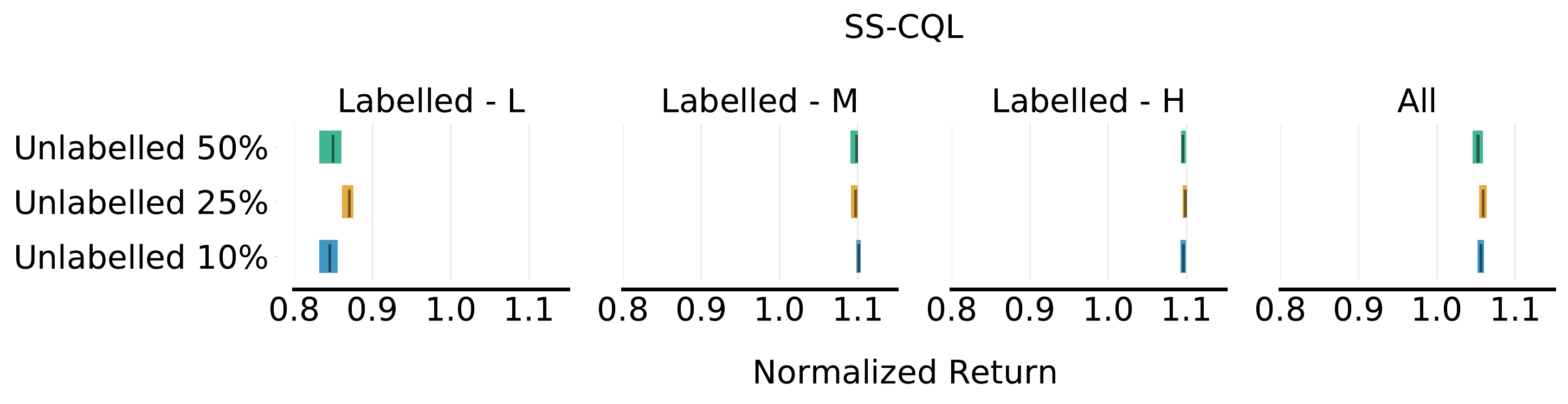}\\
        \includegraphics[width=\columnwidth]{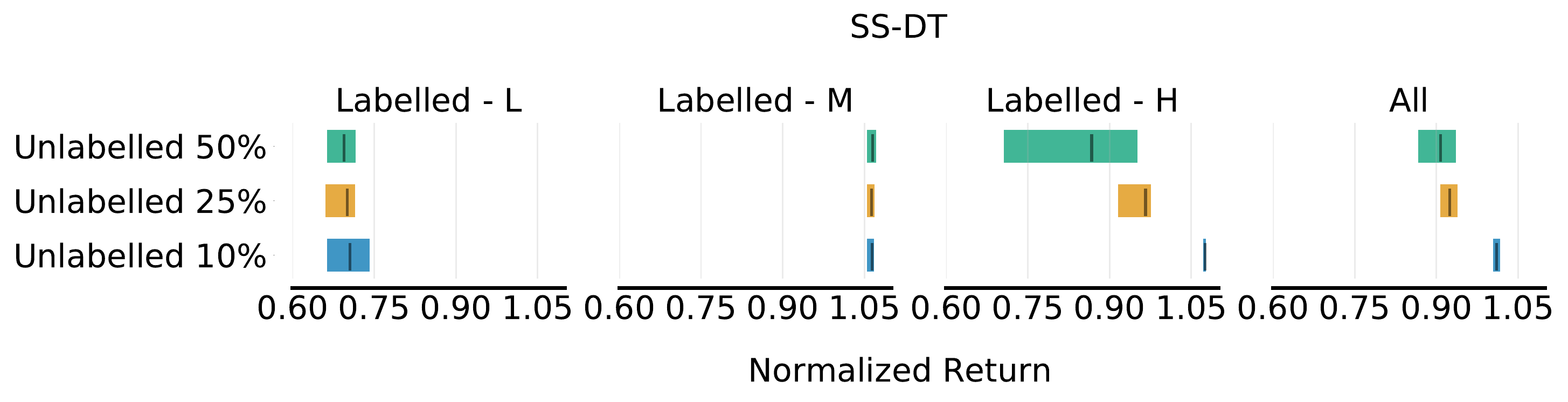}
    \caption{The $95\%$ bootstrap CIs of the IQM return of \ssdt and \sscql when the size of the unlabelled data changes. We fix the labelled data size to be $10\%$ of the offline dataset size.}
        \label{fig:ci_size_unlabelled_iqm}
    \label{fig:ci_size_iqm}
    
\end{figure}

\paragraph{Size of Labelled Data} We vary the number of labelled trajectories as $10\%$, $25\%$, and $50\%$ of the  offline dataset size, while the number of unlabelled trajectories is fixed to be $10\%$.
We train \sscql and \ssdt on the \walker-\medexpert dataset under $9$ data quality setups, where the labelled and unlabelled trajectories are respectively sampled from \texttt{Low}, \texttt{Med}, and \texttt{High} groups. Figure~\ref{fig:ci_size_labelled_iqm} plots the CIs of the IQM return.
Specifically, we consider the results aggregated over all the cases, and also for each individual labelled data quality setup. For all these cases, the performance of both \sscql and \ssdt remain relatively consistent regardless of the number of labelled trajectories. The evaluation performance of \sscql and \ssdt over the course of training for each individual environment and data setup, can be found in Figure~\ref{fig:decouple_size_labelled}.

\paragraph{Size of Unlabelled Data} 
As before, we vary the percentage of unlabelled trajectories as $10\%$, $25\%$, and $50\%$, 
while fixing the labelled data percentage to be $10\%$. We use the same data quality setups as in the previous experiment. Figure~\ref{fig:ci_size_unlabelled_iqm} reports the
$95\%$ bootstrap CIs of the IQM return. 
Interestingly, we found that \ssdt and \sscql respond slightly differently.
\sscql is relatively insensitive to changes in the size of the unlabelled data, as is \ssdt when the labelled data quality is low or moderate. However, when labelled data is of high quality, 
the performance of \ssdt deteriorates as the unlabelled data size increases.
To gain a better understanding of this phenomenon, we investigate the performance for \ssdt for each of the $9$ data quality setups. As shown in Figure~\ref{fig:decouple_size_unlabelled_dt}, when the labelled data is of high quality but the unlabelled data is of lower quality, growing the unlabelled data size harms the performance. 
Our intuition is that, in these cases, the combined dataset will have lower quality than the labelled dataset,
and supervised learning approaches like \dt can be sensitive to this.
More detaileds can be found in Figure~\ref{fig:decouple_size_unlabelled}. 
\begin{figure}[tbh]
\centering
\includegraphics[width=0.6\columnwidth]{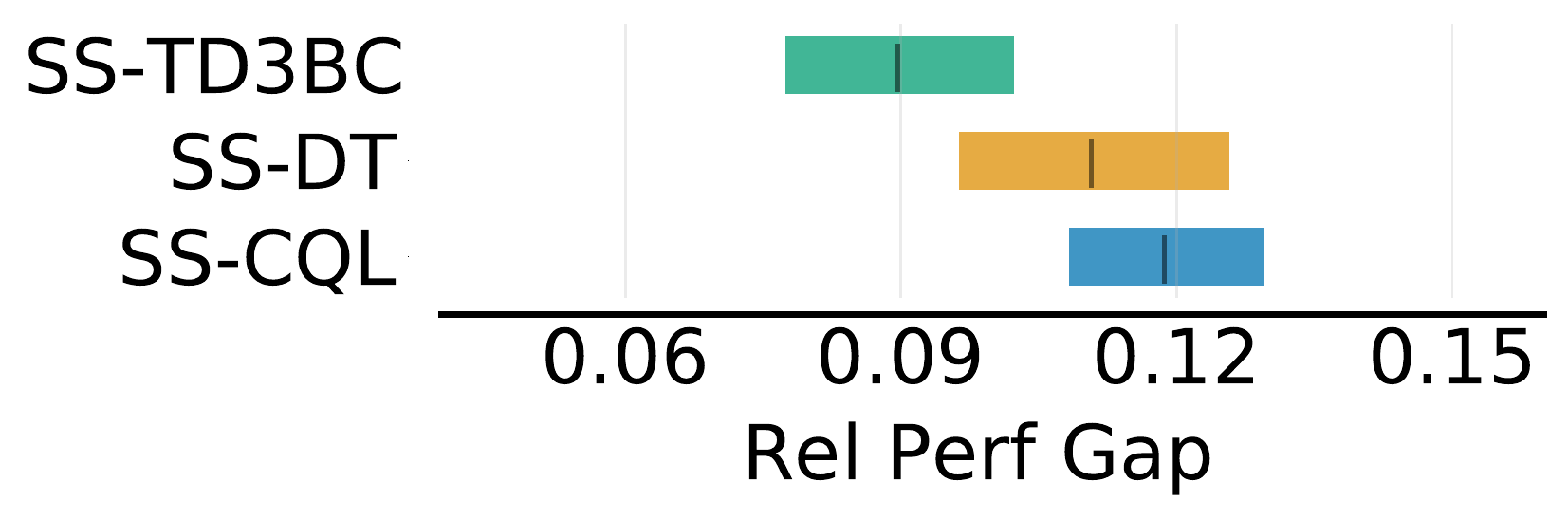}
\caption{The $95\%$ bootstrap CIs of the the relative performance gap of the \ssa agents instantiated with
different offline RL methods.}
\label{fig:perf_gap}
\end{figure}
\subsection{The Choice of Offline RL Algorithm (Q4)}
\label{sec:expr_offline}
For a chosen offline RL method,  the relative performance gap between the corresponding \ssa and oracle agents,
as defined in Equation~\eqref{eq:perf_gap}, 
illustrates how sensitive to missing actions this offline RL method is.
We train \sscql, \ssdt and \sstd on $6$ datasets (the \hopper,\walker environments with \medexpert, \medium, and \medreplay datasets), using the coupled setup as in Section~\ref{sec:expr_main} with $6$ different values of $q$: $10, 30, 50, 70, 90$ and $100$. The aggregated results, shown in Figure~\ref{fig:perf_gap}, indicate that 
\sstd has smallest relative performance gap. This suggests that \td is less sensitive to missing actions then both \dt and \cql. The performance gaps of \sscql and \ssdt are more similar, suggesting that \dt and \cql have similar sensitivity to missing actions.
\section{Conclusion}
\label{sec:conc}
We proposed a novel semi-supervised setup for offline RL where we have access to trajectories with and without action information. 
For this setting, we introduced a simple multi-stage meta-algorithmic 
pipeline. Our experiments identified key properties that enable the agents
to leverage unlabelled data and show that near-optimal
learning can be done with only $10\%$ of the actions labelled for low-to-moderate quality trajectories.
Our work is a step towards creating intelligent agents that can learn from diverse types of
auxiliary demonstrations like online videos, and
it would be interesting to study other heterogeneous data setups for offline RL in the future, including reward-free or pure state-only settings.
\section*{Acknowledgement}
The authors thank 
Zihan Ding, Maryam Fazel-Zarandi, Chi Jin, Mike Rabbat, Aravind Rajeswaran, Yuandong Tian, Lin Xiao, Denis Yarats, Amy Zhang and Dinghuai Zhang for insightful discussions.

\bibliography{main}

\begin{thebibliography}{54}
\providecommand{\natexlab}[1]{#1}
\providecommand{\url}[1]{\texttt{#1}}
\expandafter\ifx\csname urlstyle\endcsname\relax
  \providecommand{\doi}[1]{doi: #1}\else
  \providecommand{\doi}{doi: \begingroup \urlstyle{rm}\Url}\fi

\bibitem[Agarwal et~al.(2021)Agarwal, Schwarzer, Castro, Courville, and
  Bellemare]{agarwal2021deep}
Agarwal, R., Schwarzer, M., Castro, P.~S., Courville, A.~C., and Bellemare, M.
\newblock Deep reinforcement learning at the edge of the statistical precipice.
\newblock \emph{Advances in neural information processing systems},
  34:\penalty0 29304--29320, 2021.

\bibitem[Bain \& Sammut(1995)Bain and Sammut]{bain1995framework}
Bain, M. and Sammut, C.
\newblock A framework for behavioural cloning.
\newblock In \emph{Machine Intelligence 15}, pp.\  103--129, 1995.

\bibitem[Baker et~al.(2022)Baker, Akkaya, Zhokhov, Huizinga, Tang, Ecoffet,
  Houghton, Sampedro, and Clune]{VPT}
Baker, B., Akkaya, I., Zhokhov, P., Huizinga, J., Tang, J., Ecoffet, A.,
  Houghton, B., Sampedro, R., and Clune, J.
\newblock Video pretraining (vpt): Learning to act by watching unlabeled online
  videos, 2022.
\newblock URL \url{https://arxiv.org/abs/2206.11795}.

\bibitem[Bellman(1957)]{bellman1957mdp}
Bellman, R.
\newblock A markovian decision process.
\newblock \emph{Indiana Univ. Math. J.}, 1957.

\bibitem[Bentivegna et~al.(2002)Bentivegna, Ude, Atkeson, and
  Cheng]{bentivegna2002humanoid}
Bentivegna, D.~C., Ude, A., Atkeson, C.~G., and Cheng, G.
\newblock Humanoid robot learning and game playing using pc-based vision.
\newblock In \emph{IEEE/RSJ international conference on intelligent robots and
  systems}, volume~3, pp.\  2449--2454. IEEE, 2002.

\bibitem[Burda et~al.(2019)Burda, Edwards, Pathak, Storkey, Darrell, and
  Efros]{pathak18largescale}
Burda, Y., Edwards, H., Pathak, D., Storkey, A., Darrell, T., and Efros, A.~A.
\newblock Large-scale study of curiosity-driven learning.
\newblock In \emph{ICLR}, 2019.

\bibitem[Burns et~al.(2022)Burns, Yu, Finn, and Hausman]{burns2022offline}
Burns, K., Yu, T., Finn, C., and Hausman, K.
\newblock Offline reinforcement learning at multiple frequencies.
\newblock \emph{arXiv preprint arXiv:2207.13082}, 2022.

\bibitem[Chapelle et~al.(2006)Chapelle, Scholkopf, and Zien]{chapelle2006semi}
Chapelle, O., Scholkopf, B., and Zien, A.
\newblock Semi-supervised learning. 2006.
\newblock \emph{Cambridge, Massachusettes: The MIT Press View Article}, 2,
  2006.

\bibitem[Chen et~al.(2021)Chen, Lu, Rajeswaran, Lee, Grover, Laskin, Abbeel,
  Srinivas, and Mordatch]{chen2021decision}
Chen, L., Lu, K., Rajeswaran, A., Lee, K., Grover, A., Laskin, M., Abbeel, P.,
  Srinivas, A., and Mordatch, I.
\newblock Decision transformer: Reinforcement learning via sequence modeling.
\newblock In \emph{Thirty-Fifth Conference on Neural Information Processing
  Systems}, 2021.
\newblock URL \url{https://openreview.net/forum?id=a7APmM4B9d}.

\bibitem[Emmons et~al.(2021)Emmons, Eysenbach, Kostrikov, and
  Levine]{emmons2021rvs}
Emmons, S., Eysenbach, B., Kostrikov, I., and Levine, S.
\newblock Rvs: What is essential for offline rl via supervised learning?
\newblock \emph{arXiv preprint arXiv:2112.10751}, 2021.

\bibitem[Fralick(1967)]{fralick1967learning}
Fralick, S.
\newblock Learning to recognize patterns without a teacher.
\newblock \emph{IEEE Transactions on Information Theory}, 13\penalty0
  (1):\penalty0 57--64, 1967.

\bibitem[Fu et~al.(2020)Fu, Kumar, Nachum, Tucker, and Levine]{fu2020d4rl}
Fu, J., Kumar, A., Nachum, O., Tucker, G., and Levine, S.
\newblock D4rl: Datasets for deep data-driven reinforcement learning.
\newblock \emph{arXiv preprint arXiv:2004.07219}, 2020.

\bibitem[Fujimoto \& Gu(2021)Fujimoto and Gu]{fujimoto2021minimalist}
Fujimoto, S. and Gu, S.
\newblock A minimalist approach to offline reinforcement learning.
\newblock In \emph{Thirty-Fifth Conference on Neural Information Processing
  Systems}, 2021.
\newblock URL \url{https://openreview.net/forum?id=Q32U7dzWXpc}.

\bibitem[Fujimoto et~al.(2019)Fujimoto, Meger, and Precup]{fujimoto2019off}
Fujimoto, S., Meger, D., and Precup, D.
\newblock Off-policy deep reinforcement learning without exploration.
\newblock In \emph{International Conference on Machine Learning}, pp.\
  2052--2062. PMLR, 2019.

\bibitem[Ghasemipour et~al.(2021)Ghasemipour, Schuurmans, and
  Gu]{ghasemipour2021emaq}
Ghasemipour, S. K.~S., Schuurmans, D., and Gu, S.~S.
\newblock Emaq: Expected-max q-learning operator for simple yet effective
  offline and online rl.
\newblock In \emph{International Conference on Machine Learning}, pp.\
  3682--3691. PMLR, 2021.

\bibitem[Ghosh et~al.(2022)Ghosh, Ajay, Agrawal, and Levine]{ghosh2022offline}
Ghosh, D., Ajay, A., Agrawal, P., and Levine, S.
\newblock Offline rl policies should be trained to be adaptive.
\newblock In \emph{International Conference on Machine Learning}, pp.\
  7513--7530. PMLR, 2022.

\bibitem[Gupta et~al.(2017)Gupta, Devin, Liu, Abbeel, and
  Levine]{gupta2017learning}
Gupta, A., Devin, C., Liu, Y., Abbeel, P., and Levine, S.
\newblock Learning invariant feature spaces to transfer skills with
  reinforcement learning.
\newblock \emph{arXiv preprint arXiv:1703.02949}, 2017.

\bibitem[Henaff et~al.(2022)Henaff, Raileanu, Jiang, and
  Rockt{\"a}schel]{henaff2022exploration}
Henaff, M., Raileanu, R., Jiang, M., and Rockt{\"a}schel, T.
\newblock Exploration via elliptical episodic bonuses.
\newblock In Oh, A.~H., Agarwal, A., Belgrave, D., and Cho, K. (eds.),
  \emph{Advances in Neural Information Processing Systems}, 2022.
\newblock URL \url{https://openreview.net/forum?id=Xg-yZos9qJQ}.

\bibitem[Ho \& Ermon(2016)Ho and Ermon]{ho2016generative}
Ho, J. and Ermon, S.
\newblock Generative adversarial imitation learning.
\newblock \emph{Advances in neural information processing systems}, 29, 2016.

\bibitem[Ijspeert et~al.(2002)Ijspeert, Nakanishi, and
  Schaal]{ijspeert2002movement}
Ijspeert, A.~J., Nakanishi, J., and Schaal, S.
\newblock Movement imitation with nonlinear dynamical systems in humanoid
  robots.
\newblock In \emph{Proceedings 2002 IEEE International Conference on Robotics
  and Automation (Cat. No. 02CH37292)}, volume~2, pp.\  1398--1403. IEEE, 2002.

\bibitem[Janner et~al.(2021)Janner, Li, and Levine]{janner2021offline}
Janner, M., Li, Q., and Levine, S.
\newblock Offline reinforcement learning as one big sequence modeling problem.
\newblock In \emph{Thirty-Fifth Conference on Neural Information Processing
  Systems}, 2021.
\newblock URL \url{https://openreview.net/forum?id=wgeK563QgSw}.

\bibitem[Jaques et~al.(2019)Jaques, Ghandeharioun, Shen, Ferguson, Lapedriza,
  Jones, Gu, and Picard]{jaques2019way}
Jaques, N., Ghandeharioun, A., Shen, J.~H., Ferguson, C., Lapedriza, A., Jones,
  N., Gu, S., and Picard, R.
\newblock Way off-policy batch deep reinforcement learning of implicit human
  preferences in dialog.
\newblock \emph{arXiv preprint arXiv:1907.00456}, 2019.

\bibitem[Kidambi et~al.(2021)Kidambi, Chang, and Sun]{kidambi2021mobile}
Kidambi, R., Chang, J.~D., and Sun, W.
\newblock Mob{ILE}: Model-based imitation learning from observation alone.
\newblock In Beygelzimer, A., Dauphin, Y., Liang, P., and Vaughan, J.~W.
  (eds.), \emph{Advances in Neural Information Processing Systems}, 2021.
\newblock URL \url{https://openreview.net/forum?id=_Rtm4rYnIIL}.

\bibitem[Kim et~al.(2020)Kim, Gu, Song, Zhao, and Ermon]{kim2020domain}
Kim, K., Gu, Y., Song, J., Zhao, S., and Ermon, S.
\newblock Domain adaptive imitation learning.
\newblock In \emph{International Conference on Machine Learning}, pp.\
  5286--5295. PMLR, 2020.

\bibitem[Kingma \& Ba(2014)Kingma and Ba]{kingma2014adam}
Kingma, D.~P. and Ba, J.
\newblock Adam: A method for stochastic optimization.
\newblock \emph{arXiv preprint arXiv:1412.6980}, 2014.

\bibitem[Koller \& Friedman(2009)Koller and Friedman]{koller2009probabilistic}
Koller, D. and Friedman, N.
\newblock \emph{Probabilistic graphical models: principles and techniques}.
\newblock MIT press, 2009.

\bibitem[Kostrikov et~al.(2021{\natexlab{a}})Kostrikov, Fergus, Tompson, and
  Nachum]{kostrikov2021offline_divergence}
Kostrikov, I., Fergus, R., Tompson, J., and Nachum, O.
\newblock Offline reinforcement learning with fisher divergence critic
  regularization.
\newblock In \emph{International Conference on Machine Learning}, pp.\
  5774--5783. PMLR, 2021{\natexlab{a}}.

\bibitem[Kostrikov et~al.(2021{\natexlab{b}})Kostrikov, Nair, and
  Levine]{kostrikov2021offline}
Kostrikov, I., Nair, A., and Levine, S.
\newblock Offline reinforcement learning with implicit q-learning,
  2021{\natexlab{b}}.

\bibitem[Kumar et~al.(2019)Kumar, Fu, Tucker, and Levine]{kumar2019stabilizing}
Kumar, A., Fu, J., Tucker, G., and Levine, S.
\newblock Stabilizing off-policy q-learning via bootstrapping error reduction.
\newblock \emph{arXiv preprint arXiv:1906.00949}, 2019.

\bibitem[Kumar et~al.(2020)Kumar, Zhou, Tucker, and
  Levine]{kumar2020conservative}
Kumar, A., Zhou, A., Tucker, G., and Levine, S.
\newblock Conservative q-learning for offline reinforcement learning.
\newblock \emph{arXiv preprint arXiv:2006.04779}, 2020.

\bibitem[Lakshminarayanan et~al.(2017)Lakshminarayanan, Pritzel, and
  Blundell]{lakshminarayanan2017simple}
Lakshminarayanan, B., Pritzel, A., and Blundell, C.
\newblock Simple and scalable predictive uncertainty estimation using deep
  ensembles.
\newblock \emph{Advances in neural information processing systems}, 30, 2017.

\bibitem[Lee et~al.(2022)Lee, Nachum, Yang, Lee, Freeman, Xu, Guadarrama,
  Fischer, Jang, Michalewski, et~al.]{lee2022multi}
Lee, K.-H., Nachum, O., Yang, M., Lee, L., Freeman, D., Xu, W., Guadarrama, S.,
  Fischer, I., Jang, E., Michalewski, H., et~al.
\newblock Multi-game decision transformers.
\newblock \emph{arXiv preprint arXiv:2205.15241}, 2022.

\bibitem[Levine et~al.(2020)Levine, Kumar, Tucker, and Fu]{levine2020offline}
Levine, S., Kumar, A., Tucker, G., and Fu, J.
\newblock Offline reinforcement learning: Tutorial, review, and perspectives on
  open problems.
\newblock \emph{arXiv preprint arXiv:2005.01643}, 2020.

\bibitem[Liu et~al.(2018)Liu, Gupta, Abbeel, and Levine]{liu2018imitation}
Liu, Y., Gupta, A., Abbeel, P., and Levine, S.
\newblock Imitation from observation: Learning to imitate behaviors from raw
  video via context translation.
\newblock In \emph{2018 IEEE International Conference on Robotics and
  Automation (ICRA)}, pp.\  1118--1125. IEEE, 2018.

\bibitem[Mazoure et~al.(2021)Mazoure, Kostrikov, Nachum, and
  Tompson]{mazoure2021improving}
Mazoure, B., Kostrikov, I., Nachum, O., and Tompson, J.
\newblock Improving zero-shot generalization in offline reinforcement learning
  using generalized similarity functions.
\newblock \emph{arXiv preprint arXiv:2111.14629}, 2021.

\bibitem[Nachum et~al.(2019)Nachum, Dai, Kostrikov, Chow, Li, and
  Schuurmans]{nachum2019algaedice}
Nachum, O., Dai, B., Kostrikov, I., Chow, Y., Li, L., and Schuurmans, D.
\newblock Algaedice: Policy gradient from arbitrary experience.
\newblock \emph{arXiv preprint arXiv:1912.02074}, 2019.

\bibitem[Ouali et~al.(2020)Ouali, Hudelot, and Tami]{ouali2020overview}
Ouali, Y., Hudelot, C., and Tami, M.
\newblock An overview of deep semi-supervised learning.
\newblock \emph{arXiv preprint arXiv:2006.05278}, 2020.

\bibitem[Pathak et~al.(2017)Pathak, Agrawal, Efros, and
  Darrell]{pathak2017curiosity}
Pathak, D., Agrawal, P., Efros, A.~A., and Darrell, T.
\newblock Curiosity-driven exploration by self-supervised prediction.
\newblock In \emph{International conference on machine learning}, pp.\
  2778--2787. PMLR, 2017.

\bibitem[Rafailov et~al.(2021)Rafailov, Yu, Rajeswaran, and
  Finn]{rafailov2021offline}
Rafailov, R., Yu, T., Rajeswaran, A., and Finn, C.
\newblock Offline reinforcement learning from images with latent space models.
\newblock In \emph{Learning for Dynamics and Control}, pp.\  1154--1168. PMLR,
  2021.

\bibitem[Reed et~al.(2022)Reed, Zolna, Parisotto, Colmenarejo, Novikov,
  Barth-Maron, Gimenez, Sulsky, Kay, Springenberg, et~al.]{reed2022generalist}
Reed, S., Zolna, K., Parisotto, E., Colmenarejo, S.~G., Novikov, A.,
  Barth-Maron, G., Gimenez, M., Sulsky, Y., Kay, J., Springenberg, J.~T.,
  et~al.
\newblock A generalist agent.
\newblock \emph{arXiv preprint arXiv:2205.06175}, 2022.

\bibitem[Schmeckpeper et~al.(2020{\natexlab{a}})Schmeckpeper, Rybkin,
  Daniilidis, Levine, and Finn]{schmeckpeper2020reinforcement}
Schmeckpeper, K., Rybkin, O., Daniilidis, K., Levine, S., and Finn, C.
\newblock Reinforcement learning with videos: Combining offline observations
  with interaction.
\newblock \emph{arXiv preprint arXiv:2011.06507}, 2020{\natexlab{a}}.

\bibitem[Schmeckpeper et~al.(2020{\natexlab{b}})Schmeckpeper, Xie, Rybkin,
  Tian, Daniilidis, Levine, and Finn]{schmeckpeper2020learning}
Schmeckpeper, K., Xie, A., Rybkin, O., Tian, S., Daniilidis, K., Levine, S.,
  and Finn, C.
\newblock Learning predictive models from observation and interaction.
\newblock In \emph{European Conference on Computer Vision}, pp.\  708--725.
  Springer, 2020{\natexlab{b}}.

\bibitem[Sermanet et~al.(2018)Sermanet, Lynch, Chebotar, Hsu, Jang, Schaal,
  Levine, and Brain]{sermanet2018time}
Sermanet, P., Lynch, C., Chebotar, Y., Hsu, J., Jang, E., Schaal, S., Levine,
  S., and Brain, G.
\newblock Time-contrastive networks: Self-supervised learning from video.
\newblock In \emph{2018 IEEE international conference on robotics and
  automation (ICRA)}, pp.\  1134--1141. IEEE, 2018.

\bibitem[Sharma et~al.(2019)Sharma, Pathak, and Gupta]{sharma19thirdperson}
Sharma, P., Pathak, D., and Gupta, A.
\newblock Third-person visual imitation learning via decoupled hierarchical
  controller.
\newblock In \emph{NeurIPS}, 2019.

\bibitem[Stadie et~al.(2017)Stadie, Abbeel, and Sutskever]{thirdpersonIL}
Stadie, B.~C., Abbeel, P., and Sutskever, I.
\newblock Third-person imitation learning.
\newblock \emph{CoRR}, abs/1703.01703, 2017.
\newblock URL \url{http://arxiv.org/abs/1703.01703}.

\bibitem[Torabi et~al.(2018{\natexlab{a}})Torabi, Warnell, and Stone]{BCO}
Torabi, F., Warnell, G., and Stone, P.
\newblock Behavioral cloning from observation.
\newblock \emph{CoRR}, abs/1805.01954, 2018{\natexlab{a}}.
\newblock URL \url{http://arxiv.org/abs/1805.01954}.

\bibitem[Torabi et~al.(2018{\natexlab{b}})Torabi, Warnell, and Stone]{gailfo}
Torabi, F., Warnell, G., and Stone, P.
\newblock Generative adversarial imitation from observation.
\newblock \emph{CoRR}, abs/1807.06158, 2018{\natexlab{b}}.
\newblock URL \url{http://arxiv.org/abs/1807.06158}.

\bibitem[Torabi et~al.(2019)Torabi, Warnell, and Stone]{torabi2019recent}
Torabi, F., Warnell, G., and Stone, P.
\newblock Recent advances in imitation learning from observation.
\newblock \emph{arXiv preprint arXiv:1905.13566}, 2019.

\bibitem[Van~Engelen \& Hoos(2020)Van~Engelen and Hoos]{van2020survey}
Van~Engelen, J.~E. and Hoos, H.~H.
\newblock A survey on semi-supervised learning.
\newblock \emph{Machine Learning}, 109\penalty0 (2):\penalty0 373--440, 2020.

\bibitem[Wu et~al.(2019)Wu, Tucker, and Nachum]{wu2019behavior}
Wu, Y., Tucker, G., and Nachum, O.
\newblock Behavior regularized offline reinforcement learning.
\newblock \emph{arXiv preprint arXiv:1911.11361}, 2019.

\bibitem[You et~al.(2019)You, Li, Reddi, Hseu, Kumar, Bhojanapalli, Song,
  Demmel, Keutzer, and Hsieh]{you2019large}
You, Y., Li, J., Reddi, S., Hseu, J., Kumar, S., Bhojanapalli, S., Song, X.,
  Demmel, J., Keutzer, K., and Hsieh, C.-J.
\newblock Large batch optimization for deep learning: Training bert in 76
  minutes.
\newblock \emph{arXiv preprint arXiv:1904.00962}, 2019.

\bibitem[Yu et~al.(2022)Yu, Kumar, Chebotar, Hausman, Finn, and
  Levine]{yu2022leverage}
Yu, T., Kumar, A., Chebotar, Y., Hausman, K., Finn, C., and Levine, S.
\newblock How to leverage unlabeled data in offline reinforcement learning.
\newblock \emph{arXiv preprint arXiv:2202.01741}, 2022.

\bibitem[Zheng et~al.(2022)Zheng, Zhang, and Grover]{zheng2022online}
Zheng, Q., Zhang, A., and Grover, A.
\newblock Online decision transformer.
\newblock \emph{arXiv preprint arXiv:2202.05607}, 2022.

\bibitem[Zhu(2005)]{zhu05survey}
Zhu, X.
\newblock Semi-supervised learning literature survey.
\newblock Technical Report 1530, Computer Sciences, University of
  Wisconsin-Madison, 2005.

\end{thebibliography}
\bibliographystyle{icml2023}

\newpage
\appendix
\onecolumn
\section{Experiment Details}
\label{app:hp}
In this section, we provide more details about our experiments. For all the offline RL methods we consider,
we use our own implementations adopted from the following codebases:

DT \hskip5pt \url{https://github.com/facebookresearch/online-dt} \\
TD3BC \hskip5pt \url{https://github.com/sfujim/TD3_BC}\\
CQL \hskip5pt \url{https://github.com/scottemmons/youngs-cql}

We use the stochastic DT proposed by \citet{zheng2022online}. For offline RL, its performance is similar to the deterministic DT~\cite{chen2021decision}. The policy parameter is optimized by the LAMB optimizer~\cite{you2019large} with $\eps=10^{-8}$. The log-temperature parameter is optimized by the Adam optimzier~\cite{kingma2014adam}.
The architecture and other hyperparameters are listed in Tabel~\ref{tbl:hp_dt}.
For TD3BC, we optimize both the critic and actor parameters by the Adam optimizer. The complete hyperparameters are listed in Table~\ref{tbl:hp_td3bc}.
For CQL, we also use the Adam optimizer to optimize the critic, actor and the log-temperature parameters.
The architecture of critic and actor networks and the other 
hyperparameters are listed in Table~\ref{tbl:hp_cql}. We use batch size $256$ and context length $20$ for DT, where each batch contains $5120$ states. Correspondingly, we use batch size $5120$ for CQL and TD3BC.

\begin{table}[tbh]
    \centering
    \begin{tabular}{l    l  }
    \toprule
         \textbf{Hyperparameter} & \textbf{Value} \\ \midrule
         number of layers & $4$ \\
         number of attention heads & $4$\\
         embedding dimension & $512$\\
         context length & $20$\\
         dropout & $0.1$\\
         activation function & relu\\
         batch size & $256$\\
         learning rate for policy & $0.0001$ \\
         weight decay for policy & $0.001$ \\
         learning rate for log-temperature & $0.0001$ \\
         gradient norm clip & $0.25$ \\
         learning rate warmup & linear warmup for $10^4$ steps \\
         target entropy & $-\text{dim}(\mathcal{A})$ \\ 
         evaluation return-to-go & 3600 Hopper\\
                                 & 5000 Walker \\
                                 & 6000 HalfCheetah\\
         \bottomrule
    \end{tabular}
    \caption{The hyperparameters used for DT.}
    \label{tbl:hp_dt}
\end{table}

\begin{table}[h]
    \centering
    \begin{tabular}{l    l  }
    \toprule
         \textbf{Hyperparameter} & \textbf{Value} \\ \midrule
         discount factor & $0.99$ \\
         target update rate & $0.005$\\
         policy noise & $0.2$\\
         policy noise clipping & $(-0.5, 0.5)$\\
         policy update frequency & $2$ \\
         critic learning rate & $0.0003$\\
         critic hidden dim & $256$\\
         critic hidden layers & $2$\\
         actor learning rate & $0.0003$\\
         actor hidden dim & $256$\\
         actor hidden layers & $2$\\
         activation function & ReLU \\
         regularization parameter $\alpha$ & 2.5 \\
       \bottomrule
    \end{tabular}
    \caption{The hyperparameters used for TD3BC.}
    \label{tbl:hp_td3bc}
\end{table}

\begin{table}[h]
    \centering
    \begin{tabular}{l    l  }
    \toprule
         \textbf{Hyperparameter} & \textbf{Value} \\ \midrule
         discount factor & $0.99$ \\
         target update rate & $0.005$\\
         critic learning rate & $0.0003$\\
         critic hidden dim & $256$\\
         critic hidden layers & $3$\\
         actor learning rate & $0.0001$\\
         actor hidden dim & $256$\\
         actor hidden layers & $3$\\
         log-temperature learning rate & $0.0003$\\
         activation function & ReLU \\
         number of sampled actions & 10 \\
         target entropy & $-\text{dim}(\mathcal{A})$ \\
         minimum Q weight value & $5$ \\
         Lagrange & False\\
         Importance Sampling & True\\
       \bottomrule
    \end{tabular}
    \caption{The hyperparameters used for CQL.}
    \label{tbl:hp_cql}
\end{table}

\section{The Return Distributions of the D4RL Datasets}
\label{app:return_distributions}
\begin{figure}[H]
    \centering
    \includegraphics[width=0.3\columnwidth]{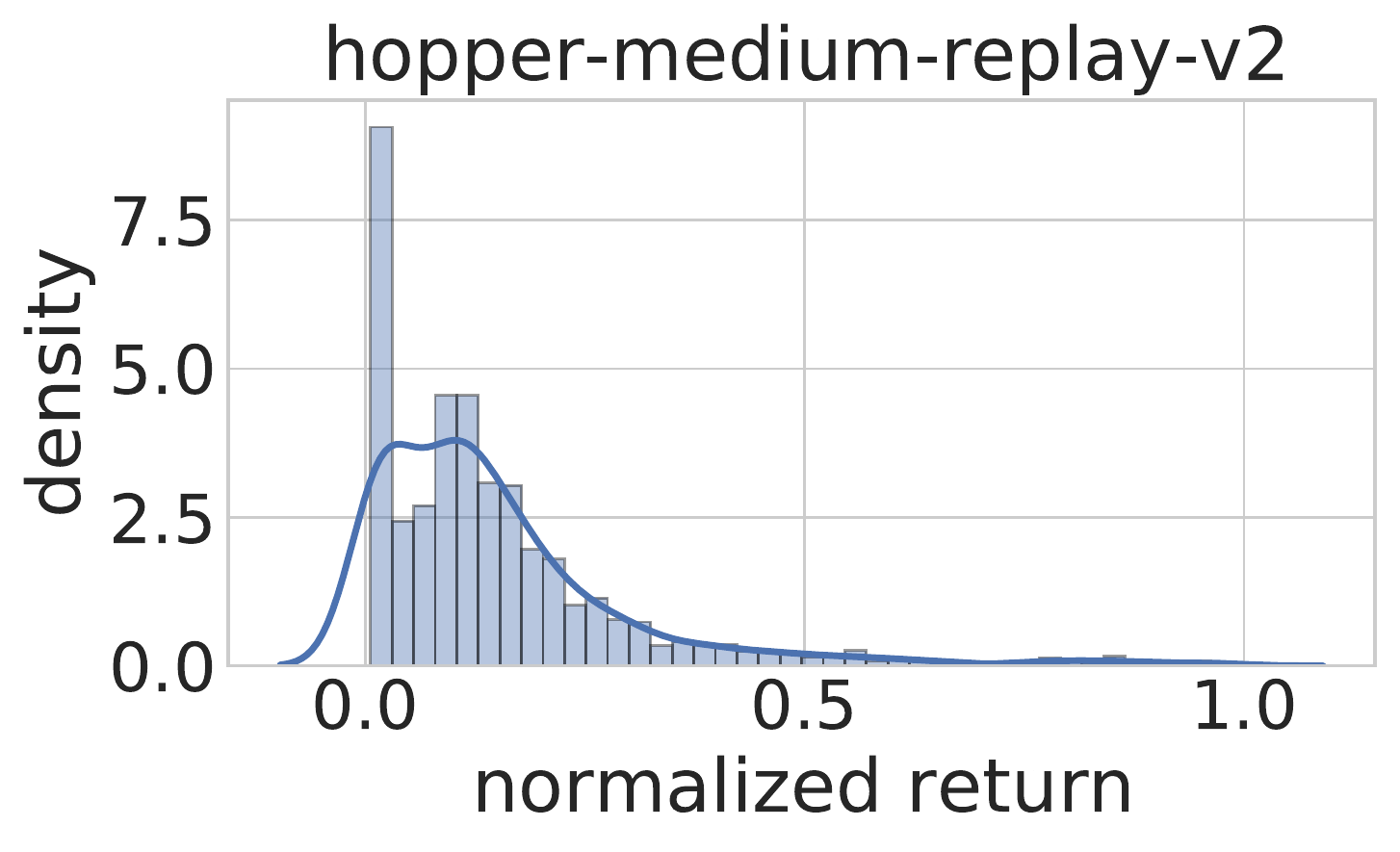}
    \includegraphics[width=0.3\columnwidth]{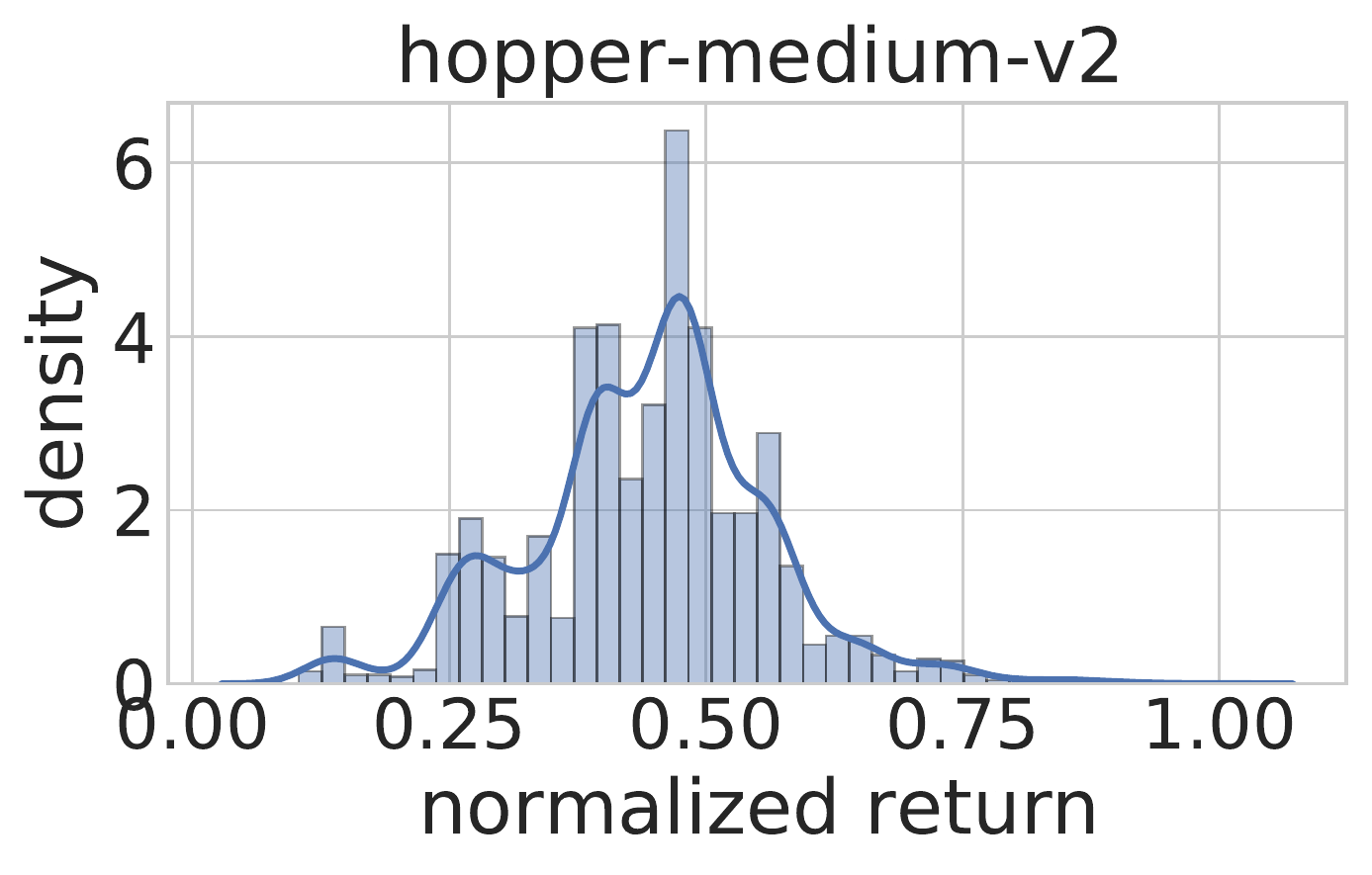}
    \includegraphics[width=0.3\columnwidth]{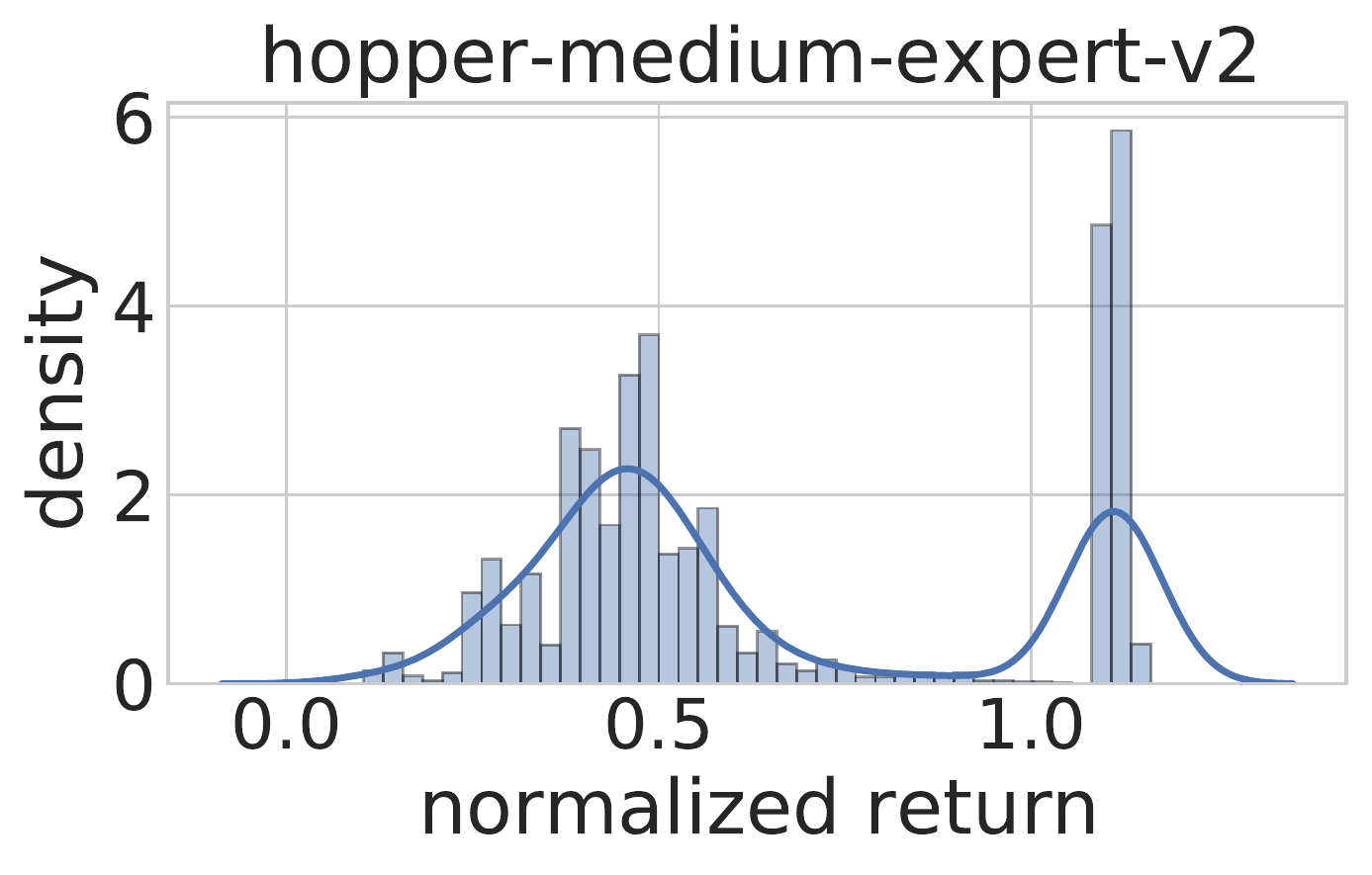} \\
    \includegraphics[width=0.3\columnwidth]{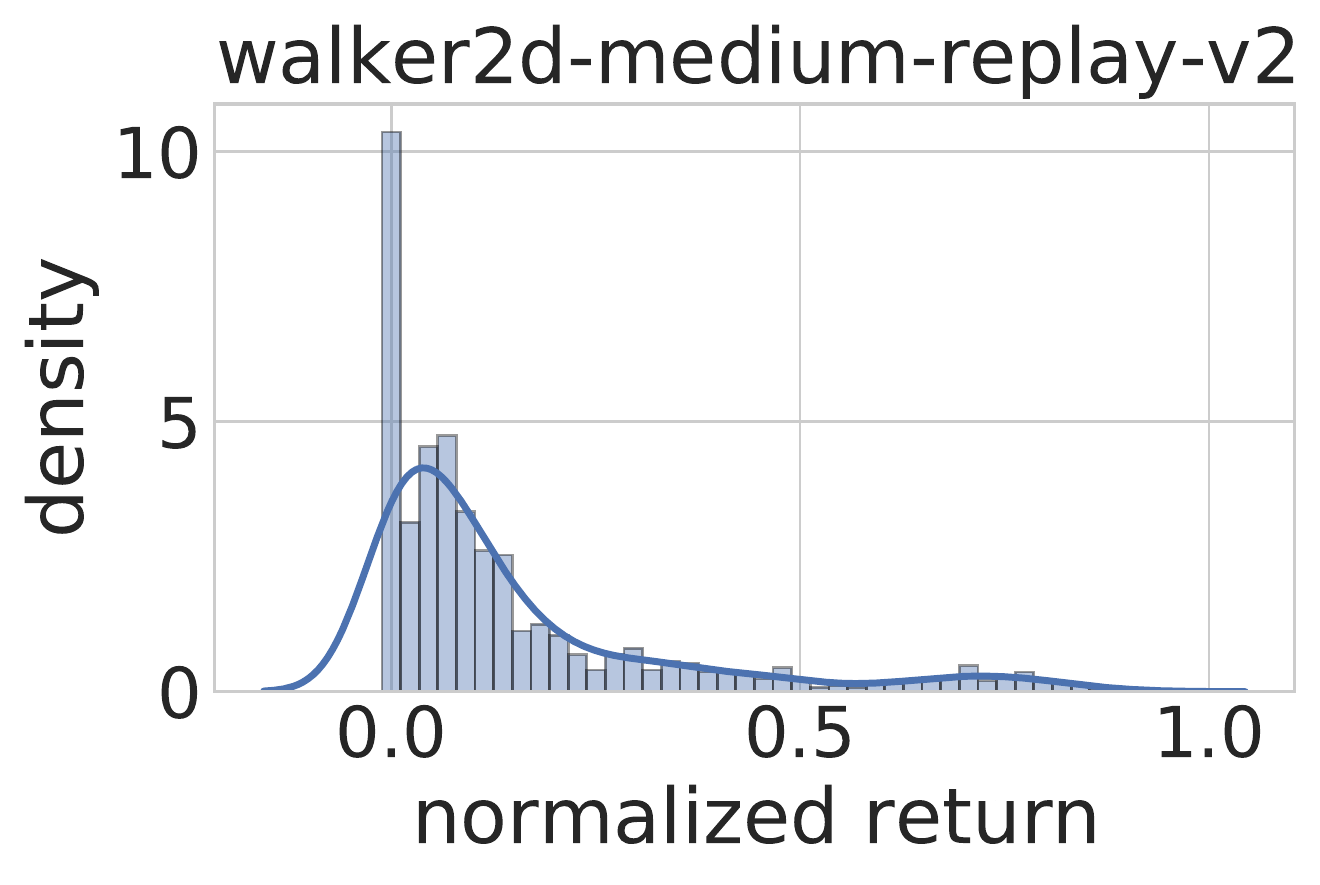}
    \includegraphics[width=0.3\columnwidth]{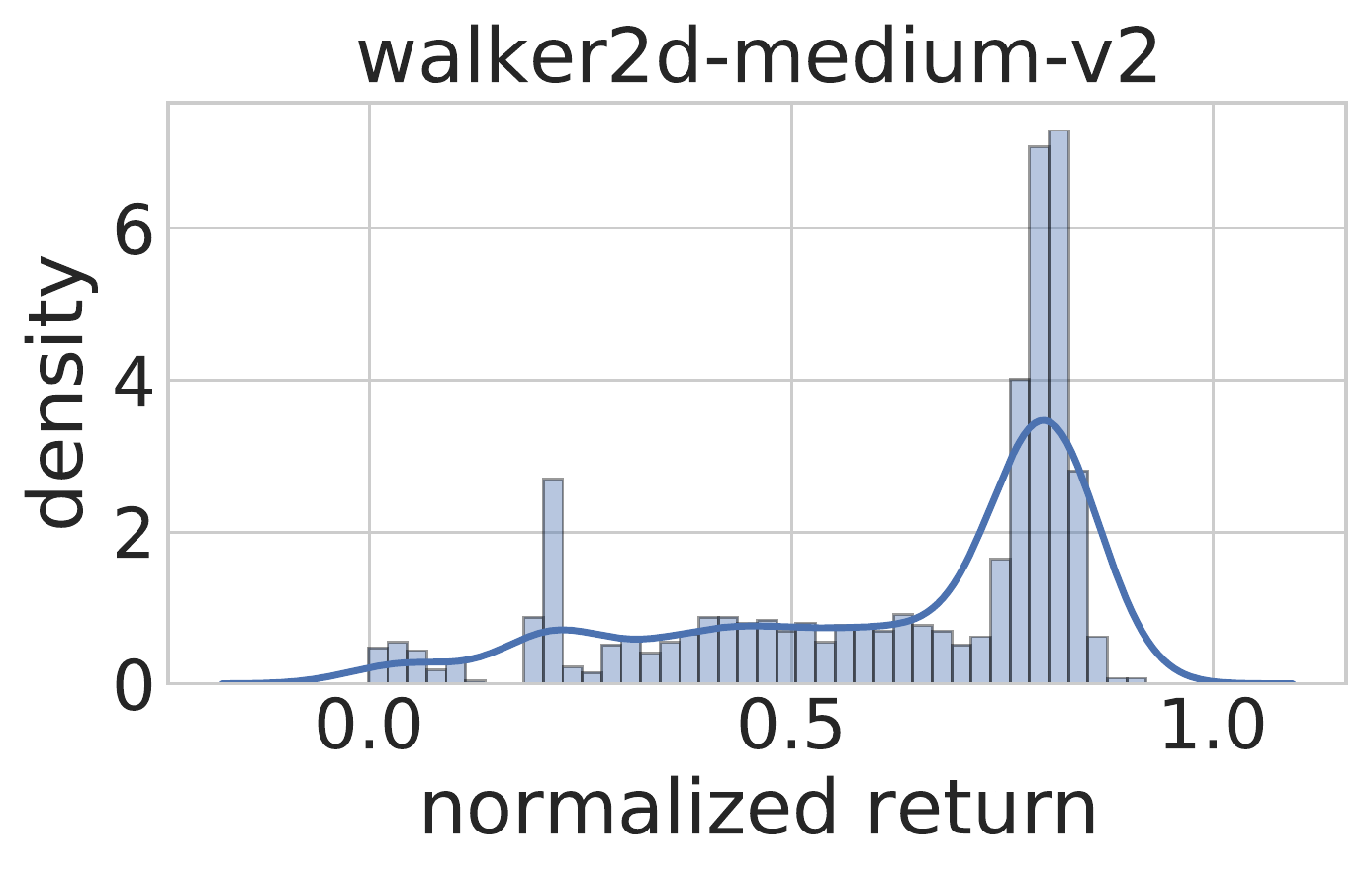}
    \includegraphics[width=0.3\columnwidth]{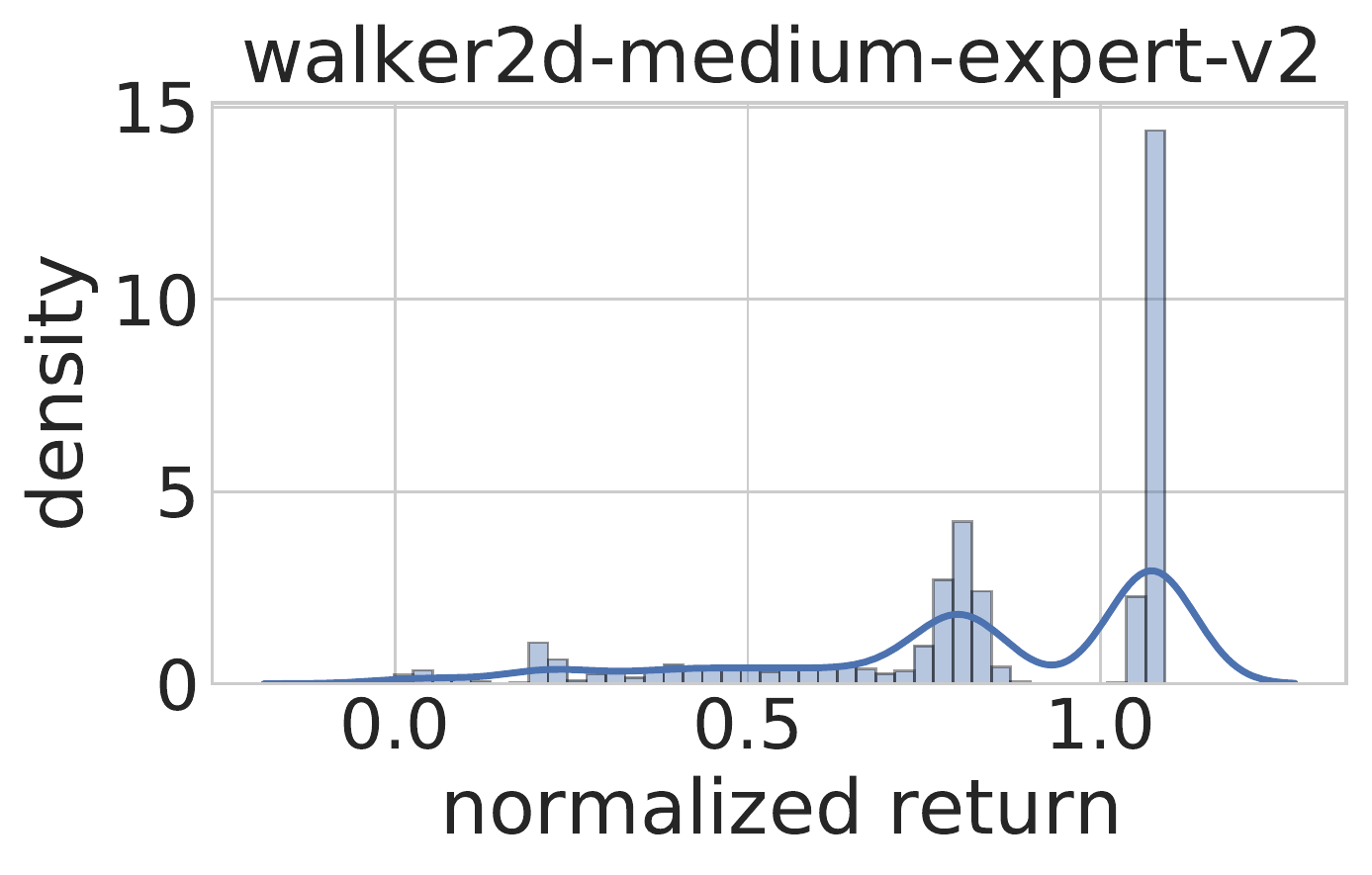} \\
    \includegraphics[width=0.3\columnwidth]{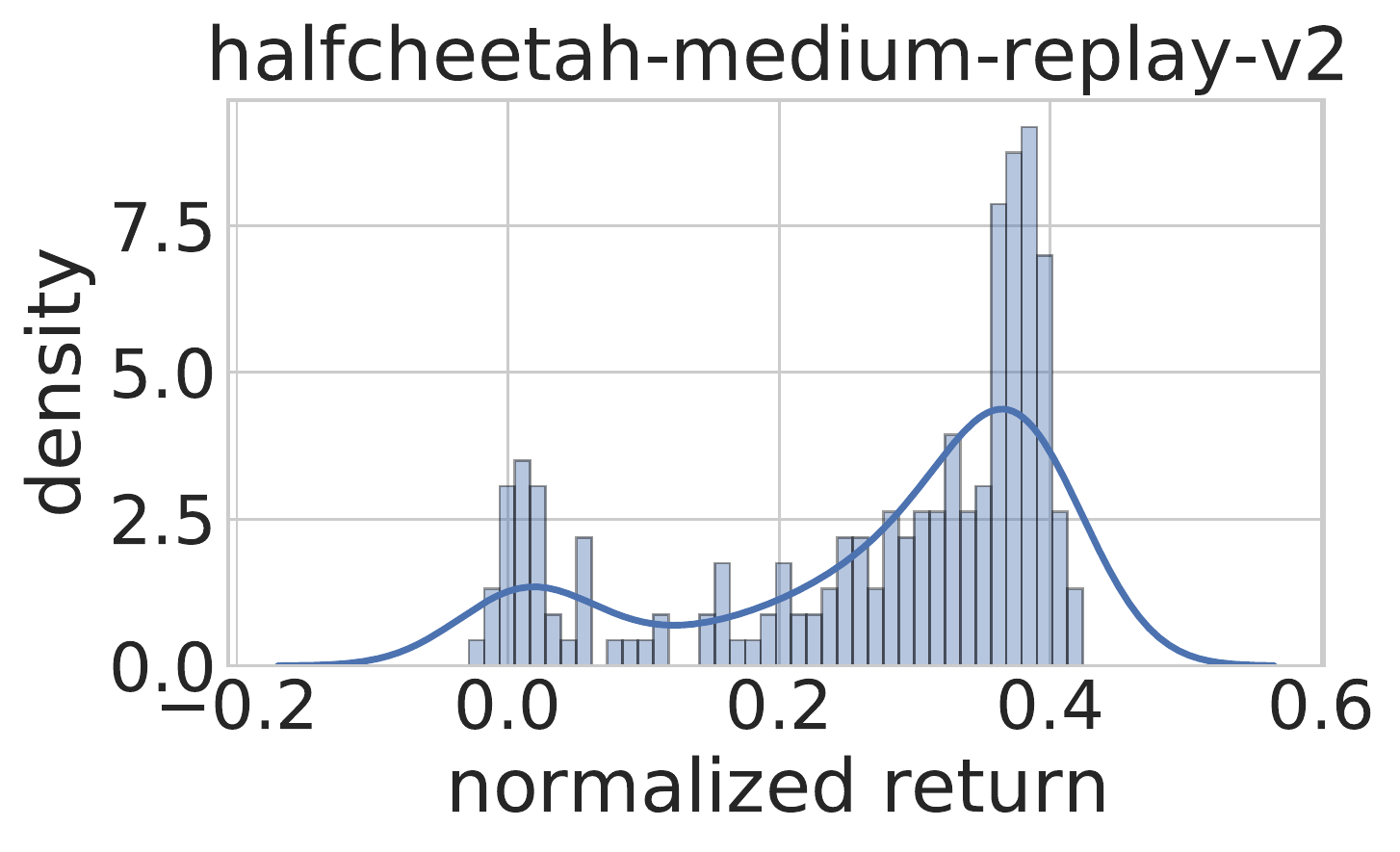}
    \includegraphics[width=0.3\columnwidth]{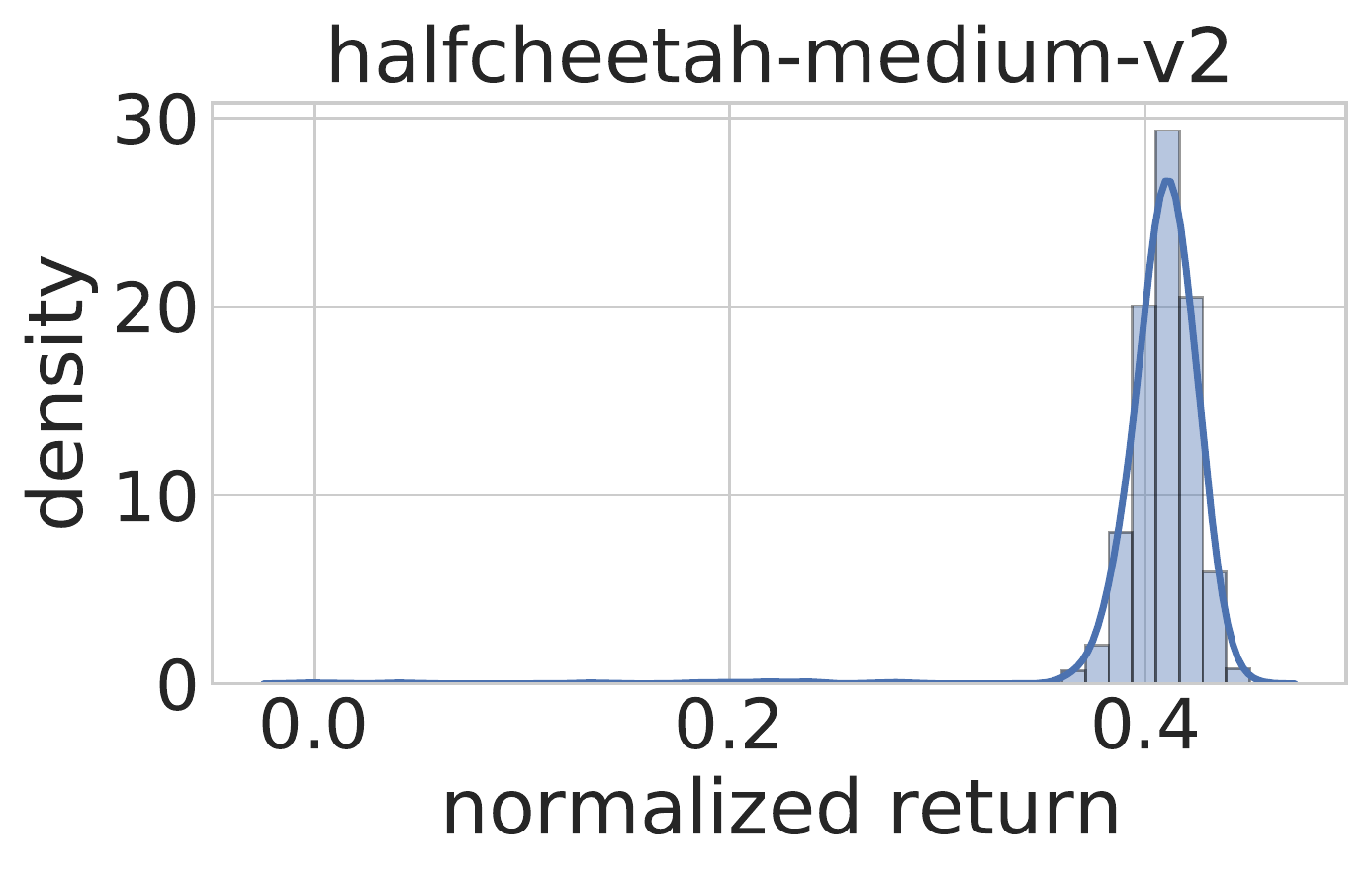}
    \includegraphics[width=0.3\columnwidth]{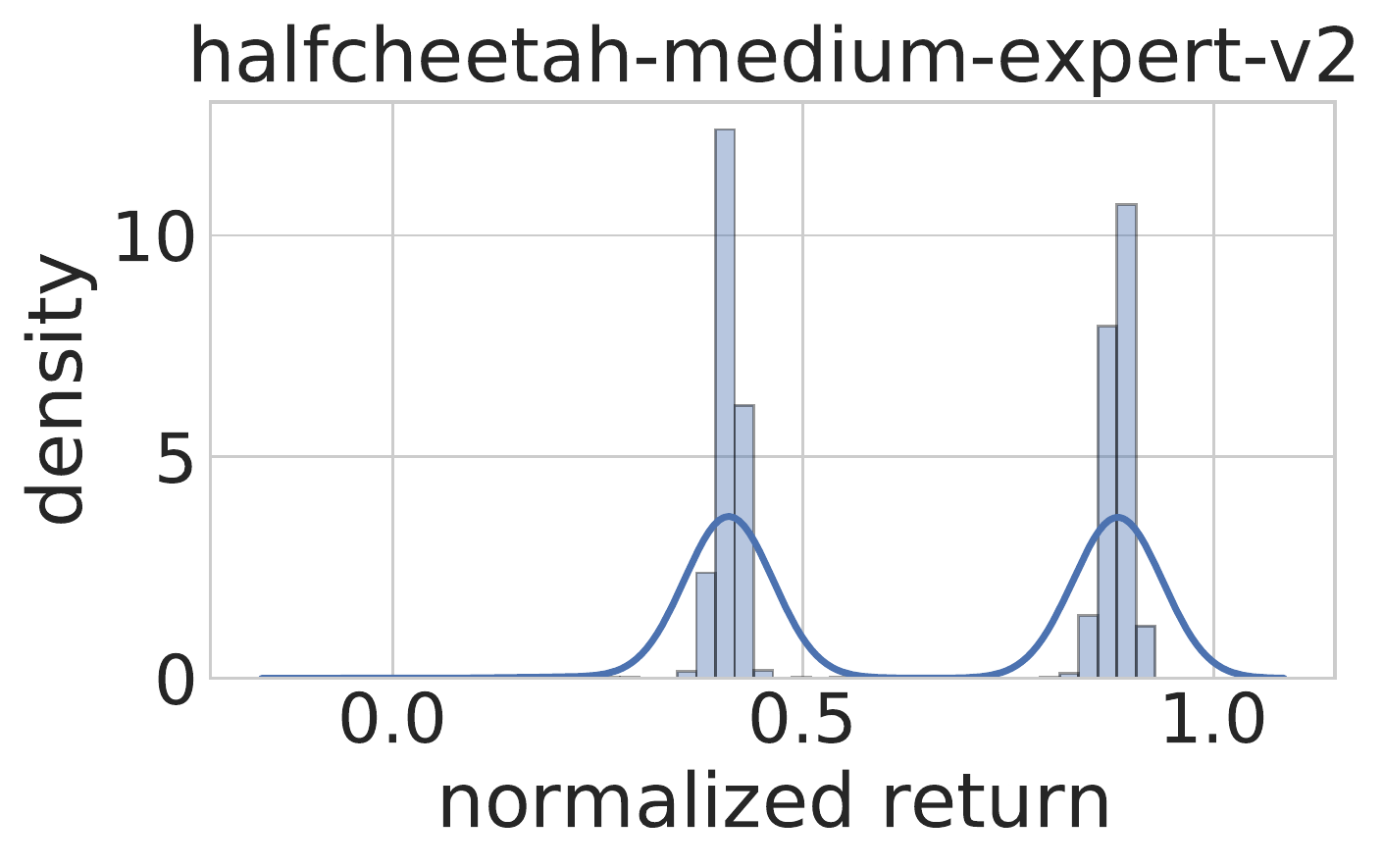}\\
     \caption{The distributions of the normalized returns of the D4RL datasets.}
    \label{fig:density_return}
\end{figure}

\section{Additional Experiments Under the Coupled Setup}
\label{app:main_extra_coupled}

\subsection{Experiments on \med and \medreplay and all \cheetah Datasets }
\label{app:main_med_and_replay}

\begin{figure}[H]
    \centering
    \includegraphics[width=0.8\columnwidth]{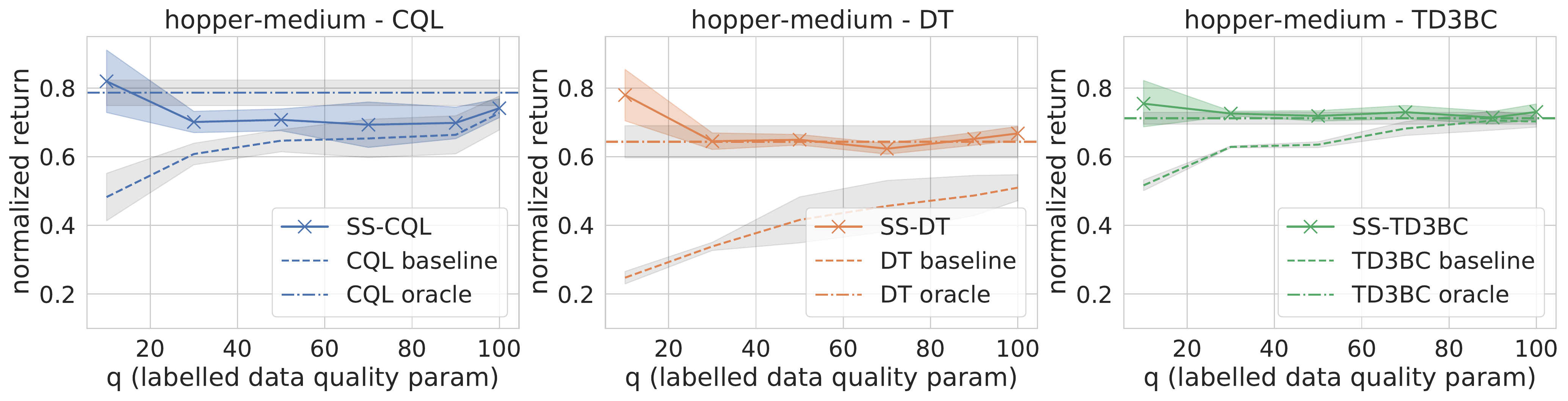}\\
    \includegraphics[width=0.8\columnwidth]{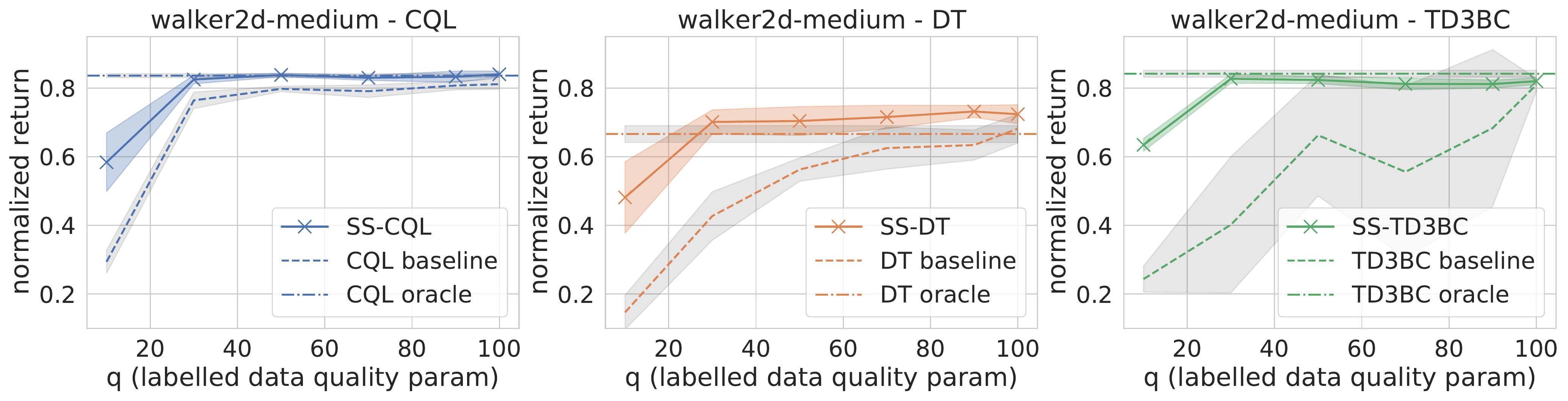}\\
    \caption{The return (average and standard deviation) of \ssa agents trained on the D4RL \medium datasets for \hopper and \walker.}
    \label{fig:main_medium}
    \vskip10pt
    \includegraphics[width=0.85\columnwidth]{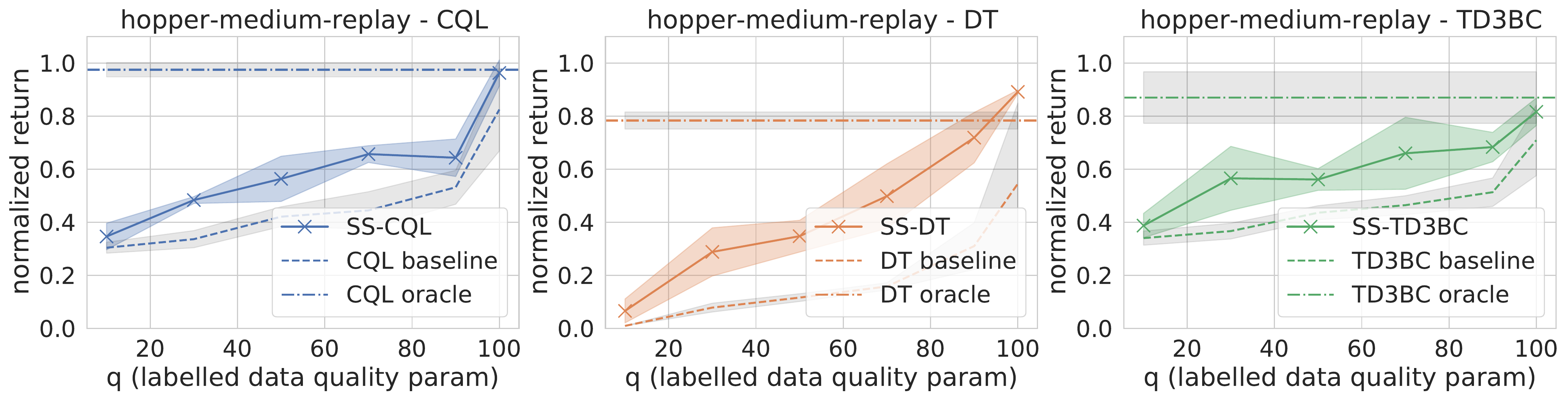}\\
    \includegraphics[width=0.85\columnwidth]{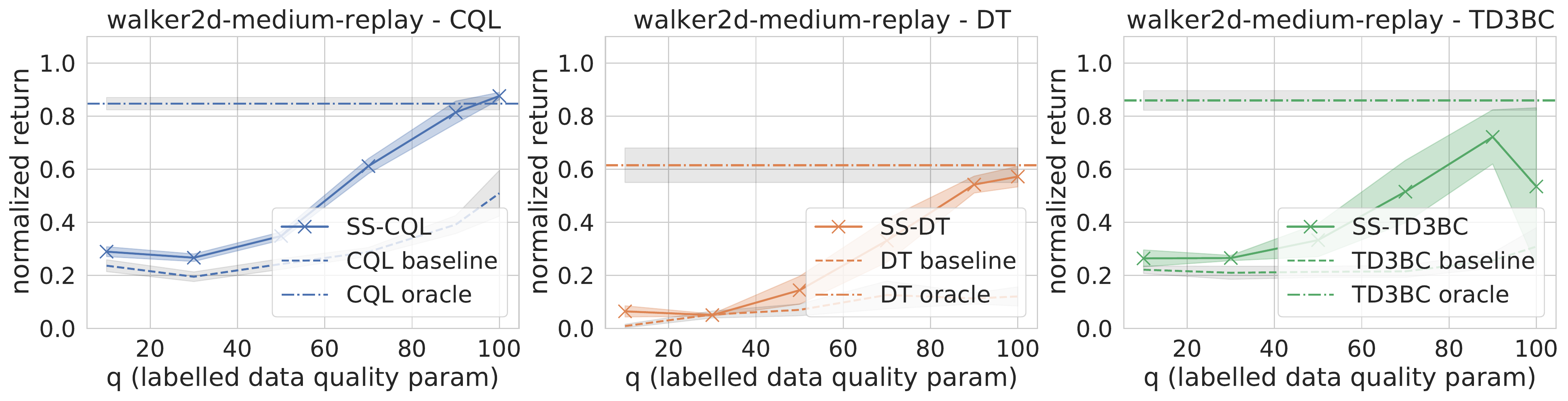}\\
    \caption{The return (average and standard deviation) of \ssa agents on the D4RL \medreplay datasets for \hopper and \walker.}
    \label{fig:main_medium-replay}
\end{figure}

\begin{figure}[H]
    \centering
    \includegraphics[width=0.85\columnwidth]{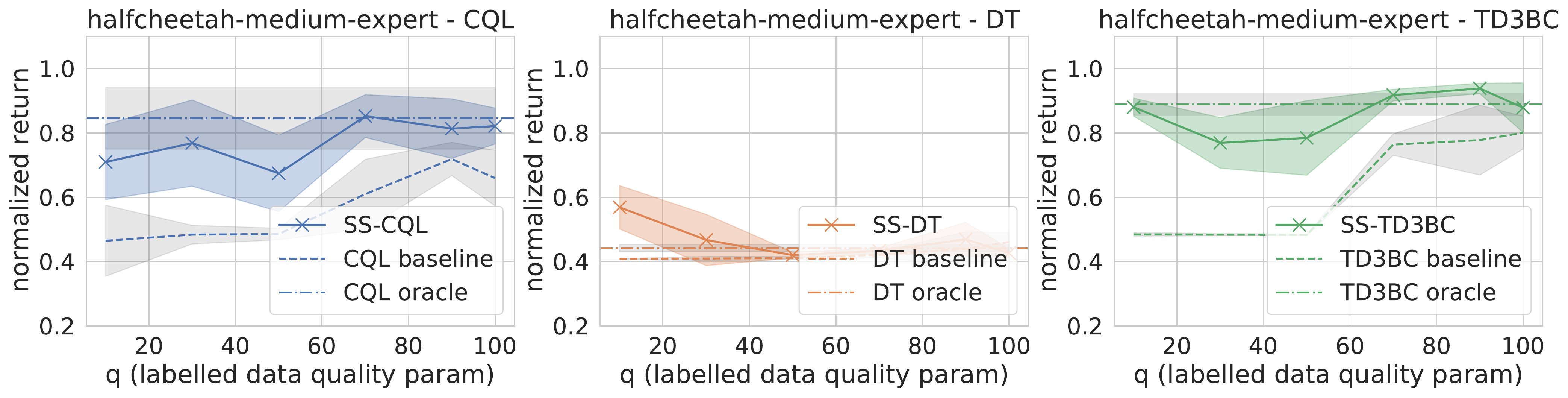}\\
    \includegraphics[width=0.85\columnwidth]{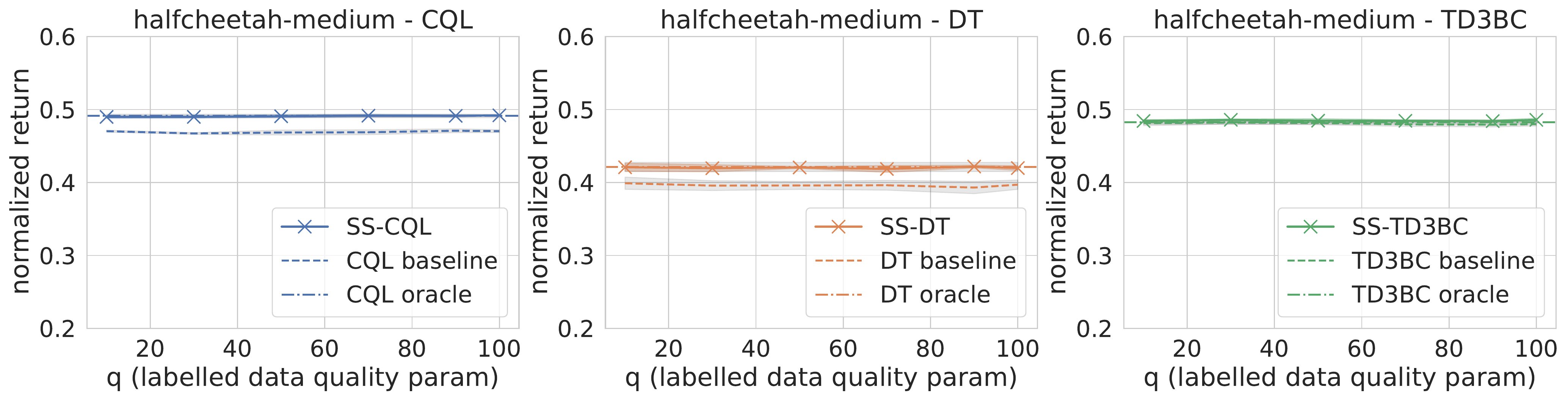}\\
    \includegraphics[width=0.85\columnwidth]{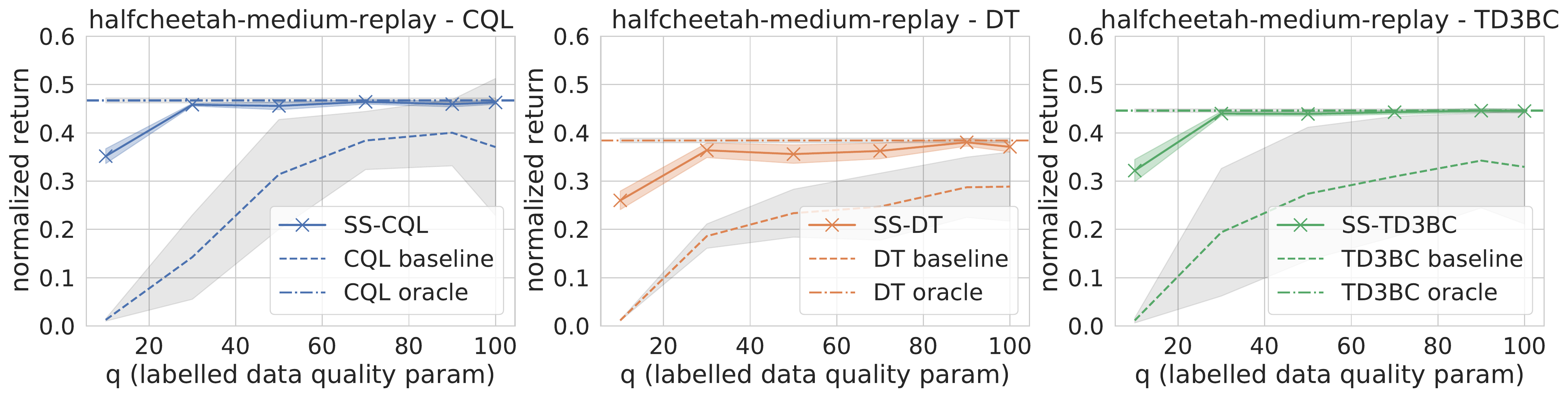}\\
    \caption{The return (average and standard deviation) of \ssa agents on the \cheetah D4RL datasets.}
    \label{fig:main_cheetah}
\end{figure}

We conduct experiments on the \medium and \medreplay datasets of D4RL benchmark for the \hopper and \walker environments, using the same setup as in Section~\ref{sec:expr_main}. Figure~\ref{fig:main_medium} and \ref{fig:main_medium-replay} reports the results. 
For completeness, we also report the results on \medexpert, \medium, and \medreplay datasets for the \cheetah environment in Figure~\ref{fig:main_cheetah}.
We found relatively suboptimal results for DT on the \cheetah environment, consistent with prior results in \citet{zheng2022online}.
The general trend is the same as that in Figure~\ref{fig:main_medium-expert}.  
We note that the results on the \cheetah-\medium dataset are less informative than the others. This is because the data distributions of \cheetah-\medium is very concentrated, similar to a Gaussian distribution with small variance, see Figure~\ref{fig:density_return}.
In such a case, varying the value of $q$ does not drastically change the labelled data distribution. To verify our hypothesis, we conduct experiments on a subsampled dataset in the next subsection.

\subsection{Performance of \ssa on a Subsampled Dataset with Wide Return Distribution}
One may notice that for the \hopper-\medreplay and \walker-\medreplay datasets, \ssa does not catch up with the oracle as quickly as on the other datasets as $q$ increases. Our intuition is that the return distributions of these two datasets concentrate on extremely low values, as shown in Figure~\ref{fig:density_return}. In our experiments, the labelled trajectories for those two datasets have average return small than $0.1$ even when $q=70$. In contrast, the return distributions of the other datasets concentrate on larger values. In contrast, for the other datasets, increasing the value of $q$ will greatly change the returns of labelled trajectories, see Table~\ref{tbl:return_labelled}.

\begin{table}[H]
    \centering
    \begin{tabular}{c|c||c|c|c|c|c}
    \toprule
dataset& q=10 & q=30 & q=50 &  q=70 &  q=90 &  q=100\\
\midrule
hopper-medium-replay   & 0.007   & 0.022 & 0.05 & 0.074 & 0.109 & 0.149 \\
walker2d-medium-replay & -0.002   & 0.005  & 0.023 & 0.048 & 0.087 & 0.156\\
halfcheetah-medium-replay & 0.001   & 0.092  & 0.179  & 0.202 & 0.269 & 0.275\\
hopper-medium  & 0.231 & 0.310 & 0.355 & 0.388 & 0.418 & 0.443\\
walker2d-medium & 0.135 & 0.287 & 0.44 & 0.557 & 0.599 & 0.618\\
halfcheetah-medium & 0.361   & 0.383  & 0.397 & 0.396 & 0.406 & 0.405\\
hopper-medium-expert   & 0.252   & 0.341  & 0.394 & 0.451 & 0.594 & 0.645\\
walker2d-medium-expert & 0.201   & 0.469  & 0.605 & 0.732 & 0.791 & 0.827\\
halfcheetah-medium-expert & 0.377   & 0.397  & 0.405 & 0.537 & 0.604 & 0.638\\
\bottomrule
    \end{tabular}
    \caption{The average return of the labelled trajectories in our experiments. Results aggregated over 5 seeds.}
    \label{tbl:return_labelled}
\end{table}

To demonstrate the performance of \ssa on dataset with a more wide return distribution, we consider a subsampled dataset for the \walker environment generated as follows.
\begin{enumerate}\itemsep0pt
    \item Combine the \walker-\medreplay and \walker-\medium datasets.
    \item Let $R_{\min}$ and $R_{\max}$ denote the minimum and maximum return in the dataset.
    We divide the trajectories into $40$ bins, where the maximum returns within each bin are linear spaced between
    $R_{\min}$ and $R_{\max}$. Let $n_i$ be the number trajectories in bin $i$.
    \item We randomly sample $1000$ trajectories. To sample a trajectory, we first first sample a bin $i \in [1, \ldots, 40]$ with weights proportional to $1/n_i$, then sample a trajectory uniformly at random from the sampled bin.
\end{enumerate}

\begin{figure}[H]
    \centering
    \includegraphics[width=0.38\textwidth]{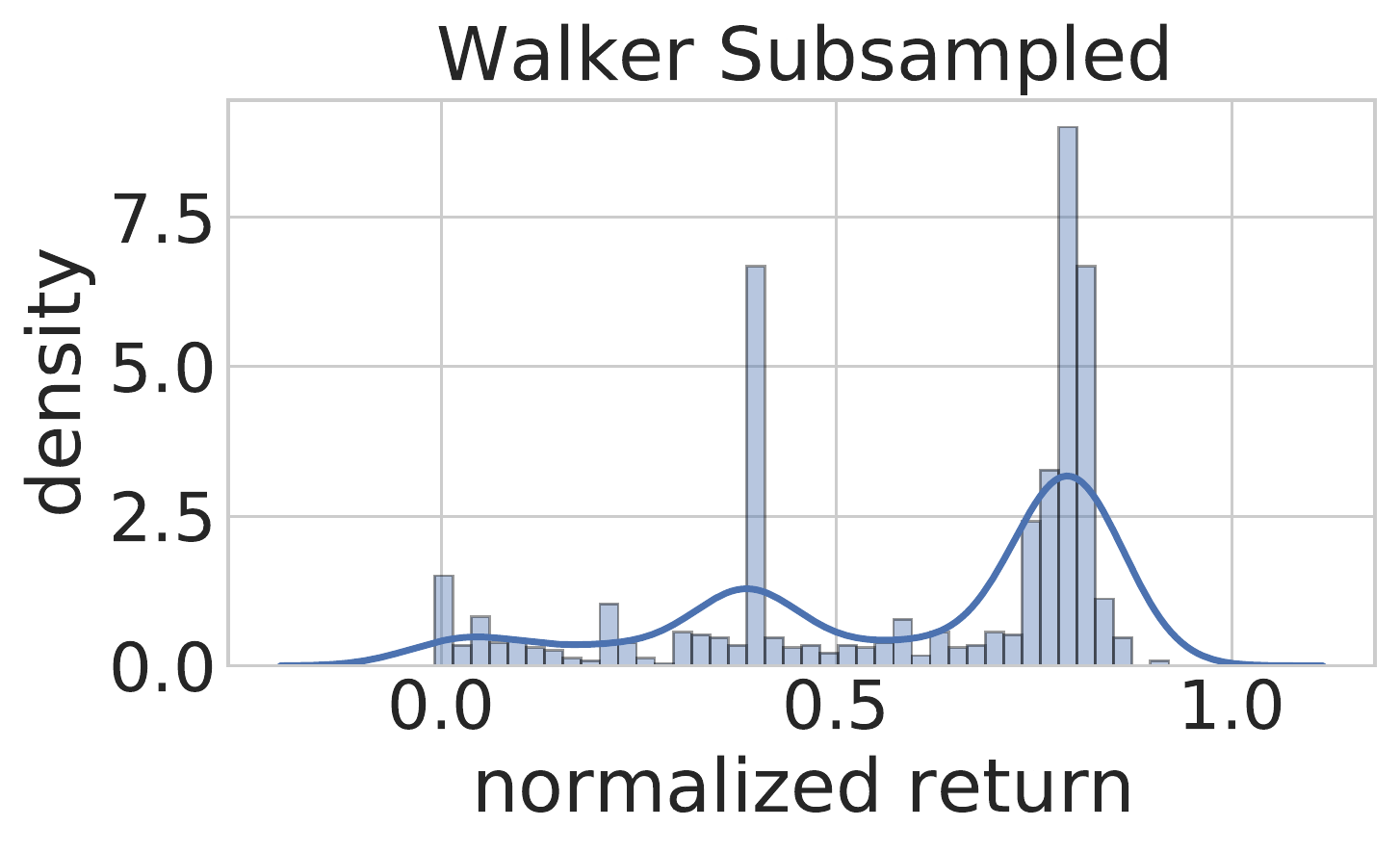}
    \caption{The density of a randomly subsampled dataset of the \walker environment.}
    \label{fig:density_walker_subsampled}
    \vspace{2em}
    \includegraphics[width=0.85\columnwidth]{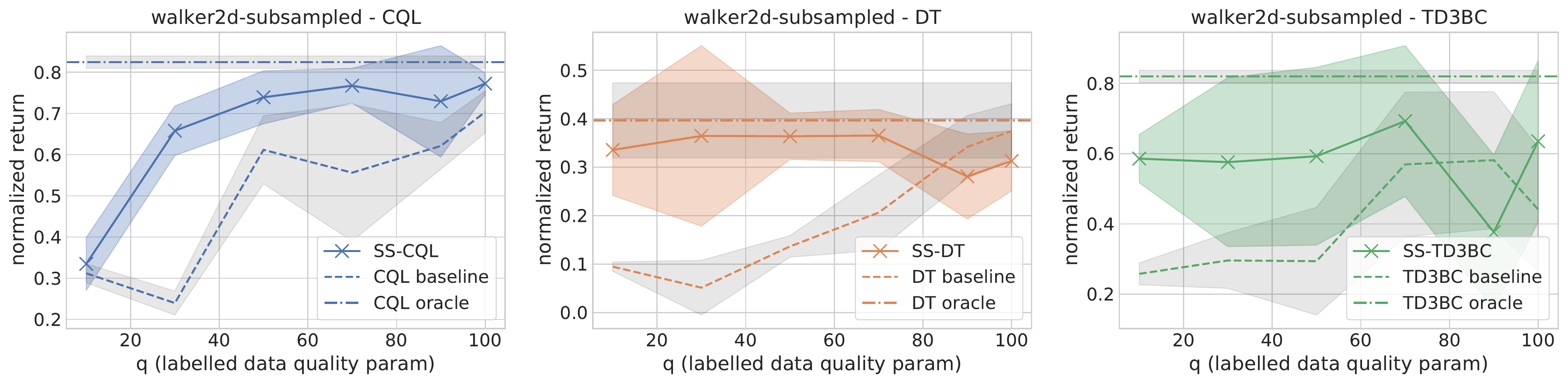}
    \caption{The return (average and standard deviation) of \ssa agents on the subsampled dataset.}
    \label{fig:main_walker_subsampled}
\end{figure}

Figure~\ref{fig:density_walker_subsampled} plots the return distribution of the subsampled dataset. It is wide and has 3 modes. We run the same experiments as before on this subsampled dataset, and Figure~\ref{fig:main_walker_subsampled} plots the results. The general trend is the same as we have found in the above experiments.

\subsection{Results on Low Percentages of Labelled Data}
\label{app:fewer_labelled_data}
\begin{figure}[H]
    \centering
    \includegraphics[width=0.85\columnwidth]{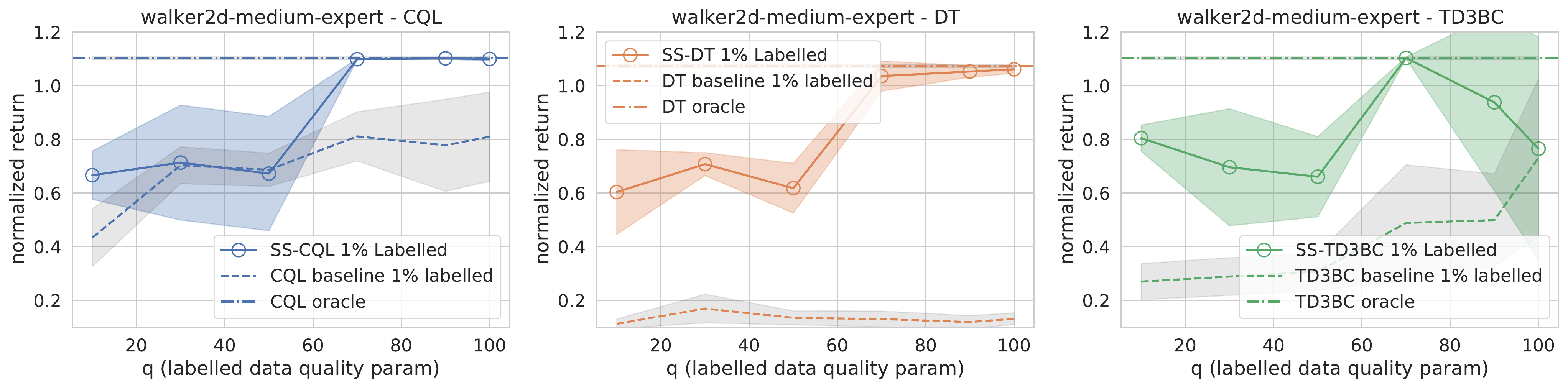}\\
    \includegraphics[width=0.85\columnwidth]{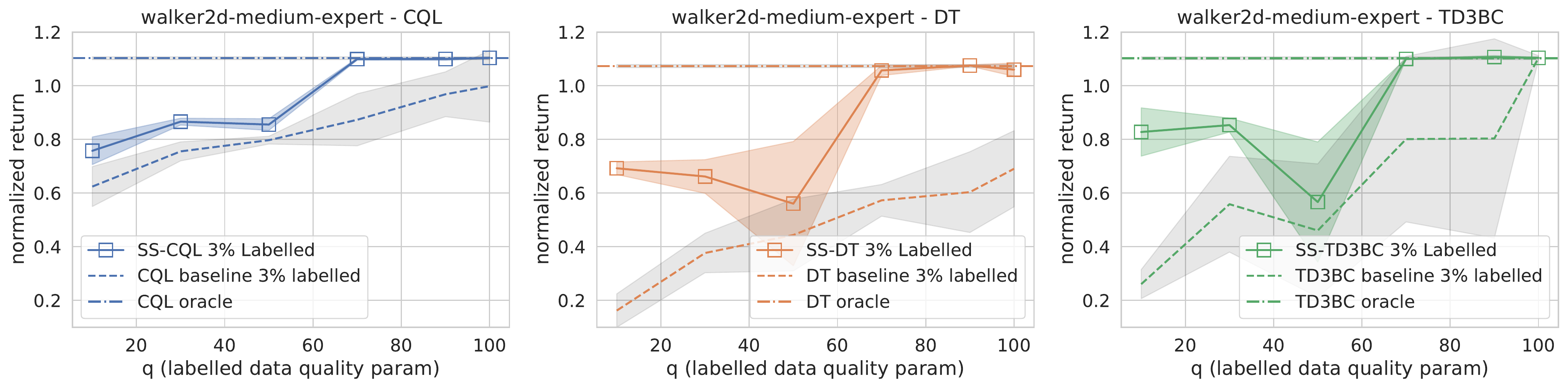}\\
    \includegraphics[width=0.85\columnwidth]{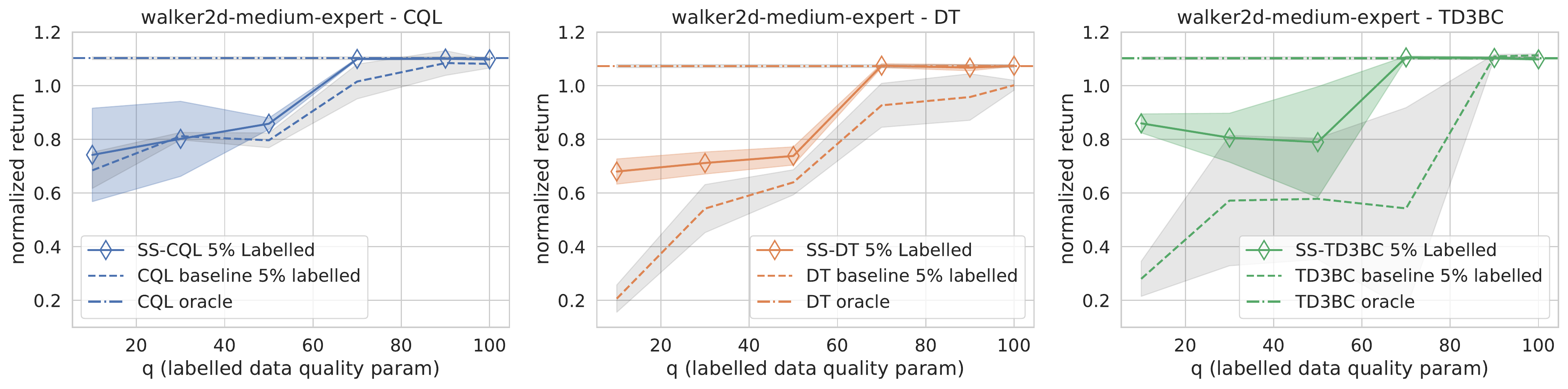}\\
    \includegraphics[width=0.85\columnwidth]{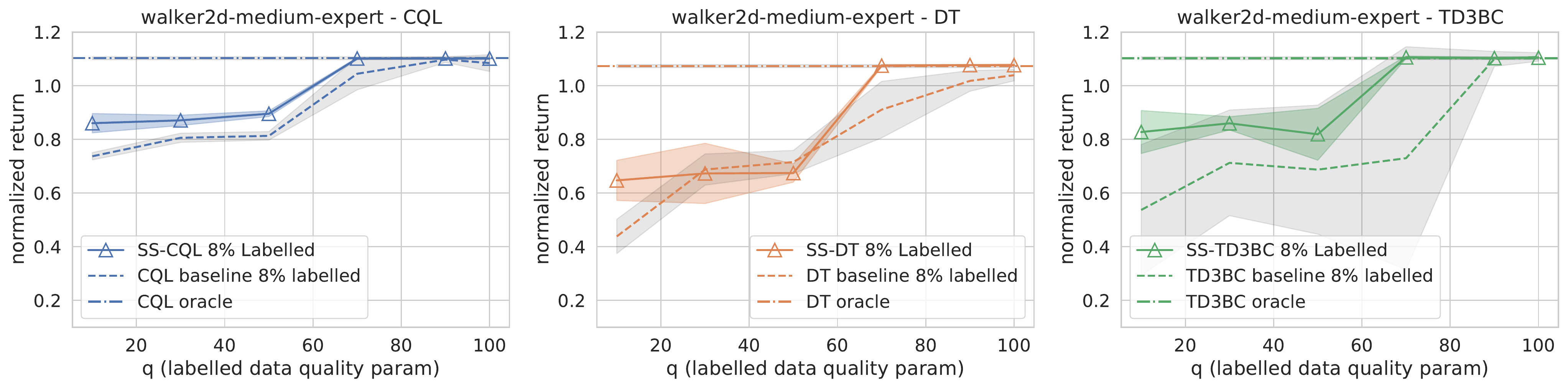}
    \caption{The return (average and standard deviation) of \ssa agents trained on the \walker-\medexpert dataset, when $1\%$, $3\%$, $5\%$ and $8\%$ of the offline trajectories are labelled. 
    }
    \label{fig:main_walker_me_fewer_labelled_data}
\end{figure}

\begin{figure}[H]
    \centering
    \includegraphics[width=0.6\columnwidth]{figure/perf_gap_labelled_1.pdf}\\
    \includegraphics[width=0.6\columnwidth]{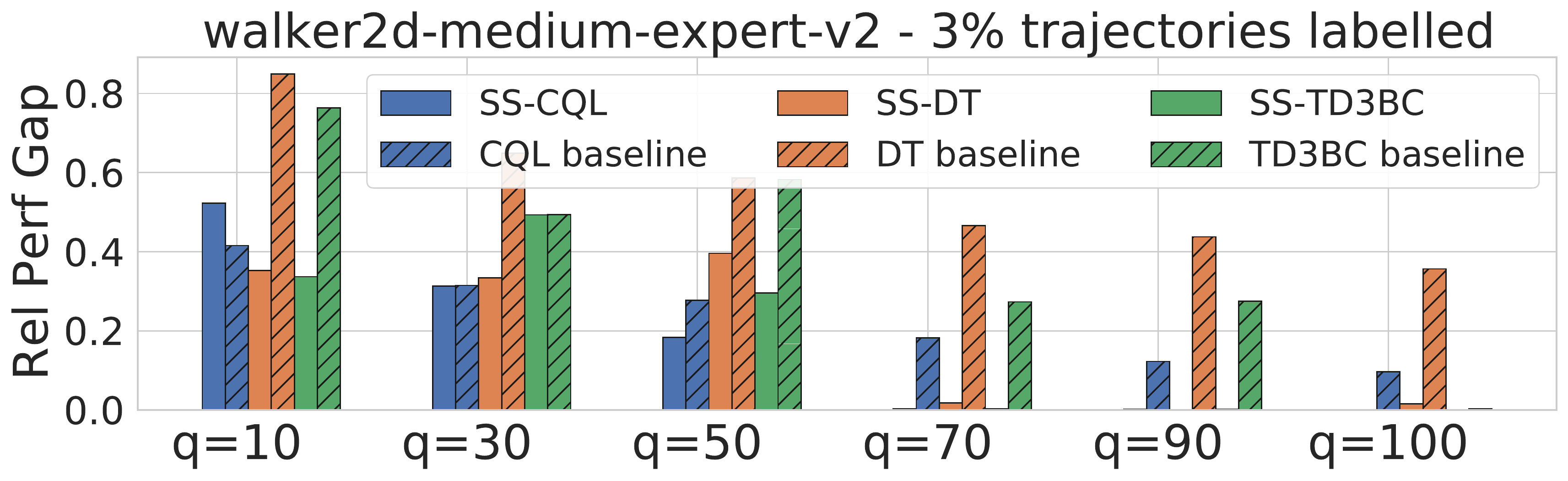}\\
    \includegraphics[width=0.6\columnwidth]{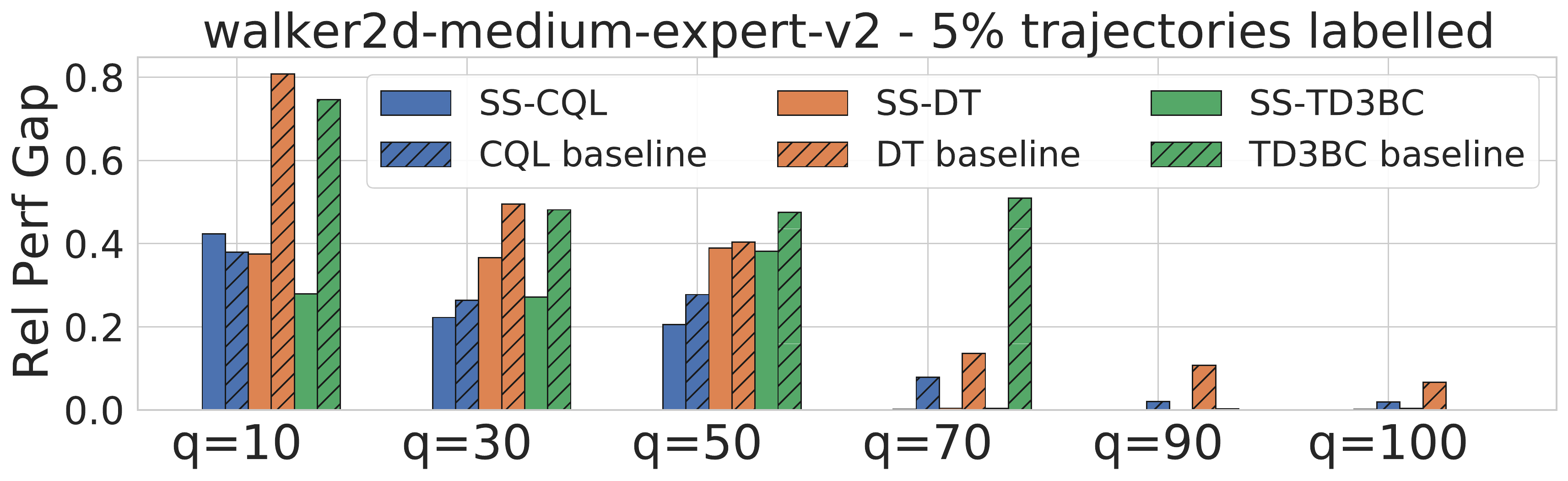}\\
    \includegraphics[width=0.6\columnwidth]{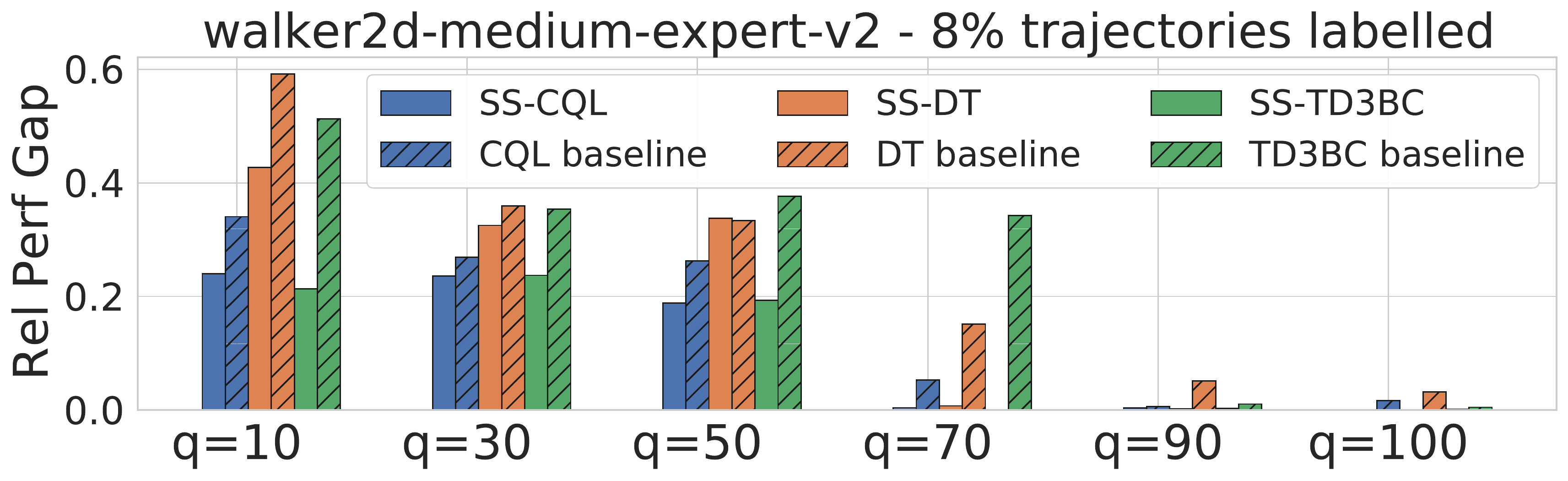}
    \caption{
    The relative performance gap of the \ssa agents and corresponding baselines when $1\%$, $3\%$, $5\%$ and $8\%$ of the offline trajectories are labelled.
    }
    \label{fig:perf_gap_fewer_labelled_data}
\end{figure}

We present the results when the number of the labelled trajectories are  $1\%$, $3\%$, $5\%$, and $8\%$ of the total offline dataset size. Figure~\ref{fig:main_walker_me_fewer_labelled_data} plots the absolute returns and  
Figure~\ref{fig:perf_gap_fewer_labelled_data} plots the relative performance gaps.
We observe the same trend as the experiments in Section~\ref{sec:expr_main}.

\section{Analysis of the Multi-Transition Inverse Dynamics Model}
\label{app:idm_proof}

Given all the past states, we can write
\begin{equation}
\label{eq:idm_all}
\begin{aligned}
p(a_t | s_{t+1}, \ldots, s_1) & = && \frac{p(a_t, s_{t+1}, \ldots, s_1)}{p(s_{t+1}, \ldots, s_1)} \\
& = && \frac{p( s_{t+1} | a_t, s_t, \ldots, s_1) p(a_t | s_t, \ldots, s_1) }{ p( s_{t+1} | s_t, \ldots, s_1) }\\
& = && \frac{p( s_{t+1} | a_t, s_t) p(a_t | s_t, \ldots, s_1) }{ p( s_{t+1} | s_t, \ldots, s_1) } \\
& = && \frac{p( s_{t+1} | a_t, s_t) p(a_t | s_t, \ldots, s_1) }{ \int_{a \in \A} p( s_{t+1} | a_t, s_t) p(a_t | s_t, \ldots, s_1) }, \\
\end{aligned}
\end{equation}
where the last two lines follow from the the Markovian transition property $p( s_{t+1} | a_t, s_t, \ldots, s_1) = p( s_{t+1} | a_t, s_t)$ inherent to a Markov Decision Process.

Let $\beta$ denote the behavior policy.  If $\beta$ is Markovian, then we have 
$p(a_t | s_t, \ldots, s_1) = \beta(a_t | s_t)$ and
it holds that
\begin{equation}
\begin{aligned}
p(a_t | s_{t+1}, \ldots, s_1) 
& = && \frac{p( s_{t+1} | a_t, s_t) \beta(a_t | s_t) }{ \int_{a \in \A }p( s_{t+1}, a | s_t) \beta(a_t | s_t)} \\
& = && p(a_t | s_{t+1}, s_t). 
\end{aligned}
\end{equation}
Similarly, if $\beta$ is non-Markovian and takes account of the previous $k$ states as well, we have
\begin{equation}
\begin{aligned}
p(a_t | s_{t+1}, \ldots, s_1)  =  p(a_t | s_{t+1}, s_t, \ldots, s_{t-k}). 
\end{aligned}
\end{equation}
While the past work commonly models $p(a_t | s_{t+1}, s_t)$~\cite{pathak2017curiosity, pathak18largescale, henaff2022exploration}, in practice, the offline dataset might contain trajectories generated by multiple behavior policies, and it is unknown if any of them is Markovian. 

Our formulation has considered these situations: 1) the behavior policy is non-Markovian, 2) there are multiple behavior policies, 3) and the environment is stochastic.
First, from the above derivation, we can see that choosing $k>0$ allows us to take into account past information before timestep $t$, which naturally copes with non-Markovian policies.
Second, for the case where there are multiple Markovian behavior policies (we assume the behavior policies are Markovian for simplicity), 
we believe it is easier to infer the actual behavior policy by a sequence of past states 
rather than a single one. 
Last, the past work usually predicts $a_t$ via a deterministic function of $(s_t, s_{t+1})$, which implicitly assumes a deterministic environment. In the contrary, our approach has stochasticty, which can potentially better cope with the stochastic environment.
Due to the practical limitation of testing environment and dataset, our experiments only show that the multi-transition IDM outperforms the classic one when the datasets are generated by multiple behavior policies, see Section~\ref{sec:expr_design}. We leave whether the multi-transition IDM provides a better solution to non-Markovian policies and stochastic environments an open question and consider it as one of future work.

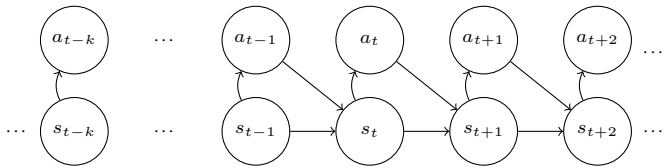
\begin{figure}[h]
\centering
\begin{tikzpicture}[]
\scriptsize{
\node[state,draw=none] (s_0) {};

\node[roundnode] (s_1) [right = 2*\interval of s_0]{$s_{t-k}$};
\node[roundnode] (a_1) [above =\interval of s_1]{$a_{t-k}$};

\node[roundnode] (s_2) [right = 5*\interval of s_1] {$s_{t-1}$};
\node[roundnode] (a_2) [above =\interval of s_2]{$a_{t-1}$};

\node[roundnode] (s_3) [right = 2*\interval of s_2] {$s_t$};
\node[roundnode] (a_3) [above =\interval of s_3]{$a_t$};

\node[roundnode] (s_4) [right = 2*\interval of s_3] {$s_{{t+1}}$};
\node[roundnode] (a_4) [above =\interval of s_4]{$a_{{t+1}}$};

\node[roundnode] (s_5) [right = 2*\interval of s_4] {$s_{{t+2}}$};
\node[roundnode] (a_5) [above =\interval of s_5]{$a_{{t+2}}$};

\node[state,draw=none] [right = 2*\interval of s_5] (s_T) {};
\node[state,draw=none] [above =\interval of s_T] (a_T) {};

}

\draw[->][] (s_2) to (s_3);
\draw[->][] (a_2) to (s_3);

\draw[->][] (s_3) to (s_4);
\draw[->][] (a_3) to (s_4);

\draw[->][] (s_4) to (s_5);
\draw[->][] (a_4) to (s_5);

\draw[->][] (s_1) to[bend left=25] (a_1);
\draw[->][] (s_2) to[bend left=25] (a_2);
\draw[->][] (s_3) to[bend left=25] (a_3);
\draw[->][] (s_4) to[bend left=25] (a_4);
\draw[->][] (s_5) to[bend left=25] (a_5);

\path (s_0) -- node[auto=false]{\ldots} (s_1);
\path (s_1) -- node[auto=false]{\ldots} (s_2);
\path (a_1) -- node[auto=false]{\ldots} (a_2);
\path (s_5) -- node[auto=false]{\ldots} (s_T);
\path (a_5) -- node[auto=false]{\ldots} (a_T);

\end{tikzpicture}
\caption{Graphical model of a Markovian behavior policy (\textit{curved}) within the transition dynamics of an MDP (\textit{straight}). For non-Markovian behavioral policies, we will have additional arrows from $s_{t-k}$ to $a_t$ for $k>0$.}
\label{fig:graphical_model_idm}
\end{figure}

A natural question to ask is whether we should incorporate any future states such as $s_{t+2}$.
Figure~\ref{fig:graphical_model_idm} depicts the graphical model of the state transitions under a MDP.
It is easy to see that given $s_t$ and $s_{t+1}$, $a_t$ is independent of $s_{t+2}$ and all the future states~\cite{koller2009probabilistic}.

In the experiments in Section~\ref{sec:expr_design}, we empirically verify that including future states do not help predicting the actions.
Meanwhile, the transition window size $k$ is a hyperparameter we need to choose. For all our experiments, we use $k=1$ and hence incorporate information about $s_{t-1}$ as well. We ablate this choice in Section~\ref{sec:expr_design}, see Figure~\ref{fig:idm_arch_perf_iqm}.

\newpage
\section{Self-Training for IDM}
\label{app:self_training}

We present the self-training algorithm used to train the IDM in Algorithm~\ref{algo:self_training}.
In each training round, we randomly sample $10\%$ of the training data as the validation set.
During the training of each individual IDM, we select the model that yields the best validation error in $100$k iterations.

\begin{algorithm}[h]
\DontPrintSemicolon
\caption{Self-Training for the Inverse Dynamics Model}\label{algo:self_training}
\textbf{Input:} labelled data $\Dlabel$, unlabelled data $\Dunlabel$,
IDM transition size $k$, ensemble size $m$,
number of augmentation rounds $N$\;
\tcp{initialize the training set}
$\D \leftarrow \Dlabel$\;
\tcp{train $m$ independent IDMs using the labelled data under the randomness of initialization and data shuffling}
$\hat{\theta}_i \leftarrow \argmin_{\theta} \sum_{(a_t, \vs_{t, -k})\; \text{in}\; \D}\left[ -\log \phi_\theta(a_t | \vs_{t, -k})  \right]$,  $i \in [m]$\;
\vskip5pt
\tcp{compute the augmentation size}
$n_\text{aug} \leftarrow \abs{\Dunlabel} / N$\\
\For{round $1, \ldots, N$}{
    \tcp{compute the estimation uncertainty }
        \For{every $(a_t, \vs_{t, -k}) \in \Dunlabel$}{
            $\nu_t \leftarrow$ variance of the Gaussian mixture $\frac{1}{m} \sum_{i=1}^m \N \left(\mu_{\hat{\theta}_i}(\vs_{t, -k}), \, \Sigma_{\hat{\theta}_i}(\vs_{t, -k}) \right)$\\
        }
        \vskip5pt
        \tcp{move examples with lowest uncertainties into the training set}
        $\D_\text{subset} \leftarrow \set{ (a_t, \vs_{t, -k}) | \nu_t \; \text{among the lowest} \; n_\text{aug} \; \text{in} \; \Dunlabel}$\\
        $\D \leftarrow \D \bigcup \D_\text{subset}$ \\
        $\Dunlabel \leftarrow \Dunlabel \backslash \D_\text{subset}$ \\
        \vskip5pt
        \tcp{train IDMs again}
        $\hat{\theta}_i \leftarrow \argmin_{\theta} \sum_{(a_t, \vs_{t, -k})\; \text{in}\; \D}\left[ -\log \phi_\theta(a_t | \vs_{t, -k})  \right]$, $i \in [m]$\;
}
\textbf{Output: $\hat{\theta}_1, \ldots, \hat{\theta}_m$}\;
\end{algorithm}

\section{Implementation Details of \gato}
\label{app:gato}
Inspired by \texttt{GATO}, the multi-task and multi-modal generalist agent proposed by~\citet{reed2022generalist},
we consider \gato, a variant of \dt that can incorporate the unlabelled data into policy training.
\gato is trained on the labelled and unlabelled data simultaneously. The implementation details are:
\begin{itemize}
    \item We form the same input sequence as \dt, where we fill in zeros for the missing actions for unlabelled trajectories. 
    \item  For the labelled trajectories, \gato predicts the actions, states and rewards; for the unlabelled ones, \gato only predicts the states and rewards. 
    \item We use the stochastic policy as in online decision transformer~\cite{zheng2022online} to predict the actions. 
    \item We use deterministic predictors for the states and rewards, which are single linear layers built on top of the Transformer outputs.
\end{itemize}  

\begin{figure}[t]
    \centering
    \includegraphics[width=0.75\textwidth]{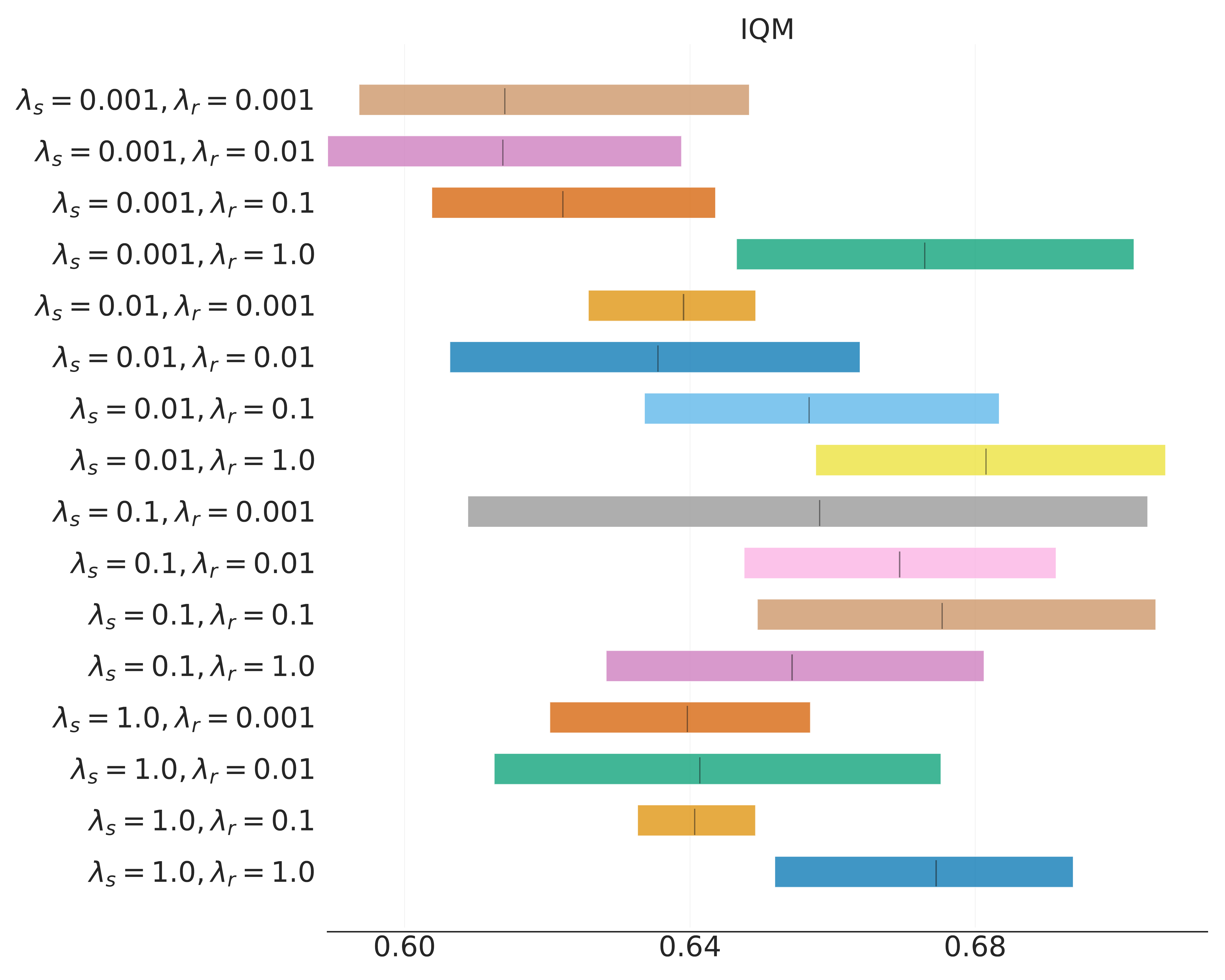}
    \caption{The $95\%$ stratified bootstrap CIs of the interquartile mean of the returns obtained by \gato agents, with different combinations of regularization parameters.}
    \label{fig:gato_lambda}
\end{figure}

Let $g_t = \sum_{t'=t}^{|\tau|i} r_{t'}$ be the return-to-go of a trajectory $\tau$ at timestep $t$. 
Let $H^{\Plabel}_{\theta}$ denotes the policy entropy included on the labelled data distribution.
For simplicity, we assume the context length of \gato is $1$, and Equation~\eqref{eq:gato} shows the training objective of \gato. (We refer the readers to \citet{zheng2022online} for the formulation with a general context length and more details.)
\begin{equation}
\begin{aligned}
    \label{eq:gato}
    \min_{\theta} &  &&
    \E_{(a_t, s_t, r_t, g_t) \sim \Plabel} \left\{ -\log \pi(a_t | s_t, g_t, \theta) +  
    \lambda_s \norm{s_t - \hat{s}_t(\theta)}^2_2 + \lambda_r \norm{r_t - \hat{r}_t(\theta)}^2_2 \right\}  \\
    & && + \E_{(s_t, r_t, g_t) \sim \Punlabel }  \left\{ \lambda_s \norm{s_t - \hat{s}_t(\theta)}^2_2 + \lambda_r \norm{r_t - \hat{r}_t(\theta)}^2_2 \right\} \\
    \text{s.t.} & && H^{\Plabel}_{\theta}[a | s, g] \geq \nu
\end{aligned}
\end{equation}
The constant $\nu$, $\lambda_s$ and $\lambda_r$ are prefixed hyper-parameters, where $\nu$ is the target policy entropy, and $\lambda_s$ and $\lambda_r$ are regularization parameters used to balance the losses for actions, states, and rewards. We use $\nu = -\text{dim}(\A)$ as for DT (see Appendix~\ref{app:hp}). To choose the regularization parameters $\lambda_s$ and $\lambda_r$ for \gato, we test 16 combinations where $\lambda_s$ and $\lambda_r$ are $1.0, 0.1, 0.01$ and $0.001$ respectively. We run experiments as in Section~\ref{sec:expr_main} for $q=10, 30, 50, 70, 90, 100$, and compute the confidence intervals for the aggregated results. Figure~\ref{fig:gato_lambda} shows that $\lambda_s = 0.01$ and $\lambda_r=0.1$ yield the best performance, and we use them in our experiments for Figure~\ref{fig:gato}.

\clearpage
\section{Influences of the Labelled and Unlabelled Data Size}
\label{app:influence_size}
\begin{figure}[H]
    \centering
    \begin{subfigure}[b]{\columnwidth}
        \centering
        \includegraphics[width=0.7\columnwidth]{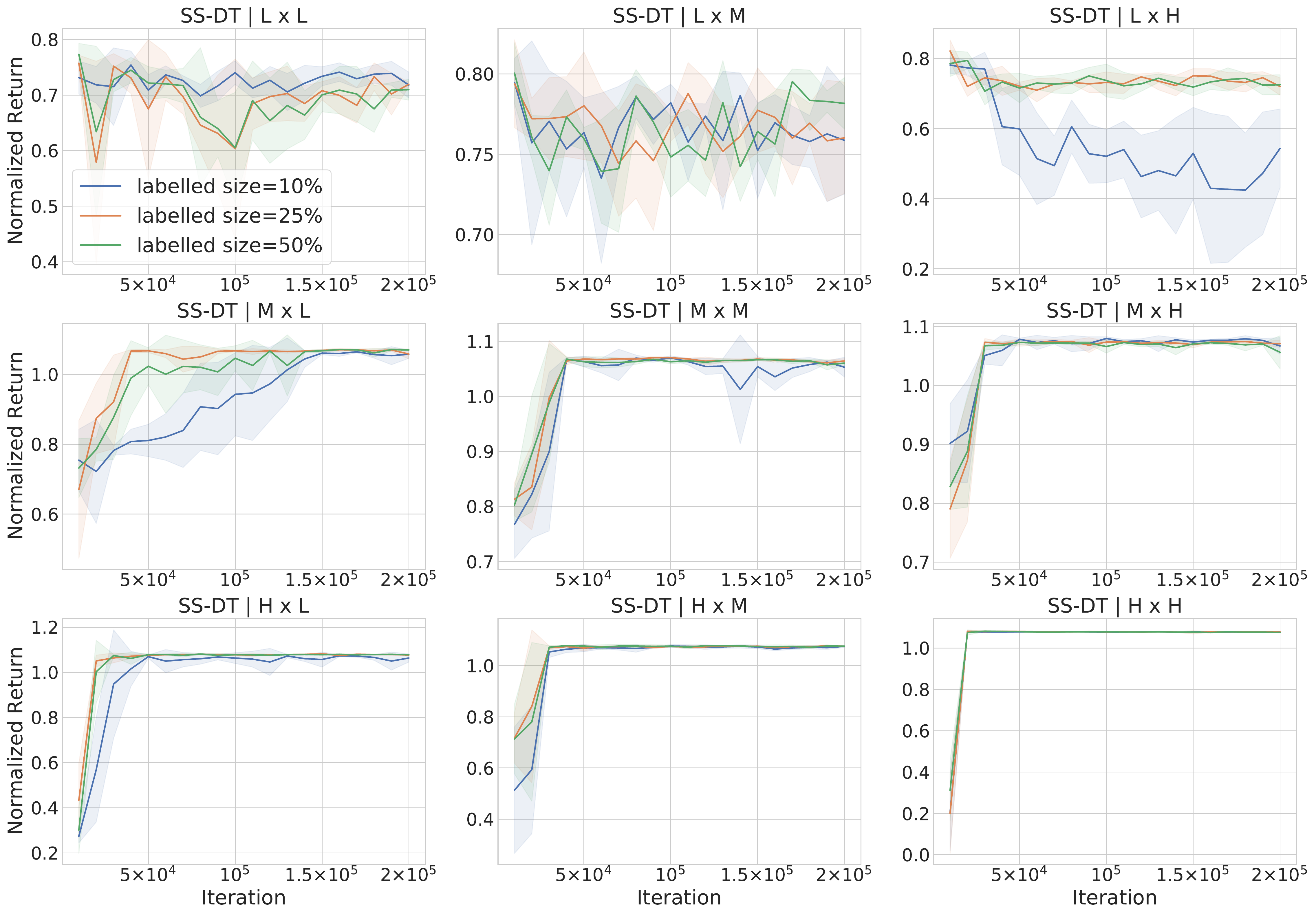}\\
        \caption{Results of \ssdt.}
        \label{fig:decouple_size_labelled_hopper}
    \end{subfigure}
    \begin{subfigure}[b]{\columnwidth}
        \centering
        \includegraphics[width=0.7\columnwidth]{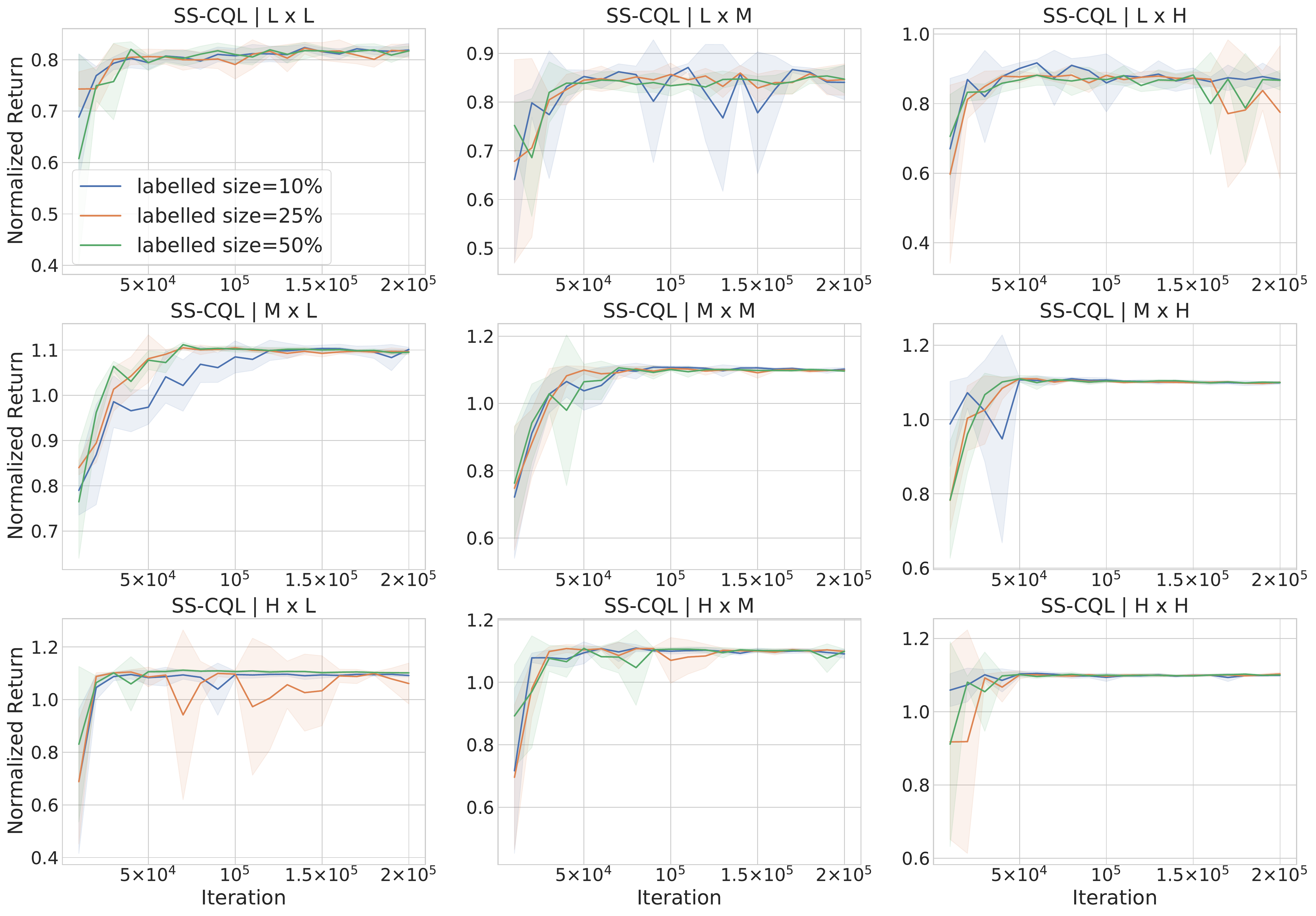}
        \caption{Results of \sscql.}
        \label{fig:decouple_size_labelled_walker}
    \end{subfigure}
    \caption{The return (average and standard deviation) of \ssdt and \sscql agents trained on the \walker-\medexpert datasets with different sizes of labelled data. The unlabelled data size is fixed to be $10\%$ of the offline dataset size. Results aggregated over 5 instances with different seeds.
    }
    \label{fig:decouple_size_labelled}
\end{figure}

\begin{figure}[H]
    \centering
    \begin{subfigure}[b]{\columnwidth}
        \centering
        \includegraphics[width=0.7\columnwidth]{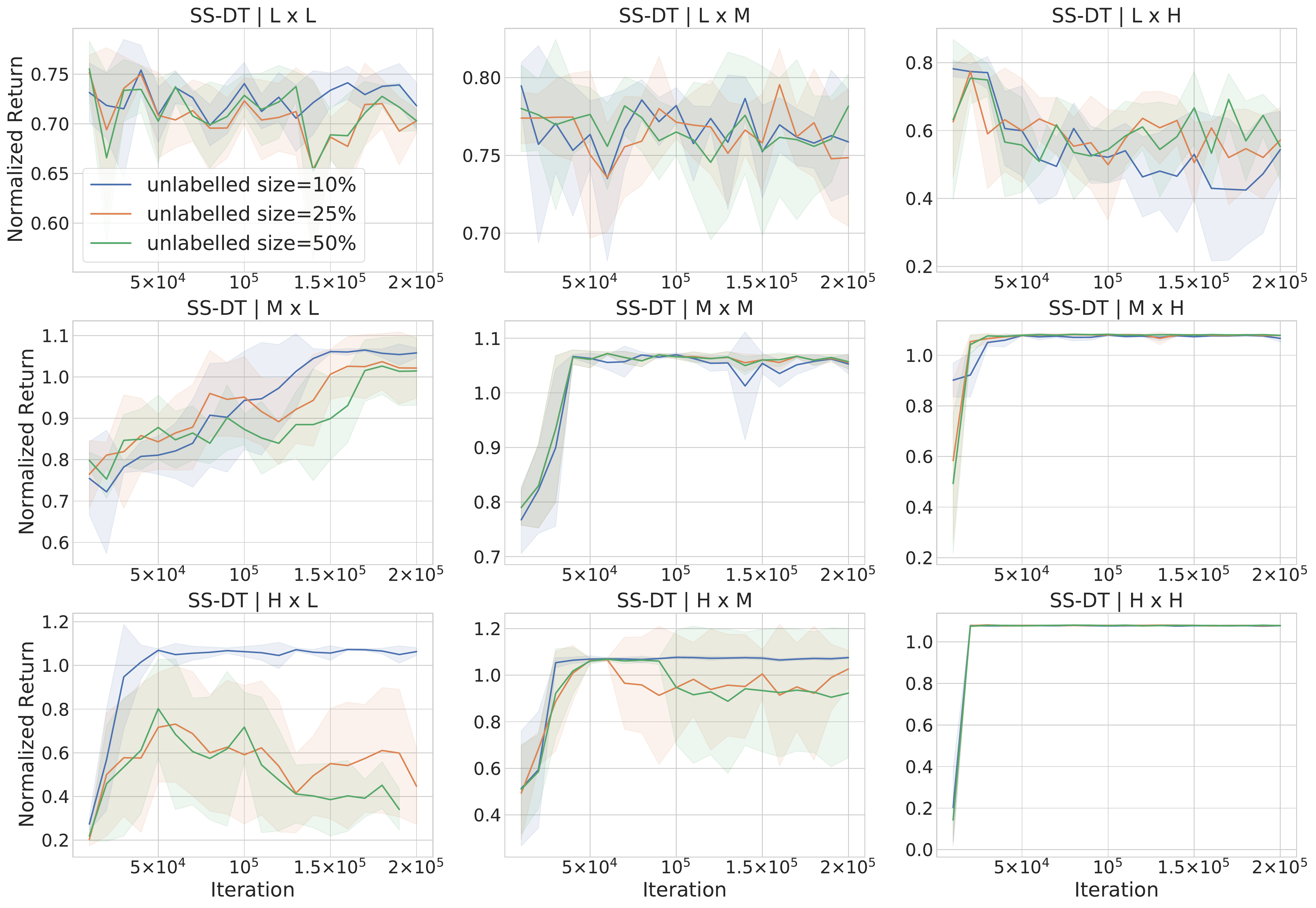}\\
        \caption{Results of \ssdt.}
        \label{fig:decouple_size_unlabelled_dt}
    \end{subfigure}
    \begin{subfigure}[b]{\columnwidth}
        \centering
        \includegraphics[width=0.7\columnwidth]{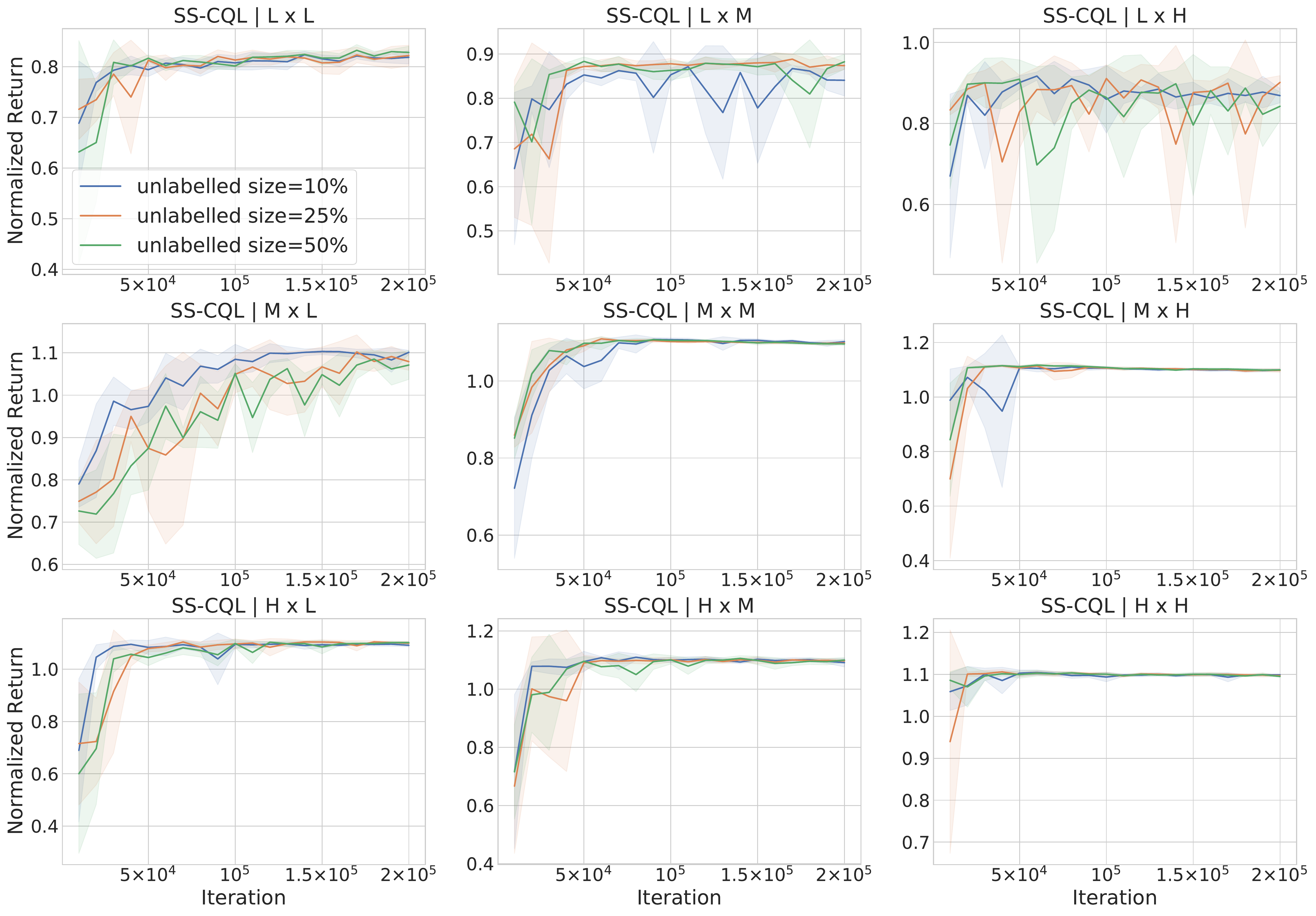}
        \caption{Results of \sscql.}
        \label{fig:decouple_size_unlabelled_cql}
    \end{subfigure}
    \caption{The return (average and standard deviation) of \ssdt and \sscql agents trained on the \walker-\medexpert datasets with different sizes of unlabelled data. The labelled data size is fixed to be $10\%$ of the offline dataset size. Results aggregated over 5 instances with different seeds.
    }
    \label{fig:decouple_size_unlabelled}
\end{figure}

Figure~\ref{fig:decouple_size_labelled} plots the average return of \ssdt and \sscql when we
 vary the number of labelled trajectories while fixing the number of unlabelled trajectories. As described in Section~\ref{sec:expr_data}, we consider $9$ data setups
where the labelled and unlabelled trajectories are sampled from
\texttt{Low}, \texttt{Medium} and \texttt{High} groups.
In all the plots, \texttt{L x H} denotes the setup where the labelled data are sampled from \texttt{Low} group and
the unlabelled data are sampled from \texttt{High} group.
Similarly, Figure~\ref{fig:decouple_size_unlabelled} plots the results when we vary the number of unlabelled trajectories, while the number of labelled ones is fixed.

\newpage
\section{Additional Experiments on the Maze2d Environment}
\label{app:additional_expr_maze2d}
The \maze environment involves moving force-actuated ball to a fixed target location. The observation consists of the location and velocities, and the reward is the negative exponentiated distance to the target location.

\begin{figure}[H]
    \centering
    \includegraphics[width=0.3\columnwidth]{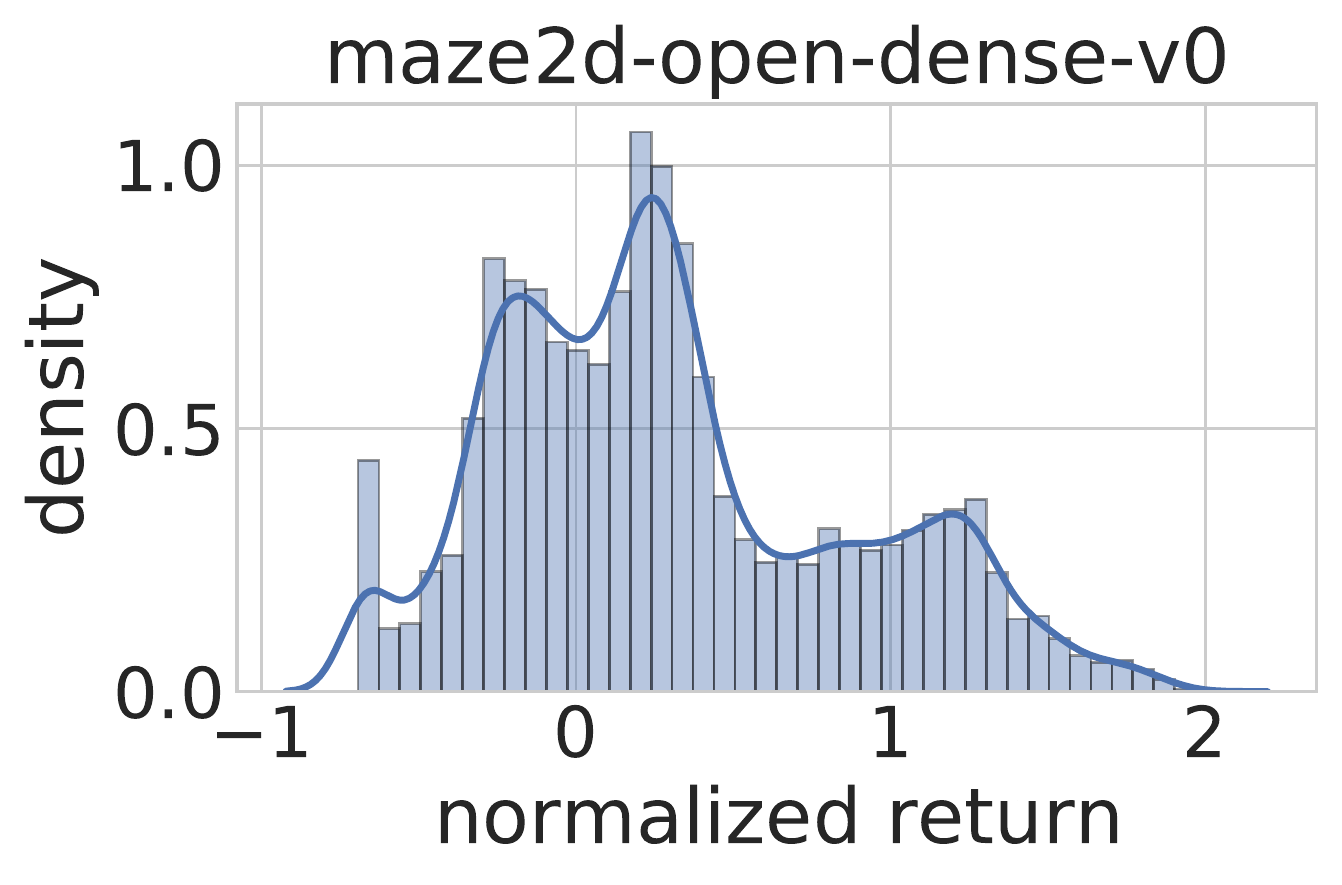}
    \includegraphics[width=0.3\columnwidth]{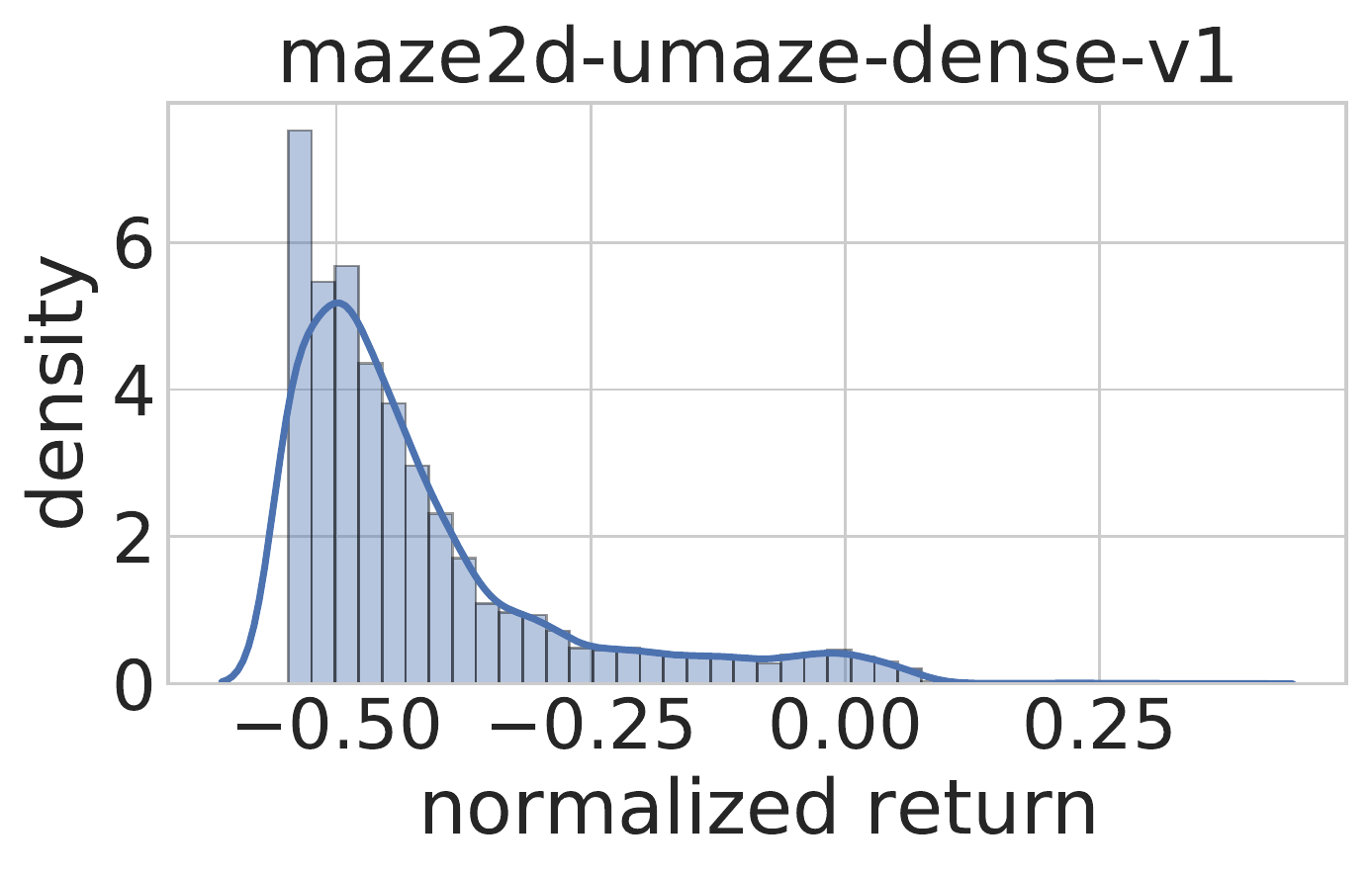} \\
    \includegraphics[width=0.3\columnwidth]{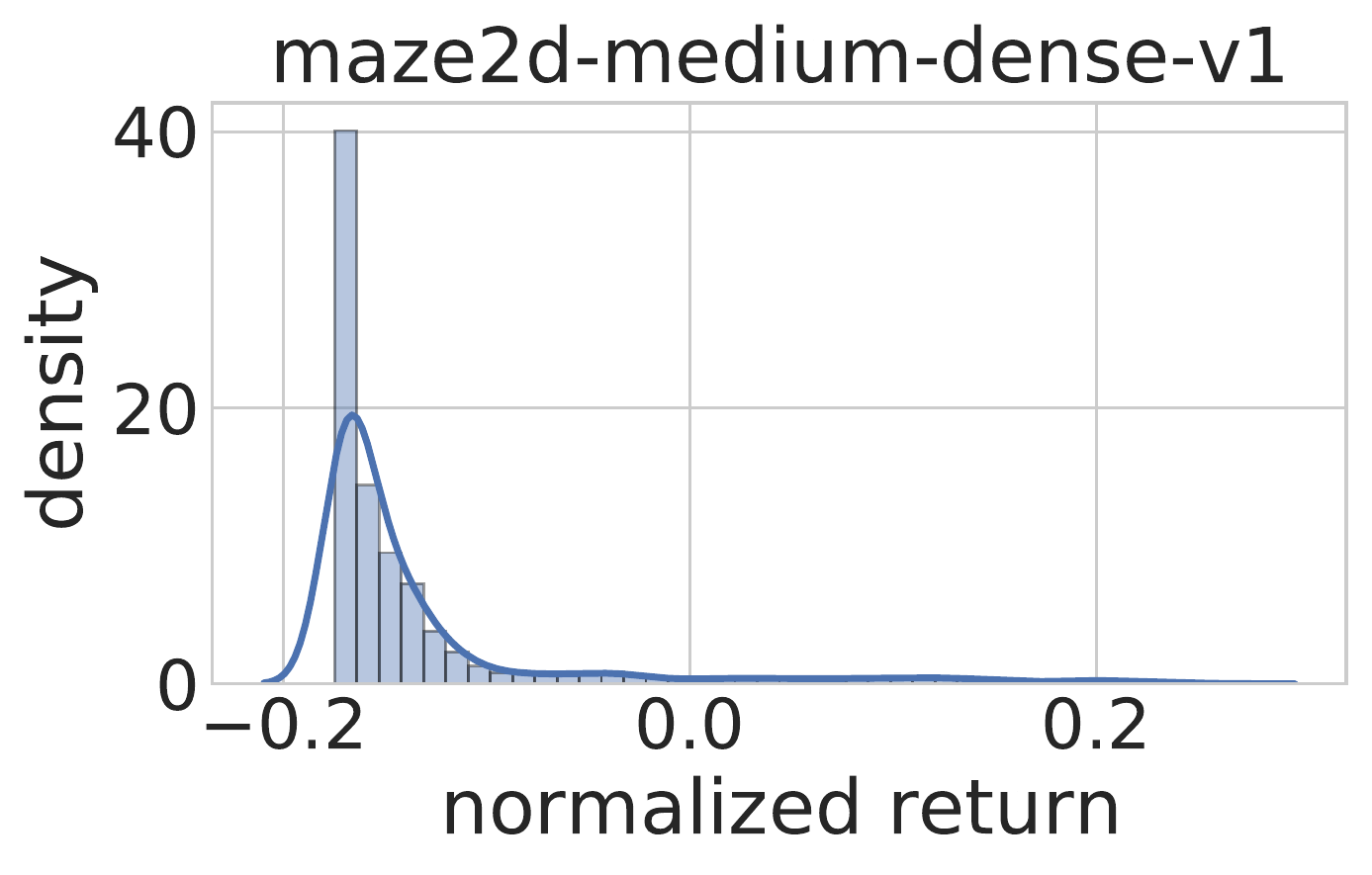}
    \includegraphics[width=0.3\columnwidth]{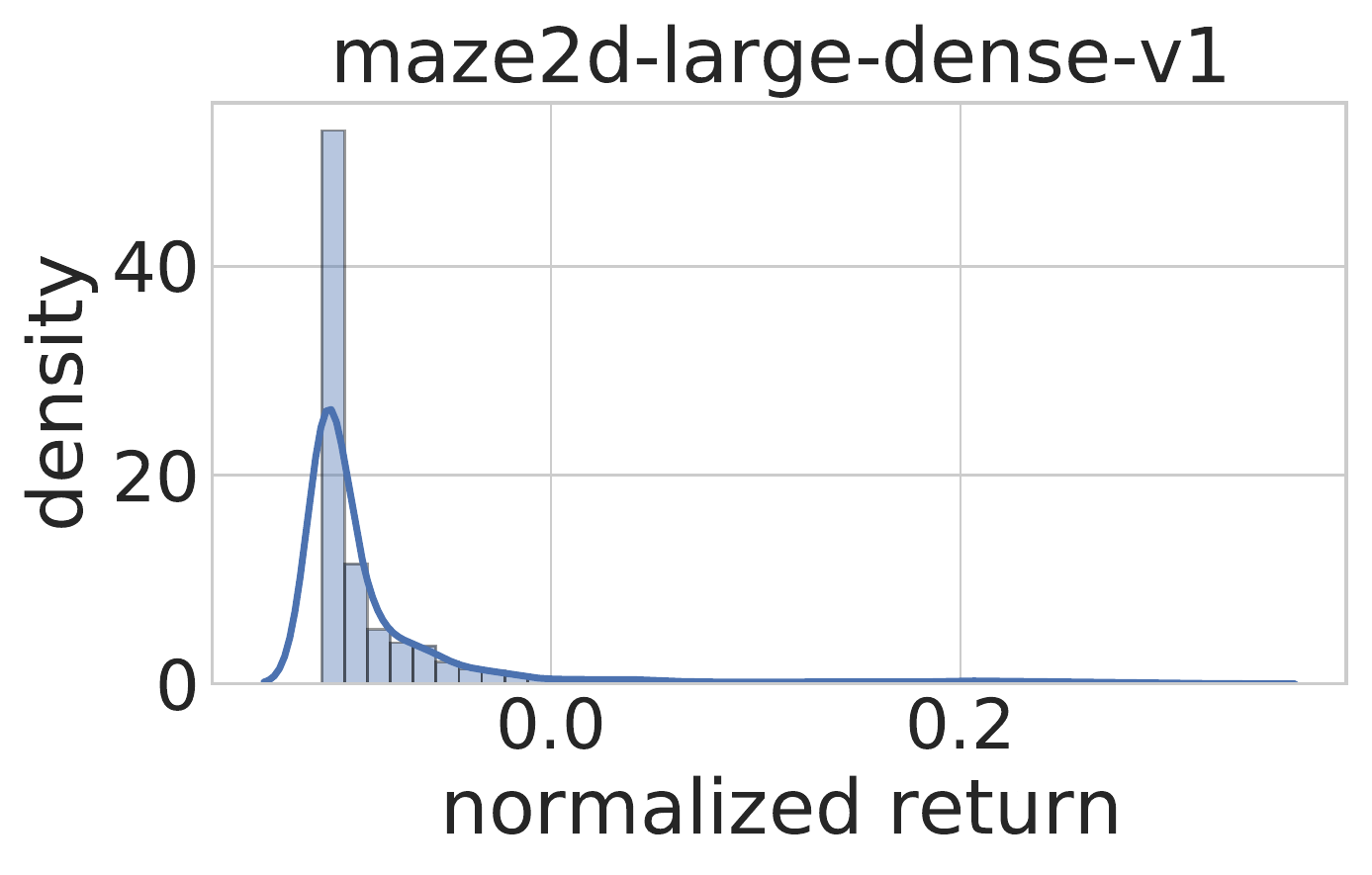}
    \caption{The distributions of the normalized returns of the \maze datasets.}
    \label{fig:density_return_maze2d}
\end{figure}

We conduct experiments on four offline dataset for the \maze environments, each corresponds to a different map: 
\texttt{maze2d-open-dense-v0}, 
\texttt{maze2d-umaze-dense-v1}, 
\texttt{maze2d-medium-dense-v1}, and
\texttt{maze2d-large-dense-v1}.
Figure~\ref{fig:density_return_maze2d} plots the normalized return distributions of these four datasets. The return distribution of \texttt{maze2d-open-dense-v0} is widely spread, while the others are heavily skewed towards the low return values. Note that many of the trajectories' normalized returns are below zero. 

We train \ssa agents instantiated with \td, under the coupled setup as in Section~\ref{sec:expr_main}. We use learning rate value $0.0001$ for both actors and critics, which is smaller than what we used for locomotion tasks. All the other hyperparameters are the same as described in Appendix~\ref{app:hp}.

Figure~\ref{fig:ssorl_main_maze2d_td3bc} plots the results. The general trend is similar to what we have seen in previous experiments for locomotion tasks. 

\begin{figure}[H]
    \centering
    \includegraphics[width=0.95\columnwidth]{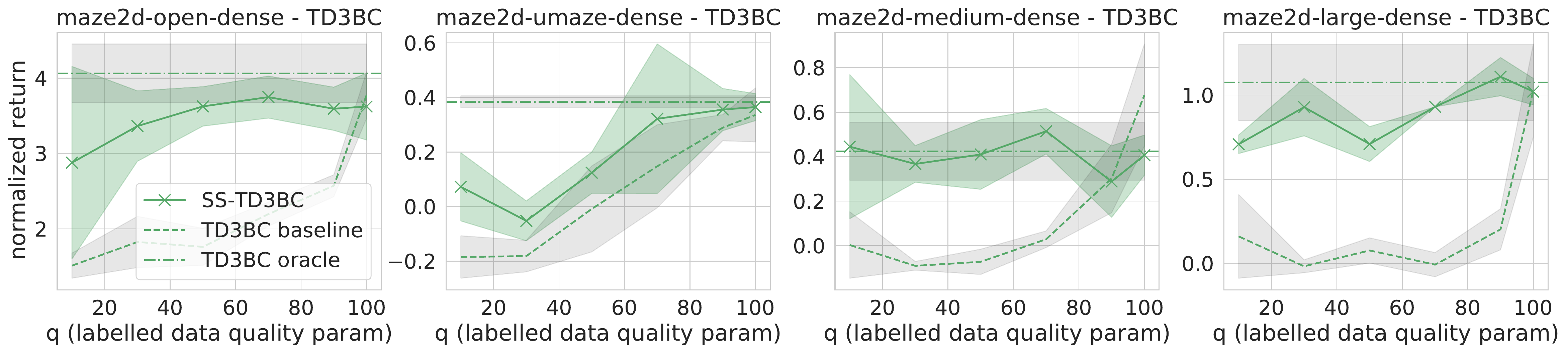}
    \caption{The return (average and standard deviation) of \sstd agents trained on the \maze dataset.}
    \label{fig:ssorl_main_maze2d_td3bc}
\end{figure}
\end{document}